\def\year{2021}\relax
\documentclass[letterpaper]{article} 
\usepackage{aaai21}  
\usepackage{times}  
\usepackage{helvet} 
\usepackage{courier}  
\usepackage{graphicx} 
\usepackage{natbib}  
\usepackage{caption} 
\usepackage{subfig}
\usepackage{amsmath}
\usepackage{amssymb}
\usepackage{booktabs}
\usepackage{color}

\title{Context-aware Attentional Pooling (CAP) for Fine-grained Visual Classification}
\author{
    Ardhendu Behera, Zachary Wharton, Pradeep Hewage, Asish Bera
    \\
}
\affiliations{
    Department of Computer Science, Edge Hill University\\

    St Helen Road, Lancashire\\
    United Kingdom, L39 4QP\\
    beheraa@edgehill.ac.uk, zachary.wharton@go.edgehill.ac.uk, pradeep.hewage@edgehill.ac.uk, beraa@edgehill.ac.uk

}

\usepackage[firstpage]{draftwatermark}
\SetWatermarkFontSize{1cm}
\SetWatermarkAngle{90}
\usepackage[printwatermark]{xwatermark}
\usepackage{xcolor}
\begin{document}
\newwatermark[pagex={1},fontfamily=bch,fontsize=11pt,color=gray,angle=0,scale=1.1,xpos=0,ypos=12cm]{Extended version of the accepted paper in $35^{th}$ AAAI Conference on Artificial Intelligence 2021}
\SetWatermarkText{AAAI 2021}

\maketitle

\begin{abstract}
Deep convolutional neural networks (CNNs) have shown a strong ability in mining discriminative object pose and parts information for image recognition. For fine-grained recognition, context-aware rich feature representation of object/scene plays a key role since it exhibits a significant variance in the same subcategory and subtle variance among different subcategories. Finding the subtle variance that fully characterizes the object/scene is not straightforward. To address this, we propose a novel context-aware attentional pooling (CAP) that effectively captures subtle changes via sub-pixel gradients, and learns to attend informative integral regions and their importance in discriminating different subcategories 
without requiring the bounding-box and/or distinguishable part annotations. We also introduce a novel feature encoding by considering the intrinsic consistency between the informativeness of the integral regions and their spatial structures to capture the semantic correlation among them. 
Our approach is simple yet extremely effective and can be easily applied on top of a standard classification backbone network. We evaluate our approach using six state-of-the-art (SotA) backbone networks and eight benchmark datasets. Our method significantly outperforms the SotA approaches on six datasets and is very competitive with the remaining two.
\end{abstract}
\section{Introduction}
Over recent years, there has been significant progress in the landscape of computer vision due to the adaptation and enhancement of a fast, scalable and end-to-end learning framework, the CNN \cite{lecun1998gradient}. This is not a recent invention, but we now see a profusion of CNN-based models achieving SotA results in visual recognition \cite{he2016deep, huang2017densely, zoph2018learning, sandler2018mobilenetv2}. 
The performance gain primarily comes from the model's ability to reason about image content by disentangling discriminative object pose and part information from texture and shape. 
Most discriminative features are often based on changes in global shape and appearance. They are often ill-suited to distinguish subordinate categories, 
involving subtle visual differences within various natural objects such as bird species \cite{wah2011caltech, van2015building}, flower categories \cite{nilsback2008automated}, dog breeds \cite{khosla2011novel}, pets \cite{parkhi2012cats} and man-made objects like aircraft types \cite{maji2013fine}, car models \cite{krause20133d}, etc. 
To address this, a global descriptor is essential which ensembles features from multiple local parts and their hierarchy so that the subtle changes can be discriminated as a misalignment of local parts or pattern. The descriptor should also be able to emphasize the importance of a part. 

There have been some excellent works on fine-grained visual recognition (FGVC) using weakly-supervised complementary parts \cite{ge2019weakly}, part attention \cite{liu2016fully}, object-part attention \cite{peng2017object}, multi-agent cooperative learning 
\cite{yang2018learning}, recurrent attention \cite{fu2017look}, and destruction and construction learning \cite{chen2019destruction}. All these approaches avoid part-level annotations and automatically discriminate local parts in an unsupervised/weakly-supervised manner. 
Many of them use a pre-trained object/parts detector and lack rich representation of regions/parts to capture the object-parts relationships better. To truly describe an image, we need to consider the image generation process from pixels to object to the scene in a more fine-grained way, not only to regulate the object/parts and their spatial arrangements but also defining their appearances using multiple partial descriptions as well as their importance in discriminating subtle changes. These partial descriptions should be rich and complementary to each other to provide a complete description of the object/image. In this work, we propose a simple yet compelling approach that embraces the above properties systematically to address the challenges associated with the FGVC. Thus, it can benefit to a wide variety of applications such as image captioning \cite{herdade2019image, huang2019attention,  li2019entangled}, expert-level image recognition \cite{valan2019automated, krause2016unreasonable}, and so on.

\noindent\textbf{Our work:} To describe objects in a conventional way as in CNNs as well as maintaining their visual appearance, we design a context-aware attentional pooling (CAP) to encode spatial arrangements and visual appearance of the parts effectively. The module takes the input as a convolutional feature from a base CNN and then \textit{learns to emphasize} the latent representation of 
multiple integral regions (varying coarseness) to describe hierarchies within objects and parts. Each region has an anchor in the feature map, and thus many regions have the same anchor due to the integral characteristics. These integral regions are then fed into a recurrent network (e.g. LSTM) to capture their spatial arrangements, and is inspired by the visual recognition literature, which suggests that humans do not focus their attention on an entire scene at once. 
Instead, they focus sequentially by attending different parts to extract relevant information \cite{zoran2020towards}. A vital characteristic of our CAP is that it generates a new feature map by focusing on a given region conditioned on all other regions and itself. Moreover, it efficiently captures subtle variations in each region by the sub-pixel gradients via bilinear pooling. The recurrent networks are mainly designed for sequence analysis/recognition. 
We aim to capture the subtle changes between integral regions and their spatial arrangements. Thus, we introduce a learnable pooling to emphasize the most-informative hidden states of the recurrent network, automatically. It learns to encode the spatial arrangement of the latent representation of integral regions and uses it to infer the fine-grained subcategories. 

\noindent\textbf{Our contributions:} Our main contributions can be summarized as: 1) an easy-to-use extension to SotA base CNNs by incorporating context-aware attention to achieve a considerable improvement in FGVC; 2) to discriminate the subtle changes in an object/scene, context-aware attention-guided rich representation of integral regions is proposed; 3) a learnable pooling is also introduced to automatically select the hidden states of a recurrent network to encode spatial arrangement and appearance features; 4) extensive analysis of the proposed model on eight FGVC datasets, obtaining SotA results; and 5) analysis of various base networks for the wider applicability of our CAP.
\begin{figure*}[t]
    \centering
    
    
    \includegraphics[width=0.33\textwidth]{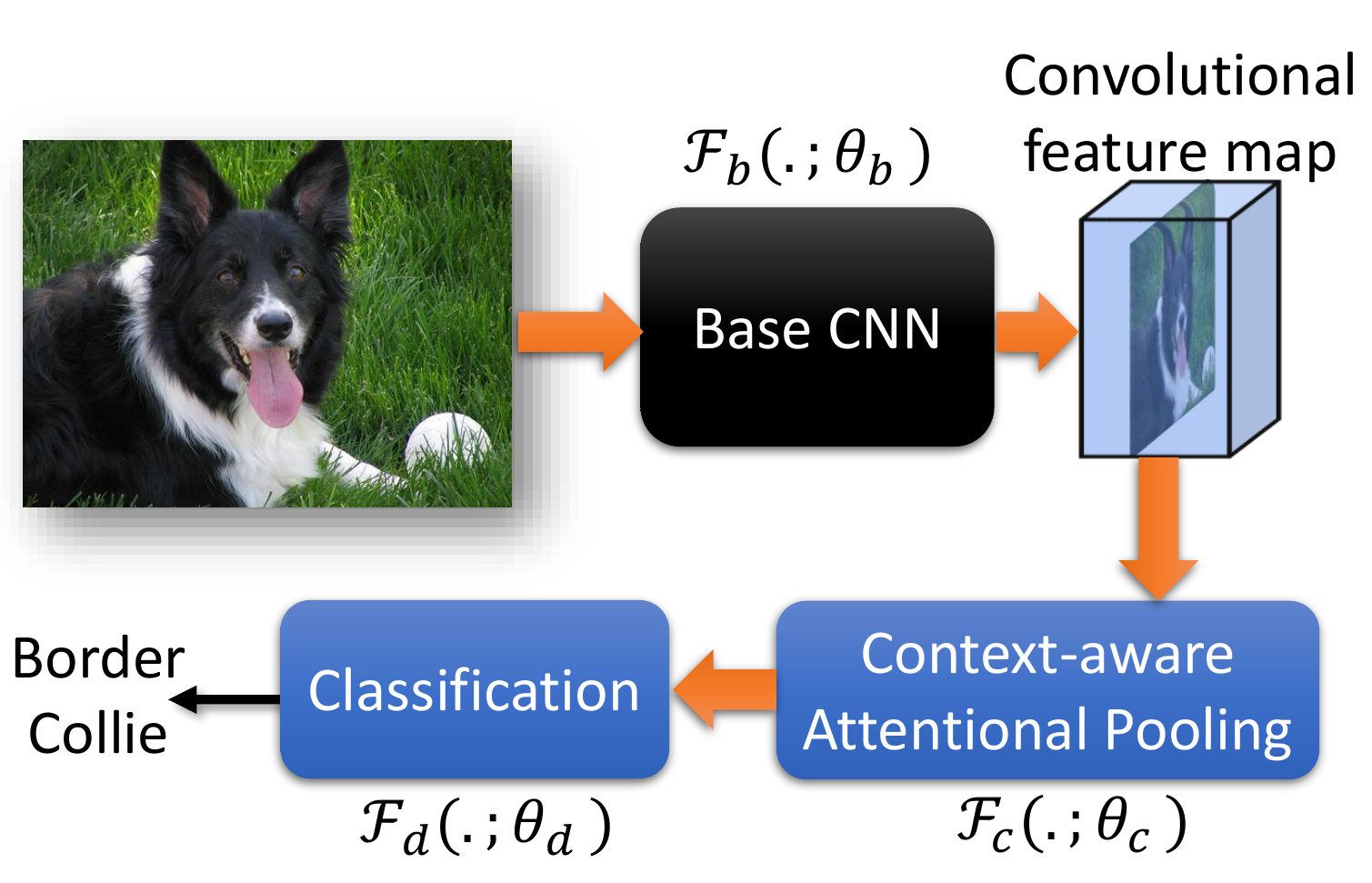} \hfill
    \includegraphics[width=0.65\textwidth]{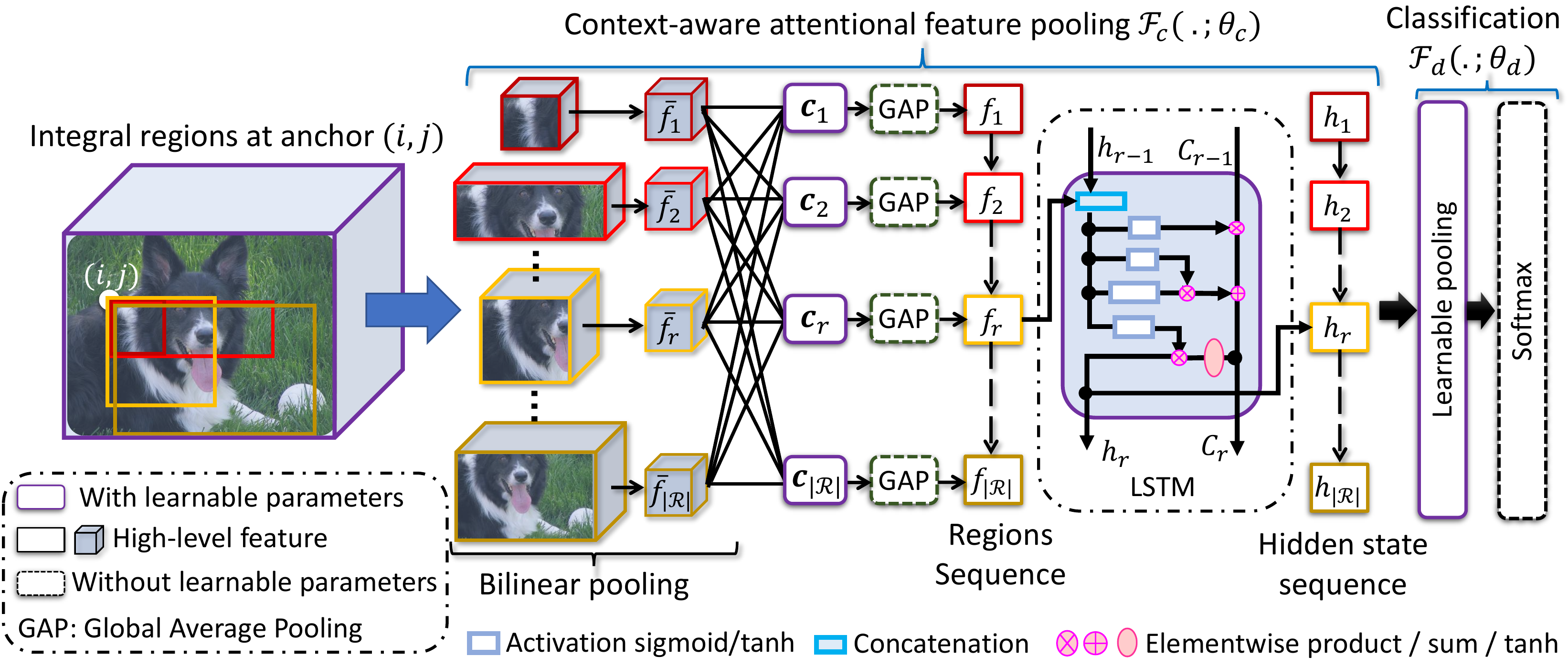}
    
    \caption{a) High-level illustration of our model (left). b) The detailed architecture of our novel CAP (right).}
    \label{fig:fig_app}
\end{figure*}
\begin{figure*}[h]
    \subfloat[Pixel-level relationships]{\includegraphics[width=0.14\textwidth] {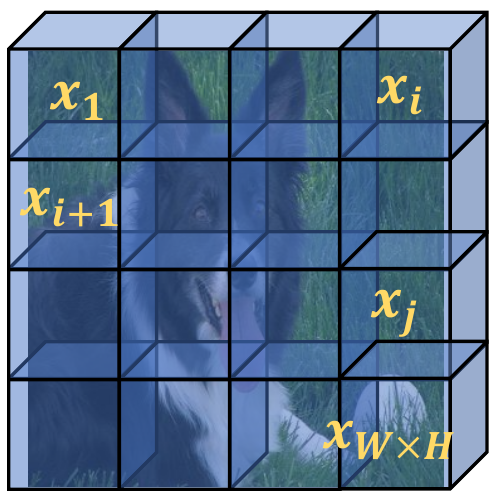}}\hfill
    \subfloat[Attention-focused contextual information from integral regions (surrounding context)]{\includegraphics[width=0.4\textwidth] {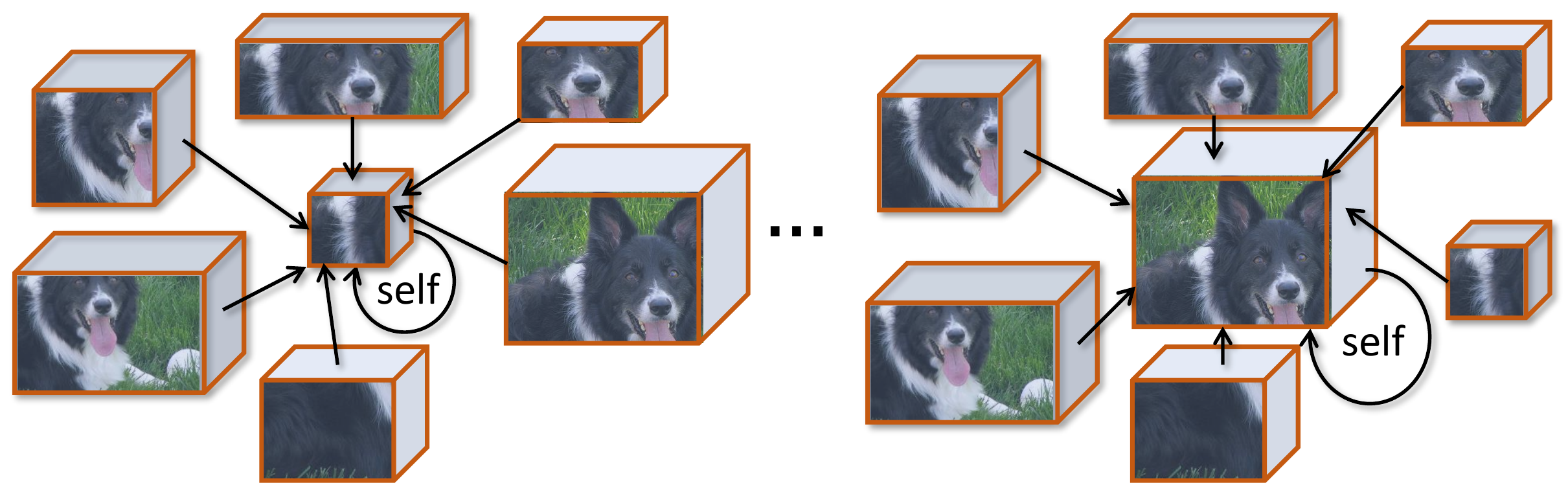}}\hfill
    \subfloat[Capturing spatial arrangements]{\includegraphics[width=0.17\textwidth] {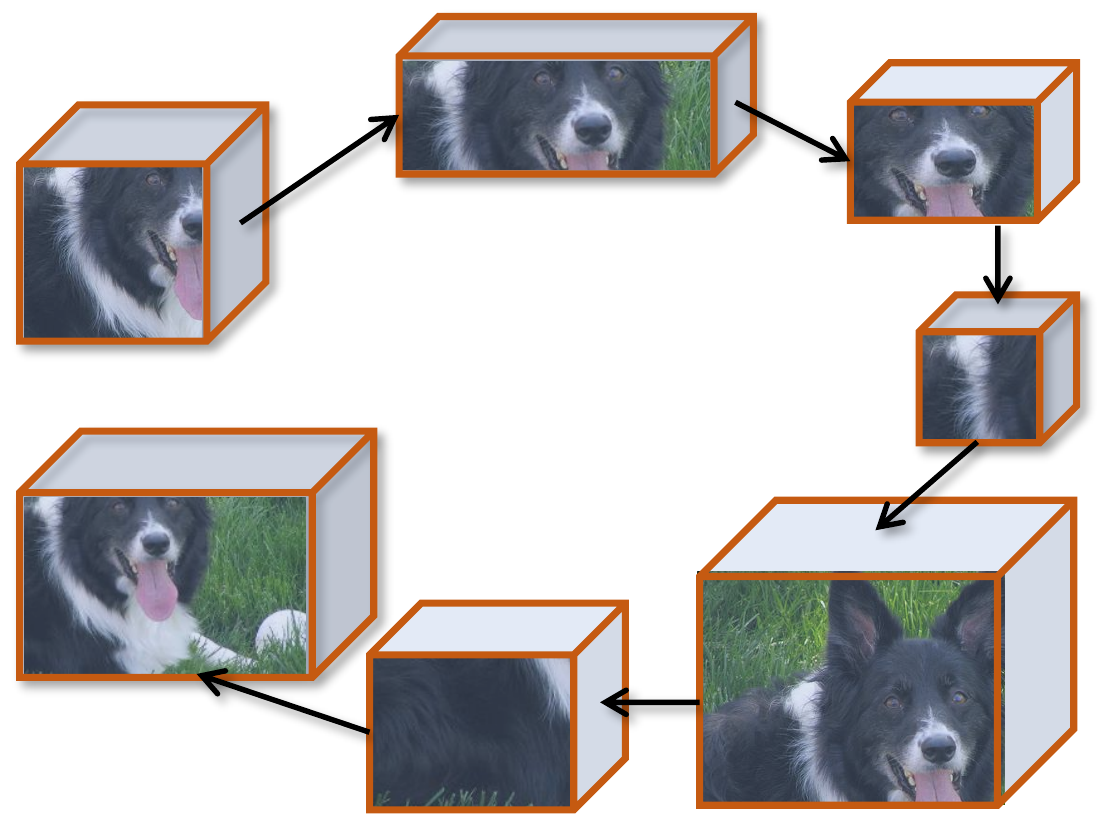}}\hfill
    \subfloat[Aggregated hidden state responses to class prediction]{\includegraphics[width=0.27\textwidth] {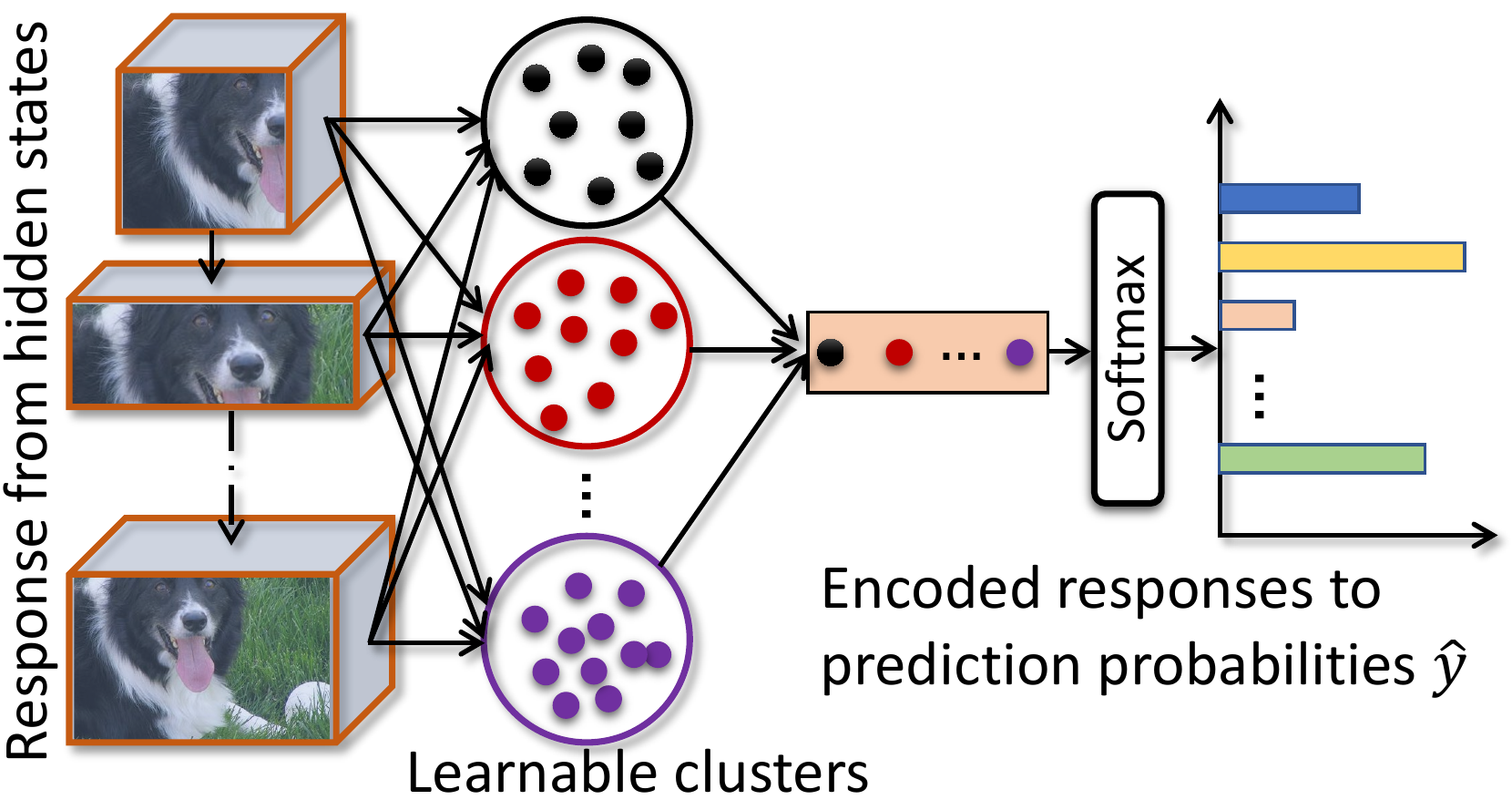}}
    \caption{a) Learning pixel-level relationships 
    from the convolutional feature map of size $W\times H\times C$. b) CAP using integral regions to capture both self and neighborhood contextual information. c) Encapsulating spatial structure of the integral regions using an LSTM. 
    d) Classification by learnable aggregation of hidden states of the LSTM. 
    }
    \label{fig:fig_struct}
\end{figure*}
\section{Related Work}\label{sec:rel_work}
%
\textbf{Unsupervised/weakly-supervised parts/regions based approaches:} Such methods learn a diverse collection of discriminative parts/regions to represent the complete description of an image. In \cite{chen2019destruction}, the global structure of an image is substantially changed by a random patch-shuffling mechanism to select informative regions. An adversarial loss is used to learn essential patches. 
In \cite{ge2019weakly}, Mask R-CNN and conditional random field are used for object detection and segmentation. A bidirectional LSTM is used to encode rich complementary information from selected part proposals for classification. A hierarchical bilinear pooling framework is presented in \cite{yu2018hierarchical} to learn the inter-layer part feature interaction from intermediate convolution layers. 
This pooling scheme enables inter-layer feature interaction and discriminative part feature learning in a mutually reinforced manner. In \cite{cai2017higher}, a higher-order integration of hierarchical convolutional features is described for representing parts semantic at different scales. A polynomial kernel-based predictor is defined for modelling part interaction using higher-order statistics of convolutional activations. 
A general pooling scheme is demonstrated in \cite{cui2017kernel} to represent higher-order and nonlinear feature interactions via compact and explicit feature mapping  using kernels. Our approach is complementary to these approaches by exploring integral regions and learns to attend these regions using a bilinear pooling that encodes partial information from multiple integral regions to a comprehensive feature vector for subordinate classification.

\noindent\textbf{Object and/or part-level attention-based approaches:} Recently, there has been significant progress to include attention mechanisms \cite{zhao2020exploring, leng2019context, bello2019attention, parmar2019stand} to boost image recognition accuracy. It is also explored in FGVC \cite{zheng2019looking, ji2018stacked, sun2018multi}. 
In \cite{zheng2019learning}, a part proposal network produces several local attention maps, and a part rectification network learns rich part hierarchies. 
Recurrent attention in \cite{fu2017look} learns crucial regions at multiple scales. 
The attended regions are cropped and scaled up with a higher resolution to compute rich features. Object-part attention model (OPAM) in \cite{peng2017object} incorporates object-level attention for object localization, and part-level attention for the vital parts selection. Both jointly learn multi-view and multi-scale features to improve performance. In \cite{liu2019bidirectional}, a bidirectional attention-recognition model (BARM) is proposed to optimize the region proposals via a feedback path from the recognition module to the part localization module. 
Similarly, in attention pyramid hierarchy \cite{ding2020weakly}, top-down and bottom-up attentions are integrated to learn both high-level semantic and low-level detailed feature representations. In \cite{lopez2020pay}, a modular feed-forward attention mechanism consisting of attention modules and attention gates is applied to learn low-level feature activations. 
Our novel paradigm is a step forward and takes inspiration from these approaches. It is advantageous over the existing methods as it uses a single network and the proposed attention mechanism learns to attend both appearance and shape information from a single-scale image in a hierarchical fashion by exploring integral regions. We further extend it by innovating the classification layer, where the subtle changes in integral regions are learned by focusing on the most informative hidden states of an LSTM. 
\section{Proposed Approach}\label{sec:proposed}
The overall pipeline of our model is shown in Fig. \ref{fig:fig_app}a. 
It takes an input image and provides output as a subordinate class label. To solve this, we are given $N$ images $I = \{I_n| n=1,\dots ,N\}$ and their respective fine-grained labels. 
The aim is to find a mapping function $\mathcal{F}$ that predicts $\hat{y}_n=\mathcal{F}(I_n)$, which matches the true label $y_n$. The ultimate goal is to learn $\mathcal{F}$ by minimizing a loss $L(y_n,\hat{y}_n)$ between the true and the predicted label. 
%
Our model consists of three elements (Fig. \ref{fig:fig_app}a): 1) a base CNN $\mathcal{F}_b(.;\theta_b)$, and our novel 2) CAP  
$\mathcal{F}_c(.;\theta_c)$ and 3) classification $\mathcal{F}_d(.;\theta_d)$ modules. We aim to learn the model's parameters $\theta=\{\theta_b,\theta_c, \theta_d\}$ 
via end-to-end training. We use the SotA CNN architecture as a base CNN $\mathcal{F}_b(.;\theta_b)$ and thus, we emphasize on the design and implementation of the rest two modules $\mathcal{F}_c(.;\theta_c)$ and $\mathcal{F}_d(.;\theta_d)$.
\subsection{Context-aware attentional pooling (CAP)} 
It takes the output of a base CNN as an input. Let us consider $\mathbf{x}=\mathcal{F}_b(I_n;\theta_b)$ to be the convolutional feature map as the output of the base network $F_b$ for input image $I_n$. 
The proposed CAP considers contextual information from pixel-level to small patches to large patches to image-level. The pixel refers to a spatial location in the convolutional feature map $\mathbf{x}$ of width $W$, height $H$ and channels $C$. 
The aim is to capture contextual information hierarchically to better model the subtle changes 
observed in FGVC tasks. Our attention mechanism learns to emphasize pixels, as well as regions of different sizes located in various parts of the image $I_n$. At pixel-level, we explicitly learn the relationships between pixels, i.e. $p(\mathbf{x}_i|\mathbf{x}_j;\theta_p)$, $\forall i$ $i\ne j$ and $1\le i,j \le W\times H$, even they are located far apart in $\mathbf{x}$. 
It signifies how much the model should attend the $i^{th}$ location when synthesizing the $j^{th}$ position in $\mathbf{x}$ (Fig. \ref{fig:fig_struct}a). To achieve this, we compute the attention map $\theta_p$ by revisiting the self-attention concept \cite{zhang2018self} where \textit{key} $k(\mathbf{x})=\mathbf{W}_{k}\mathbf{x}$, \textit{query} $q(\mathbf{x})=\mathbf{W}_{q}\mathbf{x}$ and \textit{value} $v(\mathbf{x})=\mathbf{W}_{v}\mathbf{x}$ in $\mathbf{x}$ are computed using separate $1\times 1$ convolutions. The attentional output feature map $\mathbf{o}$ is a dot-product of attention map $\theta_p$ and $\mathbf{x}$. 
$\theta_p =\{\mathbf{W}_{k}$, $\mathbf{W}_{q}, \mathbf{W}_{v}\}\in \theta_c$ is learned.

\noindent\textbf{Proposing integral regions:} To learn 
contextual information efficiently, we propose multiple integral regions with varying level of coarseness on the feature map $\mathbf{o}$. 
The level of coarseness is captured by different size of a rectangular region. Let us consider the smallest region $r(i,j,\Delta_x,\Delta_y)$ of width $\Delta_x$, height $\Delta_y$ and is located (top-left corner) at the $i^{th}$ column and $j^{th}$ row of $\mathbf{o}$. 
Using $r(i,j,\Delta_x,\Delta_y)$, we derive a set of regions by varying their widths and heights i.e. $R=\{r(i,j,m\Delta_x, n\Delta_y)\}$; $m,n=1,2,3,\dots$ and $i<i+m\Delta_x\le W$, $j<j+n\Delta_y\le H$. This is illustrated in Fig. \ref{fig:fig_app}b (left) for the given spatial location of ($i,j$). The goal is to generate the similar set of regions $R$ at various spatial locations ($0<i<W$, $0<j<H$) in $\mathbf{o}$. In this way, we generate a final set of regions $\mathcal{R}=\{R\}$ located at different places with different sizes and aspect ratios, 
as shown in Fig. \ref{fig:fig_app}b. The approach is a comprehensive context-aware representation to capture the rich contextual information characterizing subtle changes in images hierarchically.

\noindent\textbf{Bilinear pooling:} There are $|\mathcal{R}|$ regions with size 
varies from a minimum of $\Delta_x\times \Delta_y\times C$ to a maximum of $W\times H\times C$. The goal is to represent these variable size regions $(X\times Y\times C) \Rightarrow (w\times h\times C)$ with a fixed size feature vector. 
Thus, we use bilinear pooling, typically  bilinear interpolation to implement differentiable image transformations, which requires indexing operation. 
Let $T_\psi(\mathbf{y})$ be the coordinate transformation with parameters $\psi$ and $\mathbf{y}=(i,j) \in \mathbb{R}^{2}$ denotes a region coordinates at which the feature value is $\mathbf{R}(\mathbf{y})\in \mathbb{R}^C$. The transformed image $\mathbf{\tilde{R}}$ at the target coordinate $\mathbf{\tilde{y}}$ is: 
\begin{equation}
    \mathbf{\tilde{R}}(\mathbf{\tilde{y}}) = \sum_{\mathbf{y}}\mathbf{R}(T_\psi(\mathbf{y})) \text{ } K(\mathbf{\tilde{y}},T_\psi(\mathbf{y})),
    \label{eq:bilinear}
\end{equation}
where $\mathbf{R}(T_\psi(\mathbf{y}))$ is the image indexing operation and is non-differentiable; thus, the way gradients propagate through the network depends on the kernel $K(.,.)$. In bilinear interpolation, the kernel $K(\mathbf{y}_1,\mathbf{y}_2)=0$ when $\mathbf{y}_1$ and $\mathbf{y}_2$ are not direct neighbors. Therefore, the sub-pixel gradients (i.e. the feature value difference between neighboring locations in the original region) only flow through during propagation 
\cite{jiang2019linearized}. This is an inherent flaw in bilinear interpolation since the sub-pixel gradients will not associate to the large-scale changes which cannot be captured by the immediate neighborhood of a point. To overcome this, several variants \cite{jiang2019linearized, lin2017inverse} have been proposed. However, for our work, we exploit this flaw 
to capture the subtle changes in all regions 
via sub-pixel gradients. 
Note that the bilinear interpolation, although is not differentiable at all points due to the \textit{floor} and \textit{ceiling} functions, can backpropagate the error and is differentiable in most inputs as mentioned in the seminal work of Spatial Transform Networks \cite{jaderberg2015spatial}. We use bilinear kernel $K(.,.)$ in (\ref{eq:bilinear}) to pool fixed size features $\bar{f}_r$ ($w\times h\times C$) from all $r\in \mathcal{R}$. 

\noindent\textbf{Context-aware attention:} 
In this step, we capture the contextual information using our novel attention mechanism, which transforms $\bar{f}_r$ to a weighted version of itself and conditioned on the rest of the feature maps $\bar{f}_{r'}$ ($r,r'\in \mathcal{R}$). 
It enables our model to selectively focus on more relevant integral regions to generate holistic context information. 
The proposed context-aware attention 
takes a \textit{query} $\mathbf{q}(\bar{f}_r)$ and maps against a set of \textit{keys} $\mathbf{k}(\bar{f}_{r'})$ associated with the integral regions $r'$ in a given image, and then returns the output as a context vector $\mathbf{c}_r$ and is computed as:
\begin{equation}
\begin{split}
    \mathbf{c}_r = \sum_{r'=1}^{|\mathcal{R}|}&\alpha_{r,r'}\bar{f}_{r'}\text{ , } \alpha_{r,r'} = \text{softmax}\left(W_\alpha \beta_{r,r'} + b_\alpha\right)\\
    \beta_{r,r'} &= \text{tanh}\left(\mathbf{q}(\bar{f}_r) + \mathbf{k}(\bar{f}_{r'})+b_\beta\right) \\ 
    \mathbf{q}(\bar{f}_r) &= W_\beta \bar{f}_r \text{ and } \mathbf{k}(\bar{f}_{r'}) = W_{\beta^{'}} \bar{f}_{r'},  
\end{split}
\label{eq:attn}
\end{equation}
where weight matrices $W_\beta$ and $W_{\beta'}$ are for estimating the \textit{query} and \textit{key} from the respective feature maps $\bar{f}_r$ and $\bar{f}_{r'}$; $W_\alpha$ is 
their nonlinear combination; $b_\alpha$ and $b_\beta$ are the biases. 
These matrices and biases ($\{W_\beta, W_{\beta'}, W_\alpha, b_\alpha, b_\beta \}\in \theta_c$) are learnable parameters. 
The context-aware attention element $\alpha_{r,r'}$ captures the similarity between the feature maps $\bar{f}_{r}$ and $\bar{f}_{r'}$ of regions $r$ and $r'$, respectively. The attention-focused context vector $\mathbf{c}_r$ determines the \textit{strength} of $\bar{f}_r$ in focus \textit{conditioned on itself and its neighborhood context}. This applies to all integral regions $r$ (refer Fig. \ref{fig:fig_struct}b).

\noindent\textbf{Spatial structure encoding:} The context vectors $\mathbf{c}=\{\mathbf{c}_r|r=1\dots |\mathcal{R}|\}$ characterize the attention and saliency. 
To  include the structural information involving the spatial arrangements of regions (see Fig. \ref{fig:fig_app}b and \ref{fig:fig_struct}b), we represent $\mathbf{c}$ as a sequence of regions (Fig. \ref{fig:fig_struct}c) and adapt a recurrent network to capture the structural knowledge using its internal states, which is modeled via hidden units $h_r \in \mathbb{R}^n$. 
Thus, the internal state representing the region $r$ is updated as: $h_r=\mathcal{F}_h(h_{r-1}, f_r;\theta_h)$, where $\mathcal{F}_h$ is a nonlinear function with learnable parameter $\theta_h$. We use a fully-gated LSTM as $\mathcal{F}_h$ \cite{hochreiter1997long} which is capable of learning long-term dependencies. 
The parameter $\theta_h \in \theta_c$ consists of weight matrices and biases linking input, forget and output gates, and cell states of $\mathcal{F}_h$. For simplicity, we omitted equations to compute these parameters and refer interested readers to \cite{hochreiter1997long} for further details. To improve the generalizability and lower the computational complexity of our CAP, the context feature $f_r$ is extracted from the context vector $\mathbf{c}_r$ via global average pooling (GAP). 
This results in the reduction of feature map size from ($w\times h\times C)$ to ($1\times C$). 
The sequence of hidden states $h=(h_1,h_2, \dots, h_r, \dots, h_{|\mathcal{R}|})$ corresponding to the input sequence of context feature $f=(f_1,f_2, \dots, f_r, \dots, f_{|\mathcal{R}|})$ (see Fig. \ref{fig:fig_app}b) is used by our classification module $\mathcal{F}_d(.;\theta_d)$.
\subsection{Classification}\label{sec:decision}
To further guide our model to discriminate the subtle changes, we propose a learnable pooling approach (Fig. \ref{fig:fig_struct}c), which aggregates information by grouping similar responses from the hidden states $h_r$. 
It is inspired by the existing feature encoding approach, such as 
NetVLAD \cite{arandjelovic2016netvlad}. 
We adapt this differentiable clustering approach for the soft assignment of the responses from hidden states $h_r$ 
to cluster $\kappa$ and their contribution to the VLAD encoding. 
\begin{equation}
\begin{split}
    \gamma_{\kappa}(h_r)&=\frac{e^{W_\kappa^{T}h_r+b_\kappa}}{\sum_{i=1}^\mathcal{K}e^{W_i^{T}h_r+b_i}}\\
    N_{v}(o,\kappa)&=\sum_{r=1}^{|\mathcal{R}|} \gamma_{\kappa}(h_r)h_r(o) \text{, }\hat{y}=\text{softmax}(W_{N}N_v)
\end{split}
\end{equation}
where $W_i$ and $b_i$ are learnable clusters' weights and biases. $T$ signifies transpose. The term $\gamma_{\kappa}(h_r)$ refers to the soft assignment of $h_r$ to cluster $\kappa$, and $N_{v}$ is the encoded responses of hidden states from all the regions $r\in \mathcal{R}$. In the original implementation of VLAD, the weighted sum of the residuals is used i.e. $\sum_{r=1}^{|\mathcal{R}|} \gamma_{\kappa}(h_r)\left(h_r(o) - \hat{c}_\kappa(o)\right)$ in which $\hat{c}_\kappa$ is the $\kappa^{th}$ cluster center and $o\in h_r$ is one of the elements in the hidden state response. We adapt the simplified version that averages the actual responses
instead of residuals \cite{miech2017learnable}, which requires fewer parameters and computing operations. 
The encoded response is mapped into prediction probability $\hat{y}$ 
by using a learnable weight $W_N$ and \texttt{softmax}. 
The learnable parameter for the classification module $\mathcal{F}_d$ is $\theta_d = \{W_i, b_i, W_N\}$. 
\begin{table*}[t]
\centering
\begin{small}
  \begin{tabular}{lcccll}
    \toprule
    Dataset &  \#Train / \#Test & \#Classes & Our & Past Best (primary) &Past Best (primary + secondary) \\
    \midrule

Aircraft & 6,667 / 3,333 &100 &\textbf{94.9} &93.0 \cite{chen2019destruction} &  92.9 \cite{yu2018deep}\\

Food-101 &75,750 / 25,250 &101 &\textbf{98.6} &93.0 \cite{huang2019gpipe} & 90.4 \cite{cui2018large} \\

Stanford Cars & 8,144 / 8,041 &196 &\textbf{95.7}  &94.6 \cite{huang2019gpipe} & 94.8 \cite{cubuk2019autoaugment}\\

Stanford Dogs & 12,000 / 8,580 &120 &96.1 &93.9 \cite{ge2019weakly} & \textbf{97.1} \cite{ge2019weakly} \\

CUB-200 & 5,994 / 5,794 &200 &\textbf{91.8} &90.3 \cite{ge2019weakly} & 90.4 \cite{ge2019weakly} \\

Oxford Flower & 2,040 / 6,149 &102 &\textbf{97.7} & 96.4 \cite{xie2016interactive} & \textbf{97.7} \cite{chang2020devil} \\

Oxford Pets & 3,680 / 3,669 &37  &\textbf{97.3} & 95.9 \cite{huang2019gpipe} &  93.8 \cite{peng2017object} \\

NABirds &23,929 / 24,633 &555 &\textbf{91.0}  & 86.4 \cite{luo2019cross} & 87.9 \cite{cui2018large} \\
\bottomrule
\end{tabular}
\end{small}
 \caption{Dataset statistics and performance evaluation. FGVC accuracy (\%) of our model and the previous best using only the primary dataset. The last column involves the transfer/joint learning strategy consisting of more than one dataset.}
 \label{table:overall_acc}
\end{table*}
\begin{table*}[t]
  \centering
  \begin{small}
  \begin{tabular}{lc|lc|lc}
    \toprule
      \multicolumn{2}{c|}{\textbf{Aircraft}}&
      \multicolumn{2}{c|}{\textbf{Food-101}}&
      \multicolumn{2}{c}{\textbf{Stanford Cars}}\\
      {Method} & {ACC} & {Method} & {ACC} & {Method} & {ACC} \\
      \midrule
   DFL (Wang et al. 2018) & 92.0    &    WISeR (Martinel et al., 2018)  & 90.3 &
   BARM \cite{liu2019bidirectional} & 94.3  \\  
 
    BARM \cite{liu2019bidirectional} & 92.5   & DSTL$^*$ \cite{cui2018large}   & 90.4  &   
    MC$_{Loss}^*$ \cite{chang2020devil} & 94.4 \\           
    GPipe \cite{huang2019gpipe} & 92.7  & MSMVFA \cite{JiangMLL20}  & 90.6 &  
    DCL\cite{chen2019destruction} & 94.5\\
   
    MC$_{Loss}^*$ \cite{chang2020devil} & 92.9 & JDNet$^*$ \cite{zhao2020jdnet}  & 91.2 & GPipe \cite{huang2019gpipe} & 94.6 \\ 
    
    DCL \cite{chen2019destruction} & 93.0 & GPipe \cite{huang2019gpipe}  & 93.0 & AutoAug$^*$ \cite{cubuk2019autoaugment} & 94.8 \\
         \hline
    \textbf{Proposed} & \textbf{94.9}  &\textbf{Proposed} & \textbf{98.6} &\textbf{Proposed} & \textbf{95.7} \\
    \hline
    \hline
      \multicolumn{2}{c|}{\textbf{CUB-200}} &
      \multicolumn{2}{c|}{\textbf{Oxford-IIIT Pets}} &
      \multicolumn{2}{c}{\textbf{NABirds}} \\
      \midrule
   iSQRT \cite{li2018towards} & 88.7 & 
   NAC \cite{simon2015neural} & 91.6  & 
   T-Loss \cite{taha2020boosting} &	79.6  \\  
 
    DSTL$^*$ \cite{cui2018large} & 89.3   &  
    TL-Attn$^*$ \cite{xiao2015application} & 92.5 &   
    PC-CNN \cite {dubey2018pairwise}   & 82.8 \\  
DAN \cite{hu2019see} & 89.4   &
InterAct \cite{xie2016interactive} & 93.5 
& MaxEnt$^*$ \cite{dubey2018maximum} &	83.0    \\
   
BARM \cite{liu2019bidirectional}& 89.5 &
OPAM$^*$ \cite{peng2017object} & 93.8 & Cross-X \cite {luo2019cross}   & 86.4  \\ 
    
CPM$^*$ \cite{ge2019weakly} & 90.4 & 
GPipe \cite{huang2019gpipe} & 95.9 &  
DSTL$^*$  \cite{cui2018large}  &87.9\\
         \hline
    \textbf{Proposed} & \textbf{91.8}  &\textbf{Proposed} & \textbf{97.3} & \textbf{Proposed} & \textbf{91.0} \\
    \bottomrule  
  \end{tabular}
  \end{small}
  \caption{Accuracy (\%) comparison with the recent top-five SotA approaches. Methods marked with * involve transfer/joint learning strategy for objects/patches/regions consisting more than one dataset (primary and secondary). Please refer to the supplementary page in the end for the results of Stanford Dogs and Oxford Flowers.}
  \label{table:sota}
\end{table*}
\section{Experiments and Discussion}
\label{se:data_exp}
We comprehensively evaluate our model on widely used eight benchmark FGVC datasets: Aircraft \cite{maji2013fine}, Food-101 \cite{BossardGG14}, Stanford Cars \cite{krause20133d}, Stanford Dogs \cite{khosla2011novel}, Caltech Birds (CUB-200) \cite{wah2011caltech}, Oxford Flower \cite{nilsback2008automated}, Oxford-IIIT Pets \cite{parkhi2012cats}, and NABirds \cite{van2015building}. We do not use any bounding box/part annotation. 
Thus, we do not compare with methods which rely on these. Statistics of datasets and their train/test splits are shown in Table \ref{table:overall_acc}. We use the top-1 accuracy (\%) for evaluation. 

\noindent\textbf{Experimental settings:} In all our experiments, we resize images to size $256\times256$, apply data augmentation techniques of random rotation ($\pm 15$ degrees), random scaling ($1\pm 0.15$) and then random cropping to select the final size of $224\times224$ from $256\times256$. The last Conv layer of the base CNN (e.g. $7\times 7$ pixels) is increased to $42\times42$ by using an upsampling layer (as in GAN) and then fed into our CAP (Fig. \ref{fig:fig_app}a) to pool features from multiple 
integral regions $\mathcal{R}$. We fix bilinear pooling size of $w=h=7$ for each region with minimum width $\Delta_x=7$ and height $\Delta_y=7$. We use spatial location gap of 7 pixels between consecutive anchors to generate $|\mathcal{R}|=27$ integral regions. 
This is decided experimentally by considering the trade-off between accuracy and computational complexity. 
We set the cluster size to 32 in our learnable pooling approach. 
We apply Stochastic Gradient Descent (SGD) optimizer to optimize the categorical cross-entropy loss function. The SGD is initialized with a momentum of 0.99, and initial learning rate 1e-4, which is multiplied by 0.1 after every 50 epochs. The model is trained for 150 epochs using an NVIDIA Titan V GPU (12 GB). 
We use Keras+Tensorflow to implement our algorithm.

\noindent\textbf{Quantitative results and comparison to the SotA approaches:} 
Overall, our model outperforms the SotA approaches by a clear margin on all datasets except the Stanford Dogs \cite{khosla2011novel} and Oxford Flowers \cite{nilsback2008automated} (Table \ref{table:overall_acc}). 
In Table \ref{table:overall_acc}, we compare our performances with the two previous best (last two columns). One uses only the target dataset (primary) for training and evaluation (past best) and is the case in our model. The other (last column) uses primary and additional secondary (e.g. ImageNet, COCO, iNat, etc.) datasets for joint/transfer learning of objects/patches/regions during training. 
It is worth mentioning that we use only the primary datasets and our performance in most datasets is significantly better than those uses additional datasets. 
This demonstrates the benefit of the proposed approach for discriminating fine-grained changes in recognizing subordinate categories. Moreover, we use only one network for end-to-end training, and our novel CAP and classification layers are added on top of a base CNN. Therefore, the major computations are associated with the base CNNs. 

Using our model, the two highest gains are 5.6\% and 3.1\% in the respective Food-101 \cite{BossardGG14} and NABirds \cite{van2015building} datasets. In Dogs, our method (96.1\%) is significantly better than the best SotA approach (93.9\%) \cite{ge2019weakly} using only primary data. However, their accuracy increases to 97.1\% when joint fine-tuning with selected ImageNet images are used. Similarly, in Flowers, our accuracy (97.7\%) is the same as in \cite{chang2020devil} which uses both primary and secondary datasets, and we achieve an improvement of 1.3\% compared to the best SotA approach in \cite{xie2016interactive} using only primary data. We also compare our model's accuracy with the top-five SotA approaches on each dataset in Table \ref{table:sota}. 
Our accuracy is significantly higher than SotA methods using primary data in all six datasets in Table \ref{table:sota} and two in supplementary (provided in the end). Furthermore, it is also considerably higher than SotA methods, which use both primary and secondary data in six datasets (Aircraft, Food-101, Cars, CUB-200, Pets and NABirds). This clearly proves our model's powerful ability to discriminate subtle changes in recognizing subordinate categories without requiring additional datasets and/or subnetworks and thus, has an advantage of easy implementation and a little computational overhead in solving FGVC.
\begin{table}[t]
  \begin{small}
  \begin{tabular}{lcccccc}
    \toprule
      {Base CNN} & {Plane} & {Cars} & {Dogs} & {CUB} & {Flowers} & {Pets}\\
      \midrule
   ResNet-50 & \textbf{94.9}    &94.9 &95.8 &90.9  & 97.5 & 96.7\\
   Incep. V3 & 94.8  &94.8 &95.7 &91.4 &97.6 &96.2\\
   Xception & 94.1 &\textbf{95.7} &\textbf{96.1}  & \textbf{91.8} &\textbf{97.7}  &97.0\\
    DenseNet & 94.6 &93.6 &95.5 &91.6 &97.6 &96.9 \\
    NASNet-M & 93.8  &93.7 &96.0 &89.7 &\textbf{97.7} &\textbf{97.3} \\
    Mob-NetV2 & 94.4 &94.0 &95.9 &89.2 &97.4 &96.4 \\
    \bottomrule
  \end{tabular}
  \end{small}
  \caption{Our model's accuracy (\%) with different SotA base CNN architectures. Previous best accuracies for these results are; Aircraft: 93.0 \cite {chen2019destruction}, Cars: 94.6 \cite{huang2019gpipe}, Dogs: 93.9 \cite{ge2019weakly}, CUB: 90.3 \cite {ge2019weakly}, Flowers: 96.4 \cite{xie2016interactive}, and Pets: 95.9 \cite{huang2019gpipe}. The result of the Birds dataset is included in the supplementary document in the end. 
  }
  \label{table:abl_1}
\end{table}

\noindent\textbf{Ablation study:} We compare the performance of our approach using the benchmarked base CNN architectures such as ResNet-50 \cite{he2016deep}, Inception-V3 \cite{szegedy2016rethinking}, Xception \cite{chollet2017xception} and DenseNet121 \cite{huang2017densely}, as well as SotA lightweight architectures such as NASNetMobile \cite{zoph2018learning} and MobileNetV2 \cite{sandler2018mobilenetv2}. The performance is shown in Table \ref{table:abl_1}. In all datasets, both standard and lightweight architectures have performed exceptionally well when our proposed CAP and classification modules are incorporated. Even our model outperforms the previous best (primary data) for both standard and lightweight base CNNs except in Cars and CUB-200 datasets in which our model with standard base CNNs exceed the previous best. Our results in Table \ref{table:overall_acc} \& \ref{table:sota} are the best accuracy among these backbones. Nevertheless, the accuracy of our model using any standard backbones (ResNet50 / Inception V3 / Xception; Table \ref{table:abl_1}) is better than the SotA. In Flowers and Pets datasets, the lightweight NASNetMobile is the best performer, and the MobileNetV2 is not far behind (Table \ref{table:abl_1}). This could be linked to the dataset size since these two are of smallest in comparison to the rest (Table \ref{table:overall_acc}). However, in other datasets (e.g. Aircraft, Cars and Dogs), there is a little gap in performance between standard and lightweight CNNs. These lightweight CNNs involve significantly less computational costs, and by adding our modules, the performance can be as competitive as the standard CNNs. This proves the importance of our modules in enhancing performance and its broader applicability.

We have also evaluated the above base CNNs (B), and the influence of our novel CAP (+C) and the classification module (+E) in the recognition accuracy on Aircraft, Cars and Pets datasets (more in the supplementary in the end). The results are shown in Table \ref{table:abl_2}. It is evident that the accuracy improves as we add our modules to the base networks, i.e., (B+C+E) $>$ (B+C) $>$ (B+E) $>$ B, resulting in the largest gain contributed by our novel CAP (B+C). This signifies the impact of our CAP. In B+C, the minimum gain is 7.2\%, 5.7\% and 5.1\% on the respective Aircraft, Cars and Pets datasets for the Inception-V3 as a base CNN. Similarly, the highest gain is 12.5\% and 11.3\% in Aircraft and Cars, respectively. These two datasets are relatively larger than the Pets (Table \ref{table:overall_acc}) in which the highest gain (7.9\%) is achieved by using ResNet-50 as a base CNN. We also observe that there is a significant gap in baseline accuracy between lightweight and standard base CNNs in larger (Aircraft and Cars) datasets. These gaps are considerably reduced when our CAP is added. There is a further increase in accuracy when we add the classification module (B+C+E). 
This justifies the inclusion of our novel encoding by grouping hidden responses using residual-less NetVLAD and then infer class probability using learnable pooling from these encoded responses. For base CNNs, we use the standard \textit{transfer learning} by 
fine-tuning it on the target dataset using the same data augmentation and hyper-parameters. 
For our models, we use pre-trained weights for faster convergence. We experimentally found that the random initialization takes nearly double iterations to converge (similar accuracy) than the pre-trained weights. A similar observation is shown in \cite{he2019rethinking}.
\begin{table*}[t]
  \centering
  \begin{small}
  \begin{tabular}{l|cccc|cccc|cccc|l|l}
    \toprule
    &\multicolumn{4}{c|}{Aircraft/Planes}  &
      \multicolumn{4}{c|}{Stanford Cars}  &
      \multicolumn{4}{c|}{Oxford-IIIT Pets} &Param &Time\\
      {Base CNN} & {Base} & {B+C} & {B+E} & {B+C+E} & {Base} & {B+C} & {B+E} & {B+C+E} & {Base} & {B+C} & {B+E} & {B+C+E} &(M) &ms \\
      \midrule
   ResNet-50 & 79.7 &88.8 &81.1 &94.9  &84.7 &91.5 &85.7 &94.9 &86.8 &94.7 &86.3 &96.7 &36.9 &4.1 \\ 
   Incep. V3 & 82.4 &89.6 &83.3 &94.8 &85.7 &91.4 &85.7 &94.8 &90.2 &95.3 &92.4 &96.2 &35.1 &3.8\\  
   Xception  &79.5 &89.5 &89.3 & 94.1 &84.8 &91.6 &89.1 &95.7 &91.0 &96.2  &96.0 &97.0 &34.2 &4.2\\ 
    NASNet-M &77.1 &89.6 &80.4 & 93.8 &80.4 &91.7 &82.7 &93.7  &89.9 &95.6 &94.9 &97.3 &9.7 &3.5\\  
    \bottomrule
  \end{tabular}
  \end{small}
  \caption{Performance (accuracy in \%) of our model with the addition of our novel CAP (+C) and classification (+E) module to various SotA base (B) CNNs. The observed accuracy trend is (B+C+E) $>$ (B+C) $>$ (B+E) $>$ B for all base CNNs. Final model's (B+C+E) trainable parameters (Param) are given in million (M) and the respective per-frame inference time in millisecond (ms).}
  \label{table:abl_2}
\end{table*}
\begin{figure*}[!h]
    %
    \subfloat[Base CNN]{\includegraphics[width=0.24\textwidth] {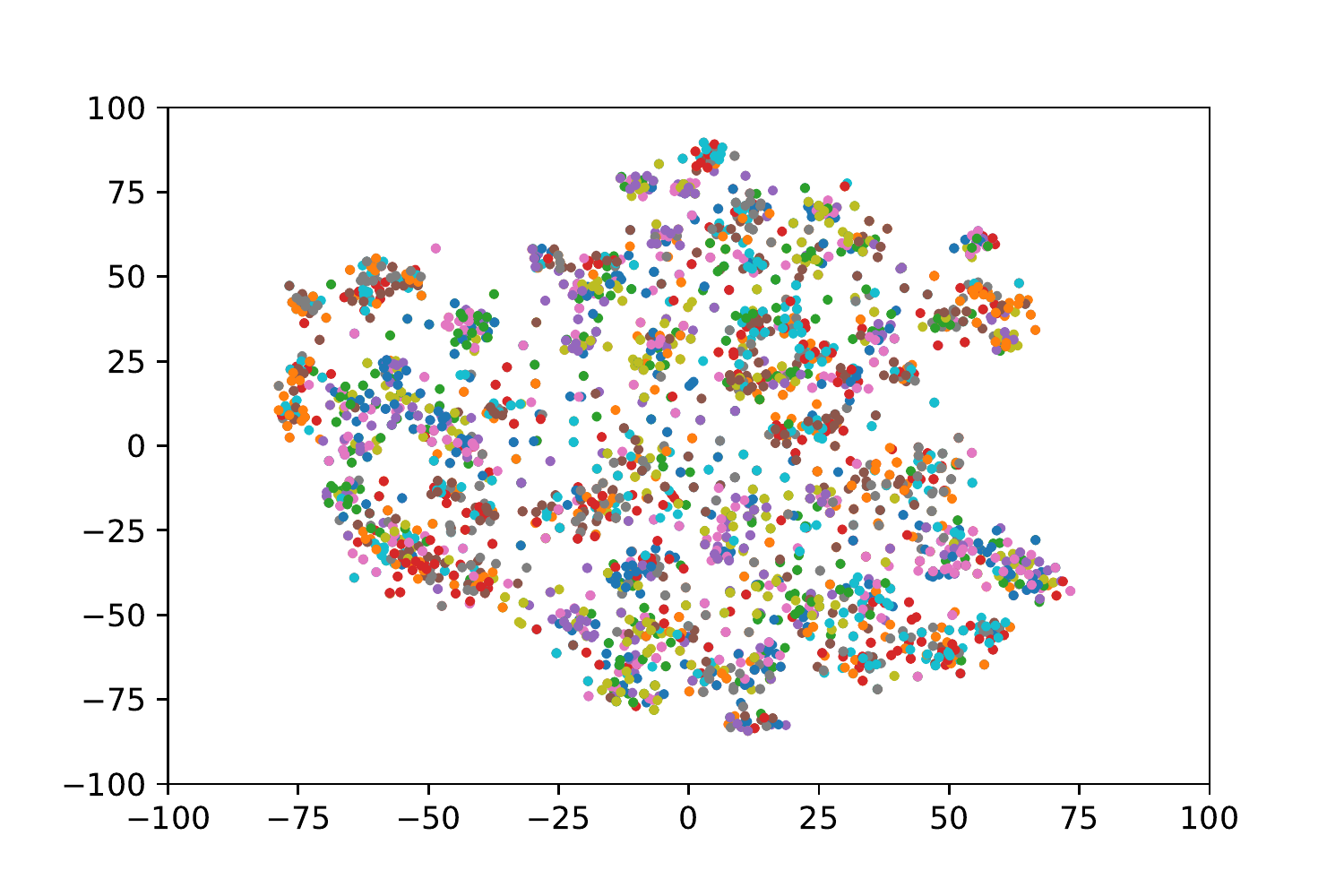}}\hfill
    \subfloat[Impact on base CNN]{\includegraphics[width=0.24\textwidth] {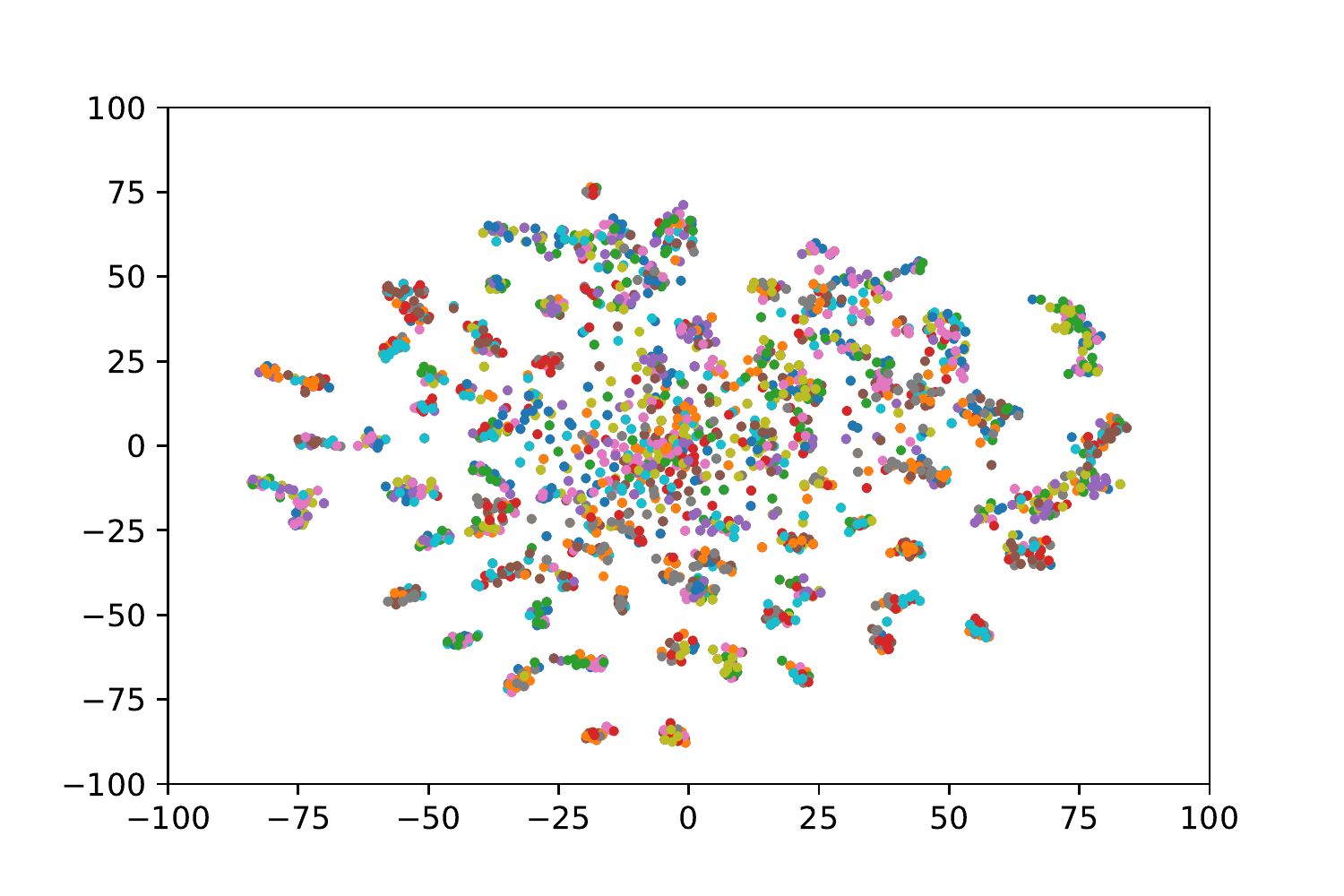}}\hfill
    \subfloat[CAP]{\includegraphics[width=0.24\textwidth] {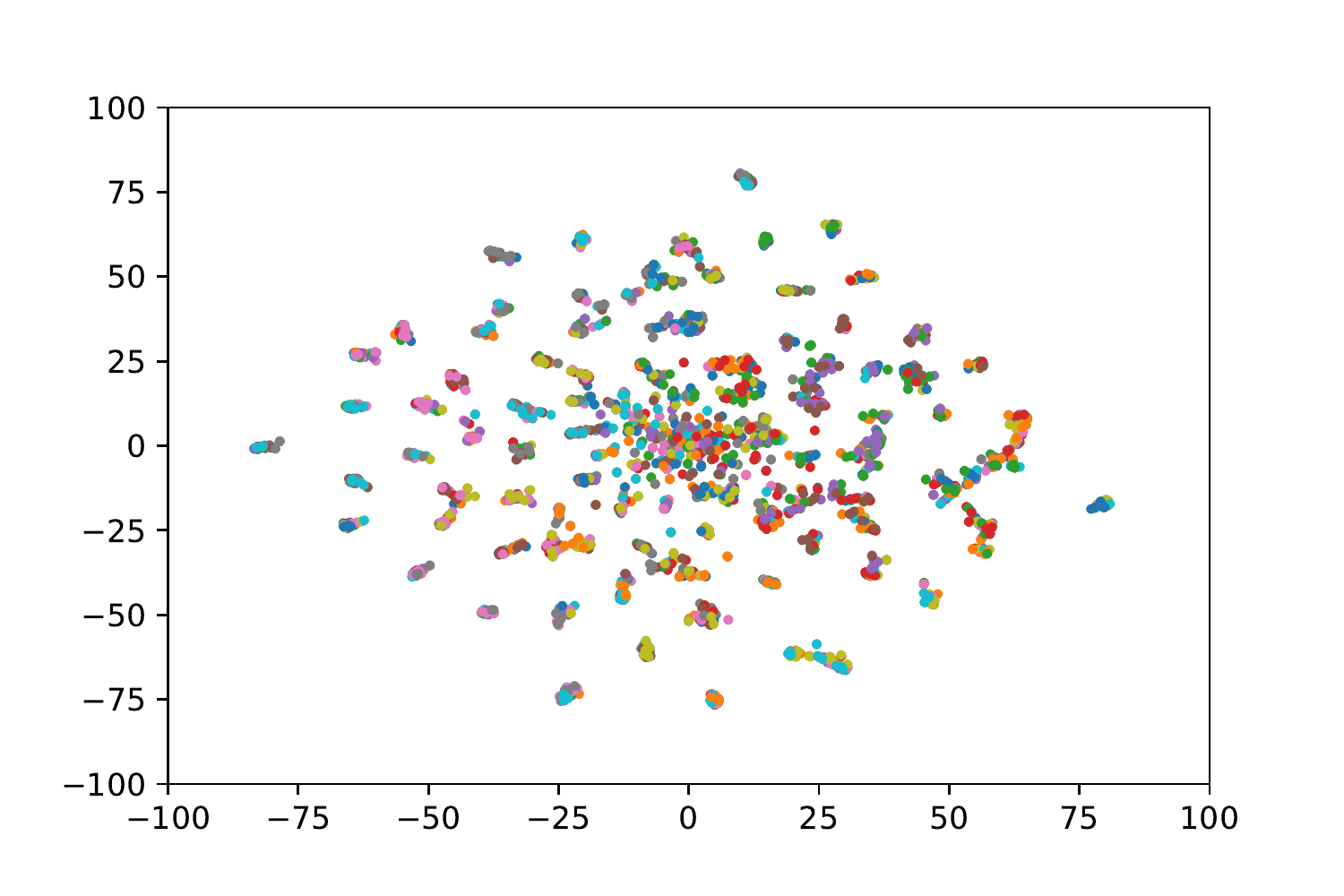}}\hfill
    \subfloat[CAP + Encoding]{\includegraphics[width=0.24\textwidth] {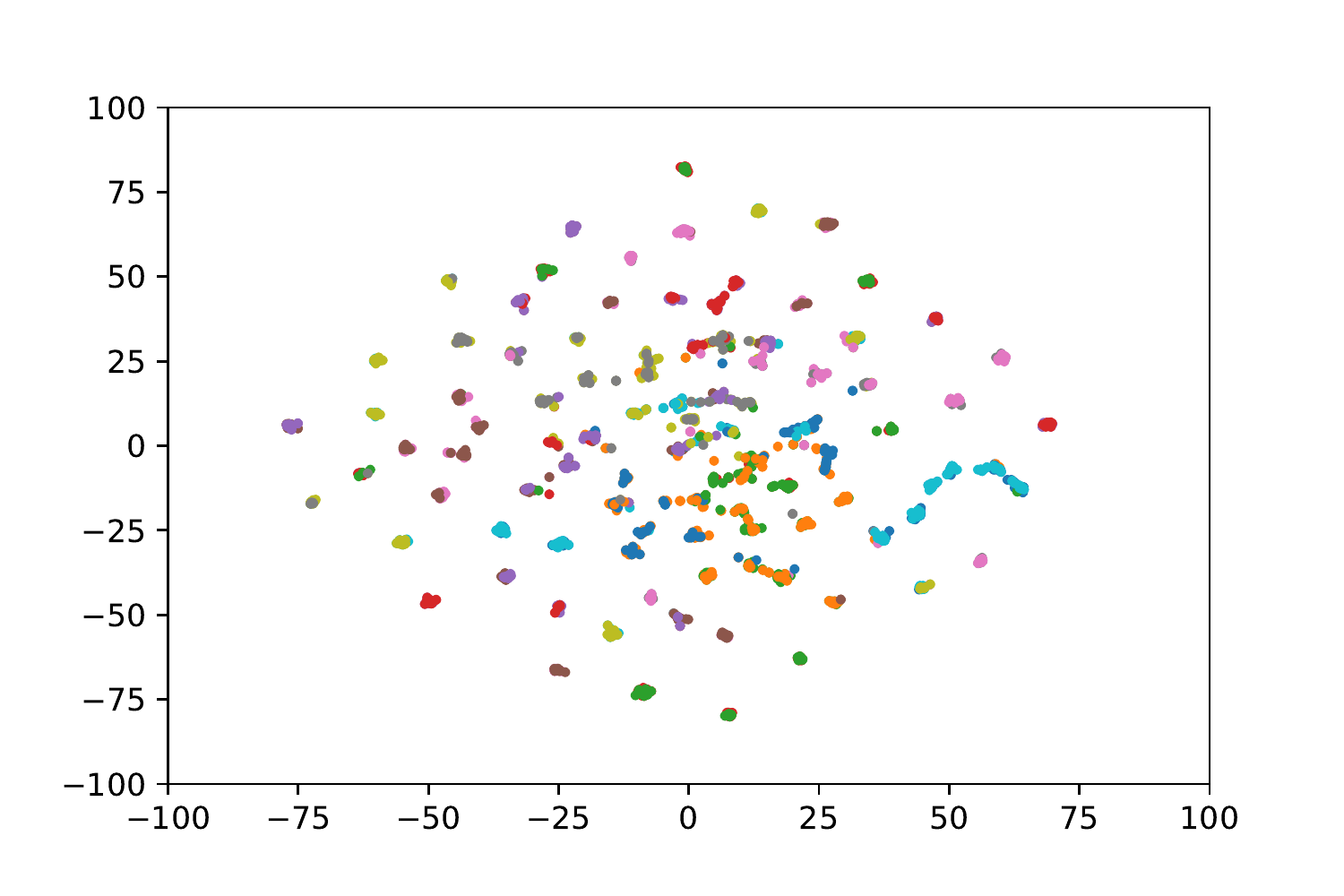}}\\
    \subfloat[$\alpha_{r,r'}$ for class 1]{\includegraphics[height=0.1\textwidth,width=0.155\textwidth] {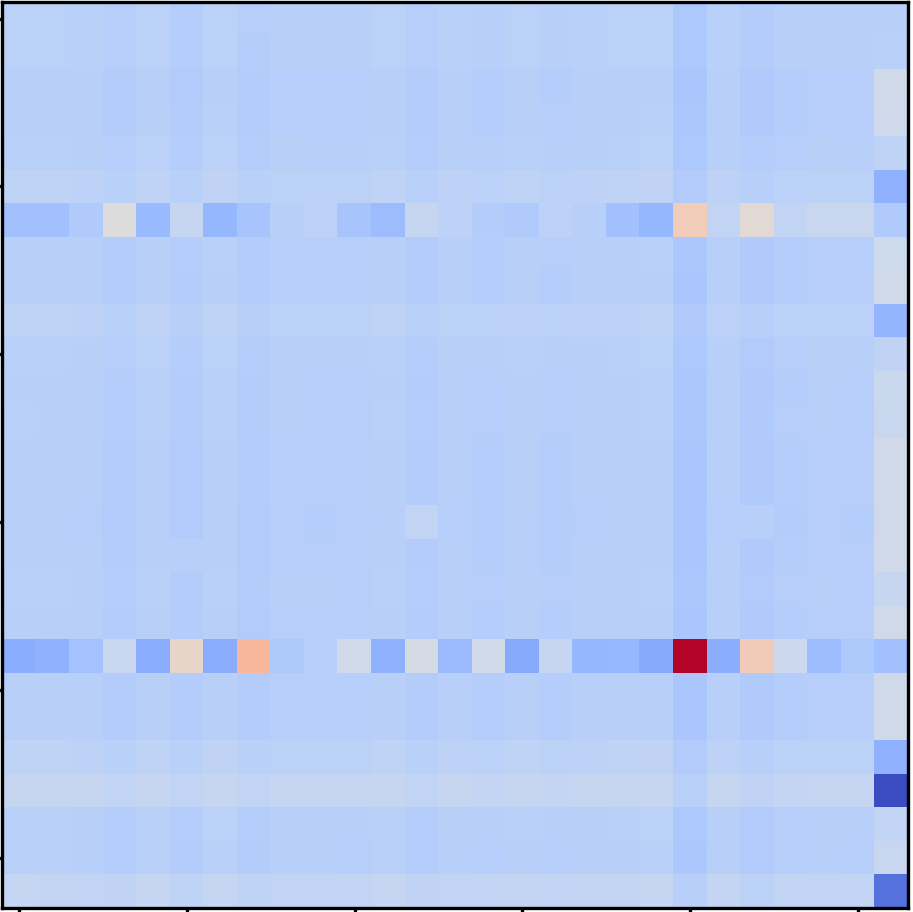}}\hfill
    \subfloat[$\alpha_{r,r'}$ for class 2]{\includegraphics[height=0.1\textwidth,width=0.155\textwidth] {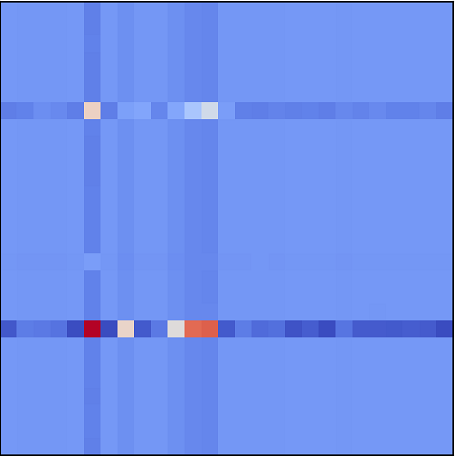}}\hfill
    \subfloat[$\mathbf{c}_r$ of region 1]{\includegraphics[height=0.1\textwidth,width=0.155\textwidth] {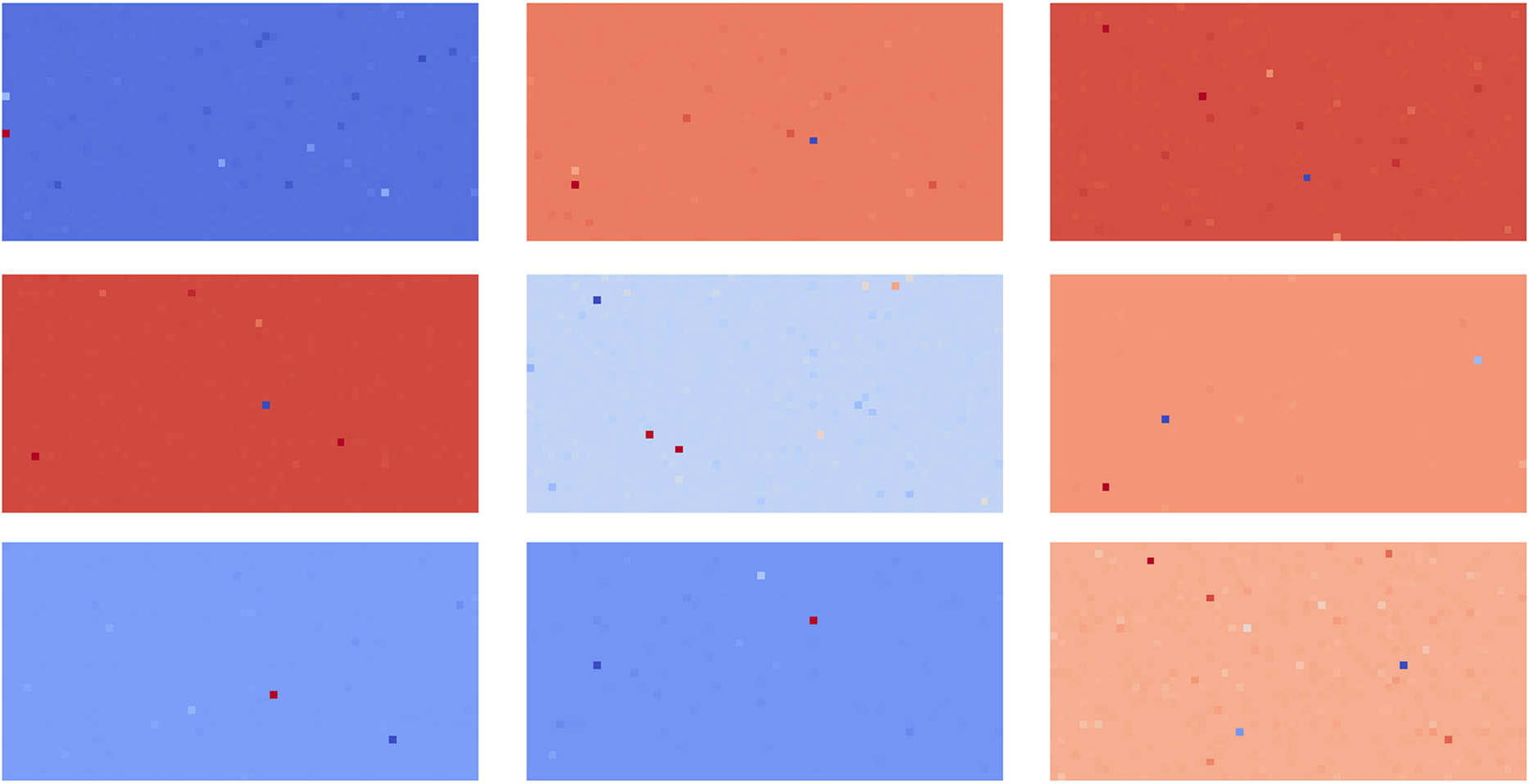}}\hfill
    \subfloat[$\mathbf{c}_r$ of region 20]{\includegraphics[height=0.1\textwidth,width=0.155\textwidth] {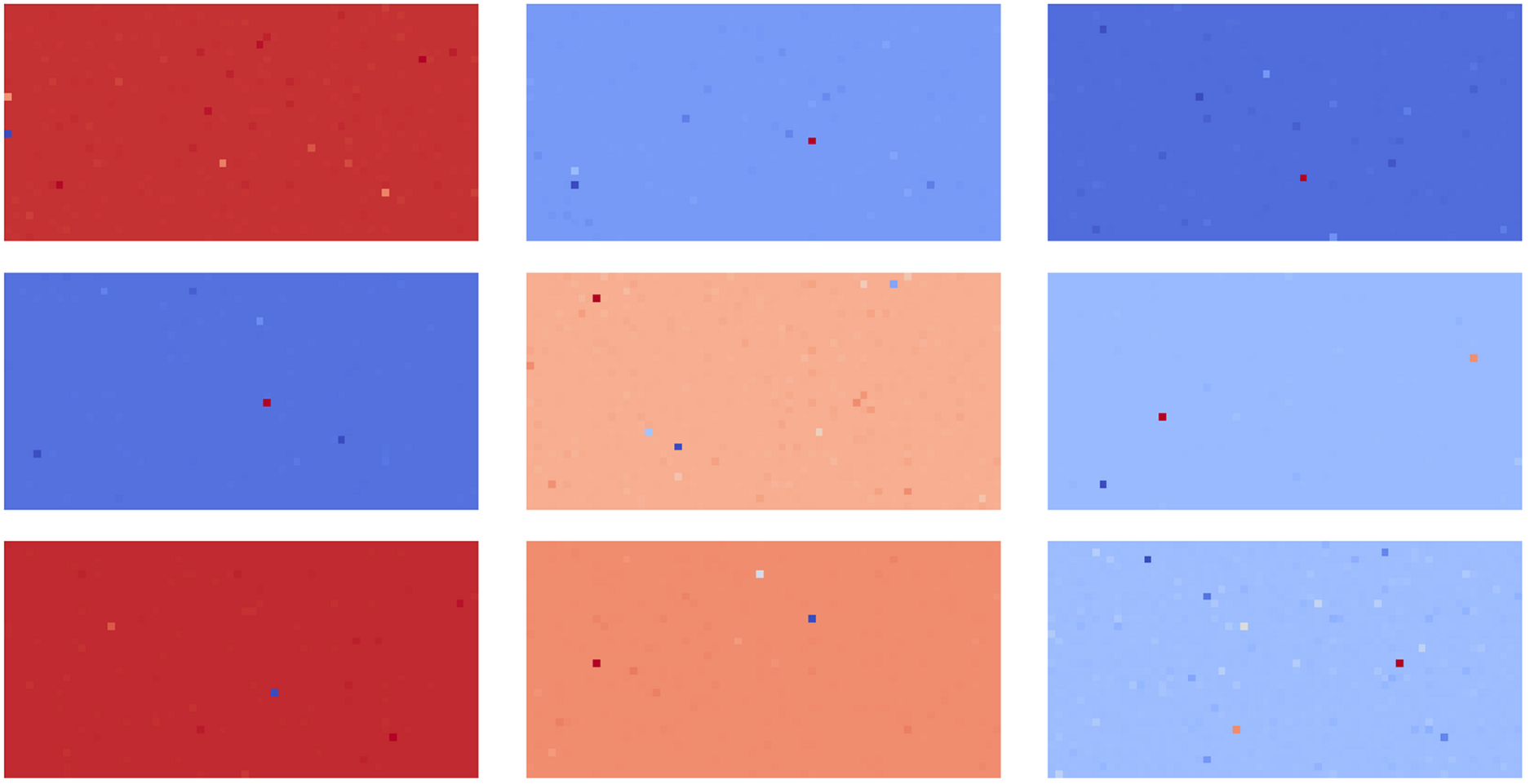}}\hfill
    \subfloat[$\mathbf{c}_r$ of class 1]{\includegraphics[height=0.1\textwidth,width=0.155\textwidth] {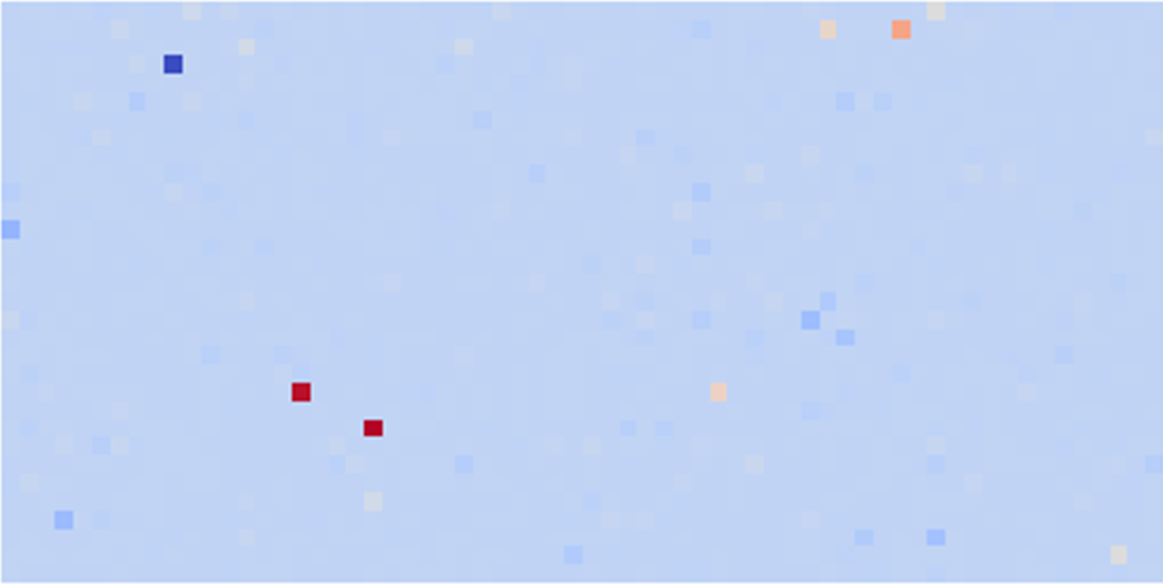}}\hfill
    \subfloat[t-SNE plot of $\mathbf{c}_r$]{\includegraphics[height=0.1\textwidth,width=0.155\textwidth] {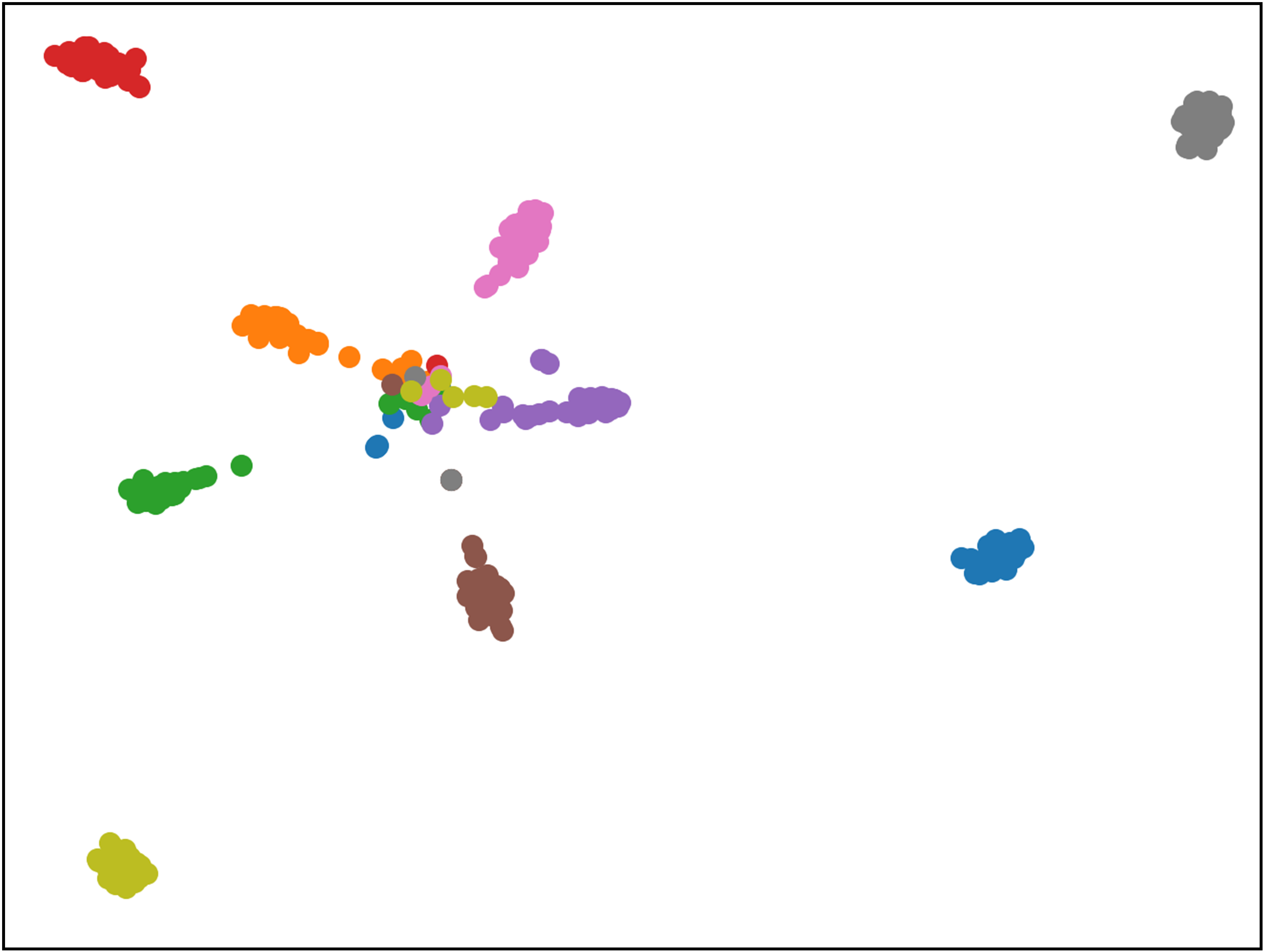}}
    \caption{Discriminability using t-SNE to visualize class separability and compactness (a-d). Aircraft test images using Xception: a) base CNN's output, b) our CAP's impact on the base CNN's output, c) our CAP's output, and d) our model's final output. Our CAP's class-specific attention-aware response for class 1 (e) and class 2 (f) to capture the similarity between 27 integral regions (27$\times$27). Class-specific $\mathbf{c}_r$ in (2) for 9 classes ($3\times 3$) from region 1 (g) and 20 (h). \textcolor{blue}{Blue} to \textcolor{red} {red} represents class-specific \textit{less} to \textit{more} attention towards that region. Class-specific individual feature response within $\mathbf{c}_r$ of the region 1 and class 4 (i). t-SNE plot of $\mathbf{c}_r$ representing images from the above 9 classes 
    (j).   
    }
    \label{fig:fig_vis}
\end{figure*}
\begin{table}
    \centering
    \begin{small}
    \begin{tabular}{l|ccc|ccc}
    \toprule
        &\multicolumn{3}{c|}{Aircraft}  &\multicolumn{3}{c}{Cars}  \\
        {Base CNN} & {\#9} & {\#27} & {\#36} & {\#9} & {\#27} & {\#36} \\
        \toprule
         ResNet-50 &85.9 &94.9 &91.2 &92.9 &94.9 & 91.9 \\
         Xception &87.8 &94.1 &90.0 &93.9 &95.7 &92.6 \\
         NASNet-M &92.7 &93.8 &90.3 &92.4 &93.7 &90.9 \\
         \bottomrule
    \end{tabular}
    \end{small}
    \caption{Accuracy (\%) of our model with a varying number of integral regions. More results in the supplementary in the end.}
    \label{table:abl_3}
\end{table}

Our model's accuracy is also compared using different numbers of regions $|\mathcal{R}|$. It is a hyper-parameter and is computed from $\Delta_x$ and $\Delta_y$. 
The results are shown in Table \ref{table:abl_3} (best $|\mathcal{R}|=27$). We have also provided results for top-N accuracy in the supplementary document provided in the end. The top-2 accuracy is around 99\% and is independent of the CNN types. 

\noindent\textbf{Model complexity:} 
It is represented as a number of trainable parameters in millions and per-image inference time in millisecond (Table \ref{table:abl_2}). It also depends on the base CNNs types (e.g. standard vs lightweight). 
Given the number of trainable parameters (9.7M) and inference time (3.5\textit{ms}), the performance of the lightweight NASNetMobile is very competitive in comparison to the rest. The role of secondary data has improved accuracy in \cite{chang2020devil, cubuk2019autoaugment, ge2019weakly, ge2017borrowing}. 
However, such models involve multiple steps and resource-intensive, resulting in difficulty in implementing. For example, 3 steps in \cite{ge2019weakly}: 1) object detection and instance segmentation (Mask R-CNN and CRF), 2) complementary part mining (512 ROIs) and 3) classification using context gating. The model is trained using 4 GPUs. In contrast, our model can be trained on a single GPU (12 GB). 
The per-image inference time is 4.1\textit{ms}. In \cite{ge2019weakly}, it is 27\textit{ms} for step 3 and additional 227\textit{ms} in step 2. FCANs \cite{liu2016fully} reported its inference time as 150\textit{ms}. Using 27 integral regions and ResNet50 as a base, the training time for the Aircraft is $\sim$4.75 hrs 
for 150 epochs (12 batch size). It is $\sim$5.7 hrs for Cars and $\sim$8.5 hrs for Dogs. 

\noindent\textbf{Qualitative analysis:} To understand the discriminability of our model, 
we use t-SNE \cite{van2014accelerating} to visualize the class separability and compactness in the features extracted from a base CNN, and our novel CAP and classification modules. We also analyze the impact of our CAP in enhancing the discriminability of a base CNN. We use test images in Aircraft and Xception as a base CNN. In Fig. \ref{fig:fig_vis}(a-d), it is evident that when we include our CAP + encoding modules, the clusters are farther apart and compact, resulting in a clear distinction of various clusters representing different subcategories. Moreover, the discriminability of the base CNN is significantly improved (Fig. \ref{fig:fig_vis}b) in comparison to without our modules shown in Fig. \ref{fig:fig_vis}a. More results are shown in the supplementary material, added in the end. 
We have also looked the inside of our CAP by visualizing its class-specific attention-aware response using $\alpha_{r,r'}$ and context vector $\mathbf{c}_r$ in (\ref{eq:attn}). Aircraft images (randomly selected 9 classes) are used 
in Fig. \ref{fig:fig_vis}(e-j). Such results clearly show our model's power in capturing the context information for discriminating subtle changes in FGVC problems. We have also included some examples, which are incorrectly classified by our model with an explanation in the supplementary information in the end. 
\section{Conclusion}
We have proposed a novel approach for recognizing subcategories by introducing a simple formulation of context-aware attention via learning where to look when pooling features across an image. Our attention allows for explicit integration of bottom-up saliency by taking advantages of integral regions and their importance, without requiring the bounding box/part annotations. We have also proposed a feature encoding by considering the semantic correlation among the regions and their spatial layouts to encode complementary partial information.
Finally, our model's SotA results on eight benchmarked datasets, quantitative/qualitative results and ablation study justify the efficiency of our approach. Code is available at https://ardhendubehera.github.io/cap/.

\section{ Acknowledgments}
This research is supported by the UKIERI-DST grant CHARM (DST UKIERI-2018-19-10). The GPU used in this research is generously donated by the NVIDIA Corporation.
\begin{quote}
\begin{small}
\bibliography{Ref}
\end{small}
\end{quote}

\clearpage
\onecolumn

\begin{center}
\Large

 \textbf{Supplementary Document}   
\vspace{0.2cm}
\large

 \vspace{0.4cm}
\end{center}

\noindent In this document, we have included the remaining quantitative and qualitative results, which we could not include in the main paper.\\

\noindent\textbf{Remaining results of Table 2:} The performance comparison (accuracy in \%) using the remaining two datasets (Stanford Dogs and Oxford Flowers) for Table 2 in the main paper. It is presented in Table \ref{table:sota1} below.

\begin{table} [!htbp]
  \caption{Performance comparison with the recent top-five SotA approaches on each dataset. Methods marked with * involve transfer/joint learning strategy for objects/patches/regions consisting more than one dataset (primary and secondary)}
  \label{table:sota1}
  \centering
  \begin{tabular}{lc|lc}
    \toprule
      \multicolumn{2}{c}{\textbf{Stanford Dogs}}&
      \multicolumn{2}{c}{\textbf{Oxford Flowers}} \\
      {Method} & {Accuracy (\%)} & {Method} & {Accuracy (\%)} \\
      \midrule
   FCANs  (Liu et al. 2016) 
   &	89.0  & InterAct (Xie et al. 2016) 
   & 96.4   \\
    SJFT$^*$  (Ge and Yu 2017)
    & 90.3 &   SJFT$^*$   (Ge and Yu 2017) 
    & 97.0 \\       
    DAN  (Hu et al. 2019)
    &92.2    &OPAM$^*$ (Peng, He, and Zhao 2018) 
    & 97.1 \\
   WARN (Rodríguez et al. 2020) 
   &92.9 &DSTL$^*$ (Cui et al. 2018) 
   & 97.6 \\
    CPM$^*$  (Ge, Lin, and Yu 2019) 
    & \textbf{97.1}& MC$_{Loss}*$
    (Chang et al. 2020) 
    & \textbf{97.7} \\ 
         \hline
    \textbf{Proposed} & {96.1} &\textbf{Proposed} & \textbf{97.7} \\
    \bottomrule
  \end{tabular}
\end{table}

\noindent\textbf{Remaining results of Table 3:} The accuracy of the proposed method is evaluated on the \textbf{NABirds} dataset using six different SotA base CNNs for Table 3 in the main paper. It is presented in Table  \ref{table:sota_baseCNN1} below.

\begin{table}  [!htbp]
  \caption{Our model’s accuracy (\%) on the \textbf{NABirds} dataset with different SotA base CNN architectures. Previous best accuracy is 86.4\% (Luo et al. 2019) for primary only and 87.9\% (Cui et al. 2018) for combined primary and secondary datasets.}
  
  \label{table:sota_baseCNN1}
  \centering
  \begin{tabular}{l| c}
    \toprule
      
      {Base CNN} & {Accuracy(\%)} \\
  \midrule
   ResNet-50 &	88.8    \\
   Inception V3 & 89.1   \\       
    Xception & \textbf{91.0}    \\
   DenseNet-121 &88.3  \\
NASNet-Mobile &88.7  \\
MobileNet V2 &89.1 \\

    \bottomrule  
  \end{tabular}
\end{table}


\noindent\textbf{Remaining results of Table 4:} In ablation study (Table 4 of the main paper), we have presented the performance of the proposed model (with the addition of our novel context-aware attentional pooling (+C) and classification (+E) module) on the Aircraft, Stanford Cars and Oxford-IIIT Pets datasets. The same evaluation procedure is performed on the Stanford Dogs, Oxford Flowers and Caltech Birds (CUB-200) datasets and the recognition accuracy (\%) is presented in Table \ref{table:abl_s3}. Like in Table 4, a similar trend is observed in the improvement of accuracy when our context-aware attentional pooling (+C) and classification (+E) modules are added to various SotA base CNN architectures (B). 

\begin{table*}[!htbp]
  \caption{Accuracy (\%) of the proposed model with the addition of our novel context-aware attentional pooling (+C) and classification (+E) module to various SotA base (B) CNN architectures. It presents the remaining evaluation of Table 4.}
  \label{table:abl_s3}
  \centering
  \begin{tabular}{l|ccc|ccc|ccc}
    \toprule
    &\multicolumn{3}{c}{\textbf {Stanford Dogs} } &
      \multicolumn{3}{c}{\textbf {Oxford Flowers} } &
      \multicolumn{3}{c}{\textbf {Caltech Birds: CUB-200}}\\
      {Base CNN} & {B} & {B+C} & {B+C+E} & {B} & {B+C} & {B+C+E} & {B} & {B+C} & {B+C+E} \\
      \midrule
   
   Inception-V3 &78.7 &94.2 &95.7 &92.3 &94.9 &97.6 &76.0 &87.1 &91.4 \\  
   Xception     &82.7 &94.8 &96.1 &91.9 &94.9 &97.7 &75.6 &87.4 &91.8 \\ 
   DenseNet-121  &79.5 &94.5 &95.5 &94.4 &95.1 &97.6 &79.1 &87.2 &91.6\\ 
   NASNet-Mobile &79.5 &94.7 &96.0 &90.7 &95.0 &97.7 &73.0 &86.8 &89.7 \\ 
   MobileNetV2  &76.5 &94.3 &95.9 &92.3 &95.0 &97.4 &74.5 &87.0 &89.2 \\
   \midrule
   Previous Best    & \multicolumn{2}{c}  (Ge et al. 2019)   &93.9 &   \multicolumn{2}{c}(Xie et al. 2016)   &96.4  & \multicolumn{2}{c} (Ge et al. 2019) &90.3  \\ 
    \bottomrule
  \end{tabular}\vspace{-1mm}
\end{table*}

\vspace{15mm}

\noindent\textbf{Remaining results of Table 5:} The performance is evaluated using a different number of integral regions on the Aircraft and Stanford Cars datasets (Table 5). The same experiment is also carried out on the Stanford Dogs dataset, and the results are given in Table \ref{table:abl_s5} below. 

\begin{table}  [!htbp]
  \caption{Accuracy (\%) of our model with numbers of 9, 27, and 36 integral regions on \textbf{Stanford Dogs } dataset.}
  
  \label{table:abl_s5}
  \centering
  \begin{tabular}{l|ccc}
    \toprule
      
      {Base CNN} & {\#9} & {\#27}  & {\#36}  \\
  \midrule
   ResNet-50 &	90.5 &95.8 &92.1     \\
   Xception  &95.3 &96.1 &95.2   \\       
    NASNet-M  &91.7 &96.0 &93.3    \\

    \bottomrule  
  \end{tabular}
\end{table}

\noindent\textbf{Top-N Accuracy (\%):} We have also evaluated the proposed approach using top-N accuracy metric on Oxford-IIIT Pets, Stanford Cars and Aircraft datasets. The performance of our modules on top of various base architectures is presented in Table \ref{table:abl_s4} below. On all three datasets, the top-2 accuracy is around 99\% and is independent of the type of base CNN architecture used. Moreover, the top-5 accuracy is nearly 100\%. This justifies the significance of our novel attentional pooling and encoding modules in enhancing performance and their wider applicability.

\begin{table*}[!htbp]
  \caption{Top-N accuracy (in \%) of the proposed model using different base architectures on Oxford-IIIT Pets, Stanford Cars and Aircraft datasets. The top-2 accuracy is around 99\% and is independent of the type of base CNN architecture used. The top-5 accuracy is nearly 100\%. This shows the significance of the proposed attentional pooling and encoding modules.}
  \label{table:abl_s4}
  \centering
  \begin{tabular}{l|lcccc}
  \toprule
  Dataset &Base CNN architecture &Top 1 &Top 2 &Top 3 &Top 5 \\
  \midrule
  Oxford-IIIT Pets &Inception-V3 &96.2 &99.0 &99.5 &99.9 \\
  &Xception &97.0 &99.7 &99.9 &99.9 \\
  &DenseNet121 &96.9 &99.2 &99.6 &99.7 \\
  &NASNetMobile &97.3 &99.4 &99.8 &99.9 \\
  &MobileNetV2 &96.4 &98.9 &99.5 &99.6 \\
  \midrule
  Stanford Cars &Inception-V3  &94.8 &99.4 &99.7 &99.8 \\
  &Xception &95.7 &99.3 &99.7 &99.8 \\
  &DenseNet121 &93.6 &98.7 &99.5 &99.9 \\
  &NASNetMobile &93.7 &99.1 &99.7 &99.8 \\
  &MobileNetV2 &94.0 &99.3 &99.8 &99.9 \\
  \midrule
  Aircraft  &Inception-V3 &94.8 &99.1 &99.7 &99.8 \\
  &Xception &94.1 &98.9 &99.2 &99.5 \\
  &DenseNet121 &94.6 &98.8 &99.3 &99.4 \\
  &NASNetMobile &93.8 &99.4 &99.8 &99.8 \\
  &MobileNetV2 &94.4 &99.1 &99.7 &99.8 \\
  \bottomrule
  \end{tabular}
\end{table*}




\vspace{1 cm}
\noindent\textbf{Additional Qualitative Analysis: \\}

\begin{figure*}[!ht]
     \centering
     \subfloat[][747-200 vs 747-100]{\includegraphics[width=0.16\textwidth, height=0.13\textwidth]{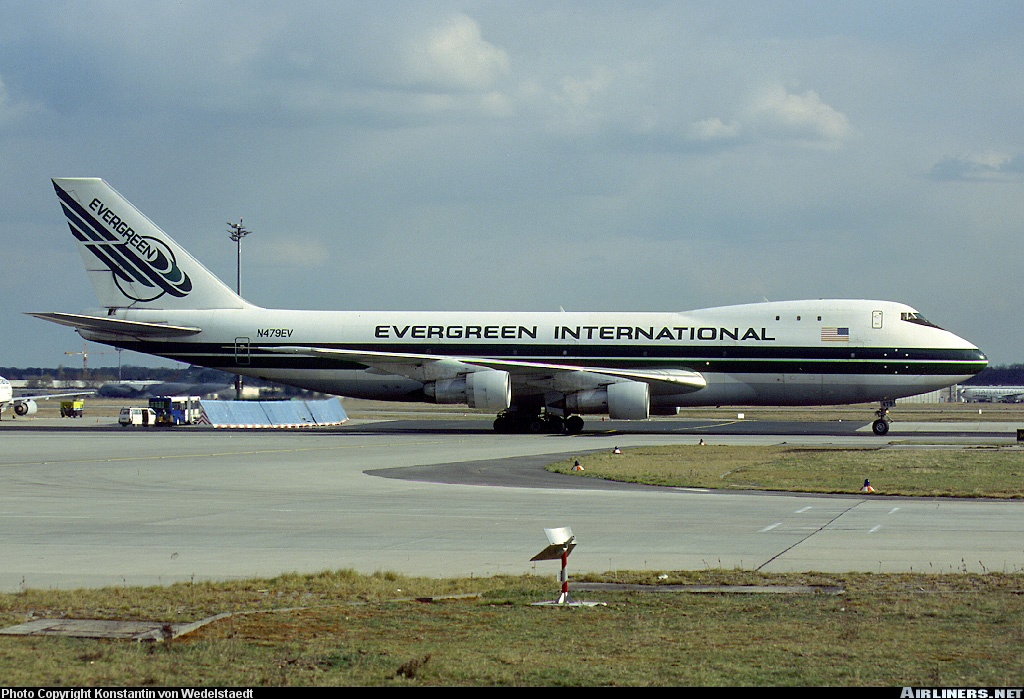}
     \includegraphics[width=0.16\textwidth, height=0.13\textwidth]{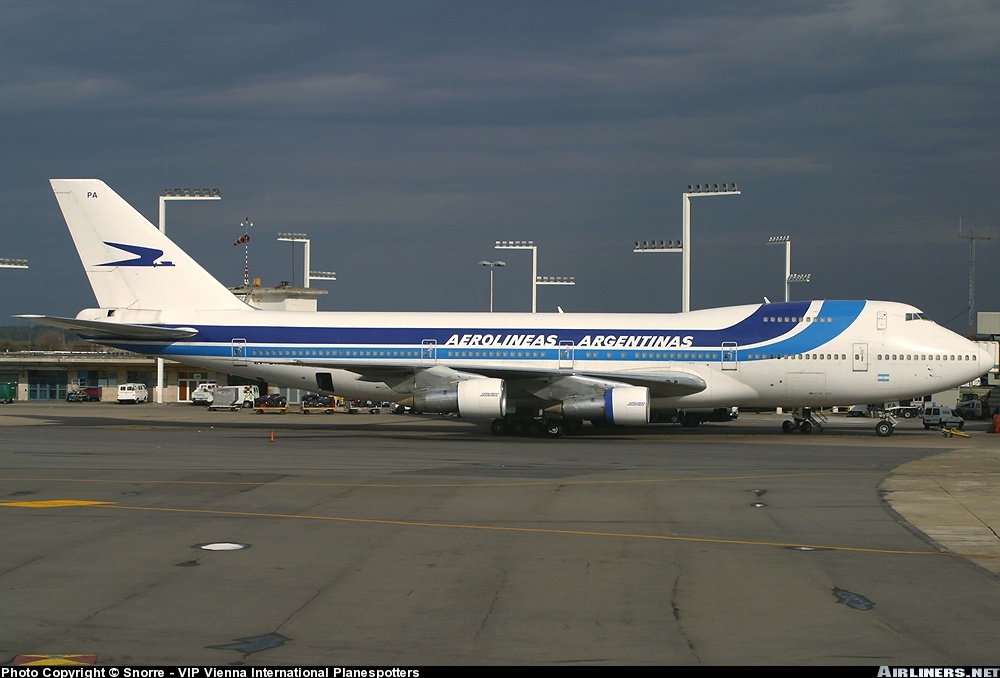}}\hfill
     \subfloat[][747-300 vs 747-400]{\includegraphics[width=0.16\textwidth, height=0.13\textwidth]{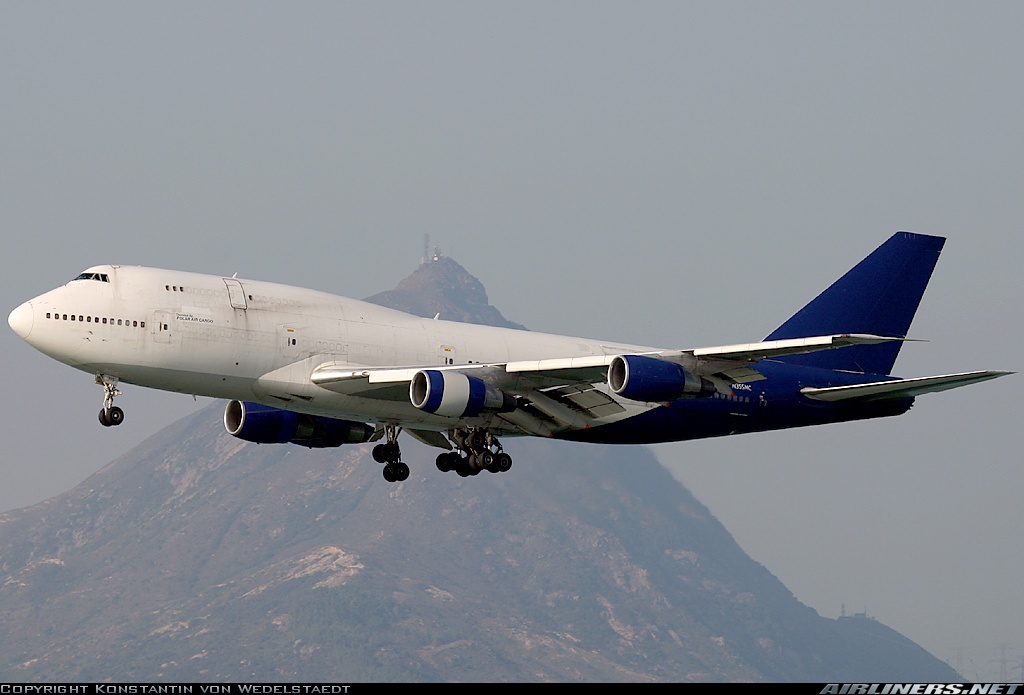}
     \includegraphics[width=0.16\textwidth, height=0.13\textwidth]{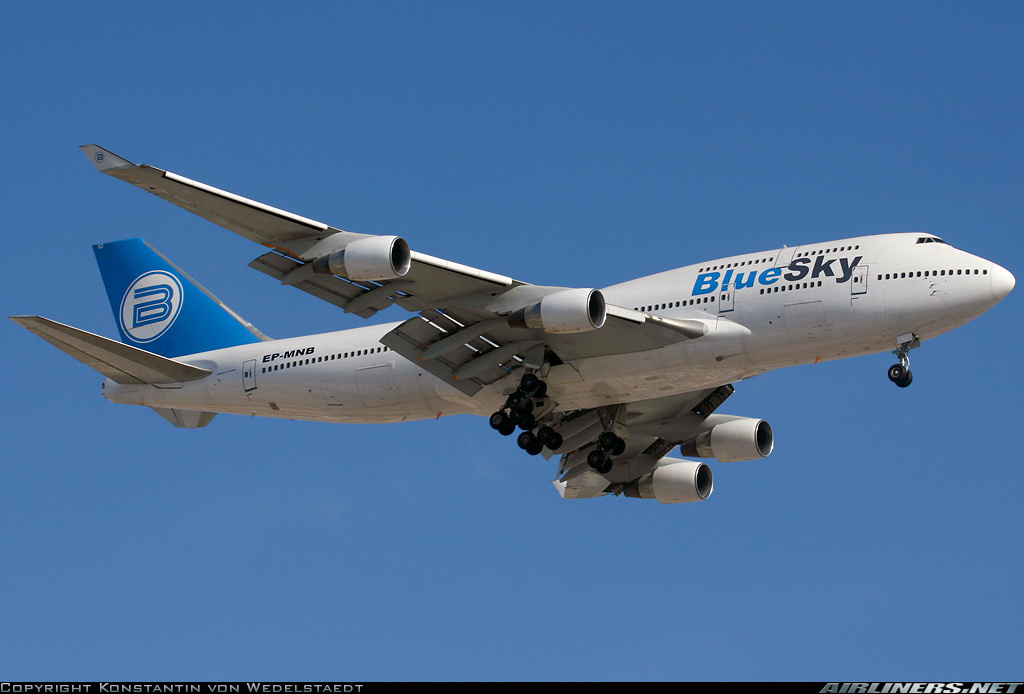}}\hfill
     \subfloat[][C-47 vs DC-3]{\includegraphics[width=0.16\textwidth, height=0.13\textwidth]{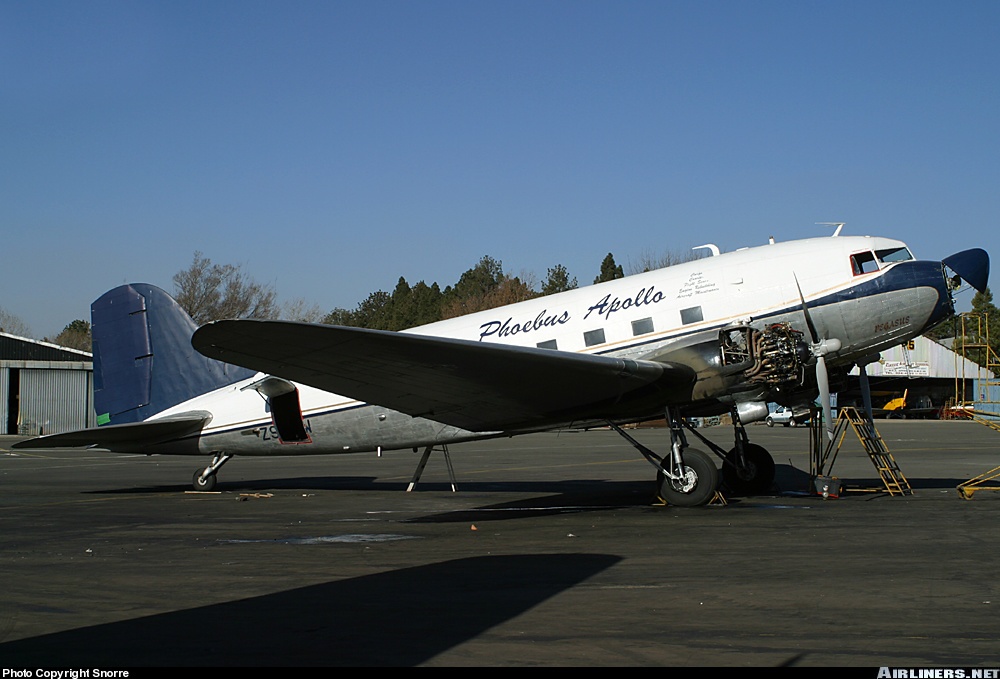}
     \includegraphics[width=0.16\textwidth, height=0.13\textwidth]{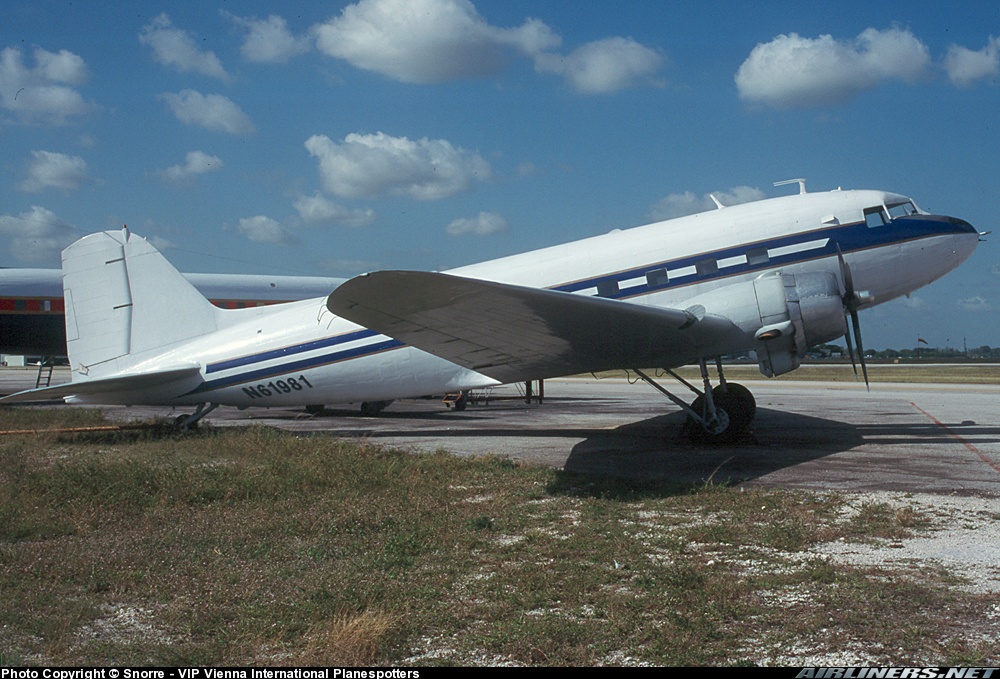}}\hfill
     
     \subfloat[][Audi TTS Coupe 2012 vs Audi TT RS Coupe 2012]{\includegraphics[width=0.16\textwidth, height=0.13\textwidth]{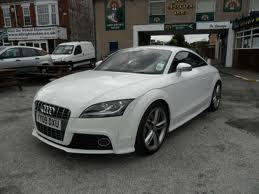}
     \includegraphics[width=0.16\textwidth, height=0.13\textwidth]{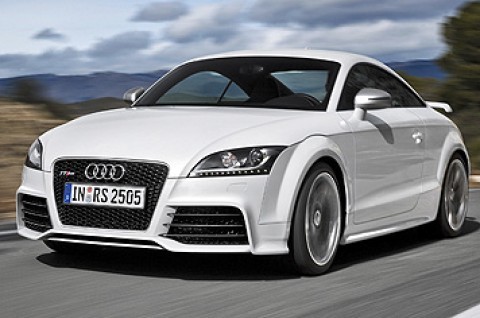}}\hfill
     \subfloat[][Bentley Continental GT Coupe 2012 vs Bentley Continental GT Coupe 2007]{\includegraphics[width=0.16\textwidth, height=0.13\textwidth]{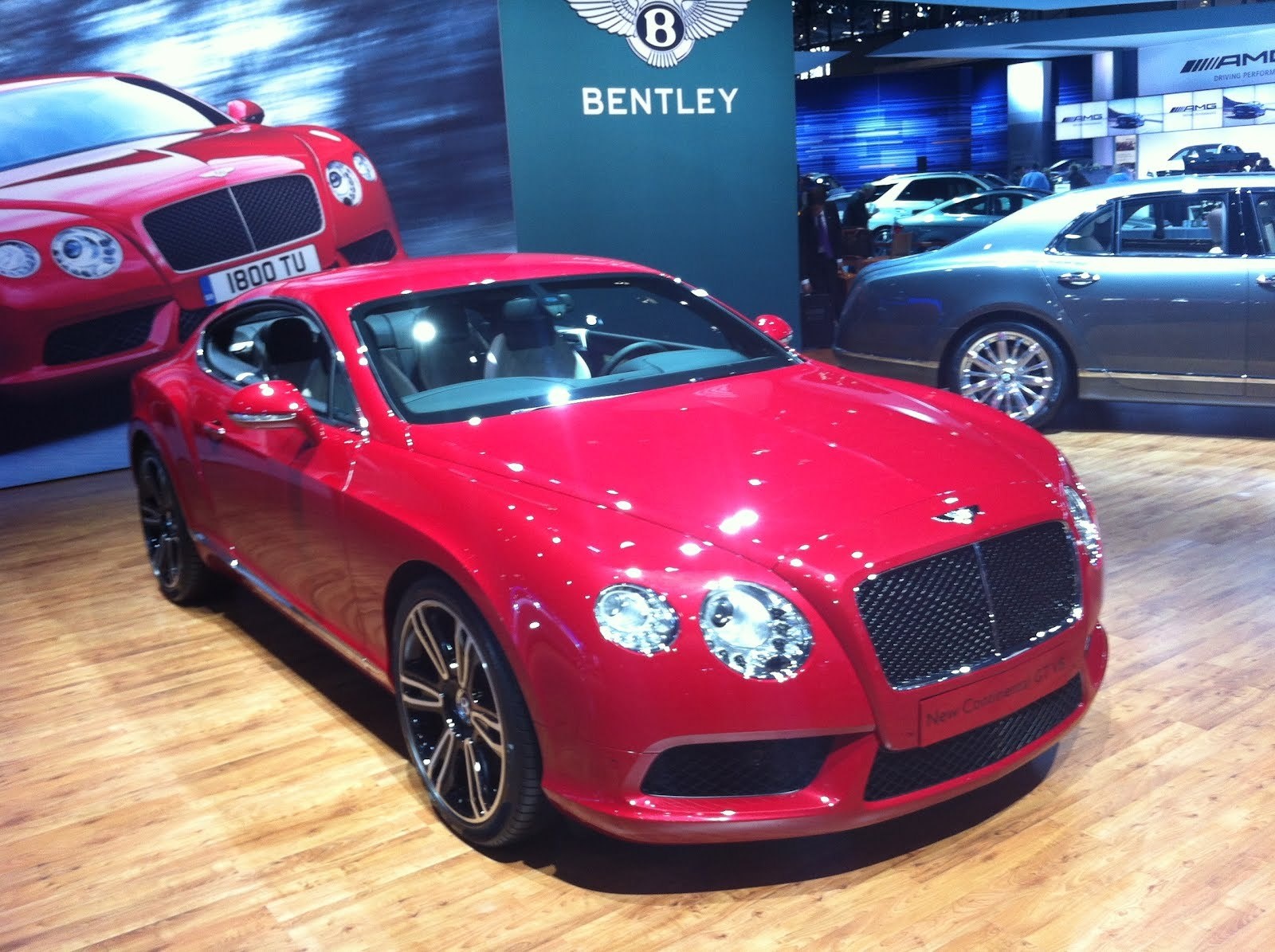}
     \includegraphics[width=0.16\textwidth, height=0.13\textwidth]{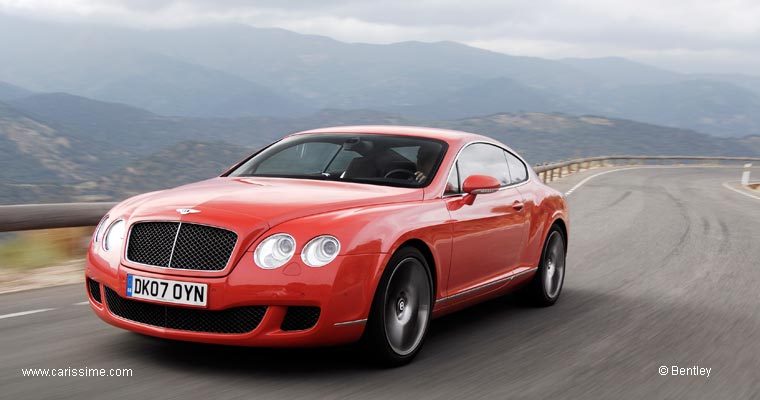}}\hfill
     \subfloat[][Chevrolet Express Cargo Van 2007 vs Chevrolet Express Van 2007]{\includegraphics[width=0.16\textwidth, height=0.13\textwidth]{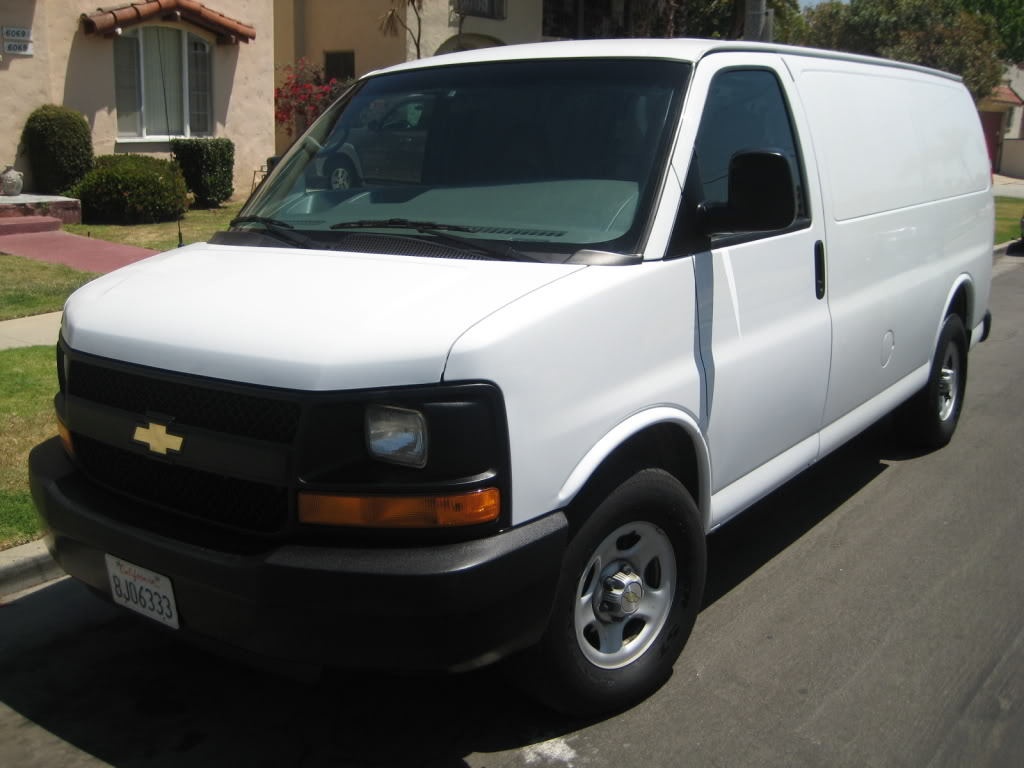}
     \includegraphics[width=0.16\textwidth, height=0.13\textwidth]{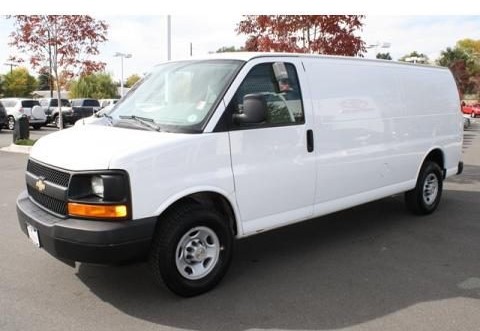}}\hfill
     
     \subfloat[][Birman vs Ragdoll]{\includegraphics[width=0.16\textwidth, height=0.13\textwidth]{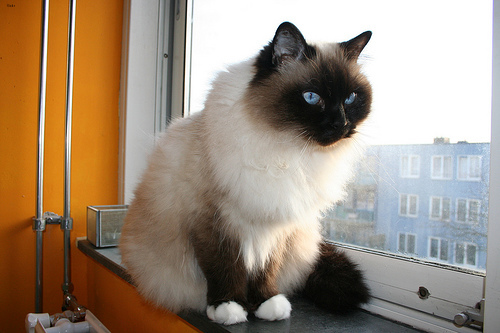}
     \includegraphics[width=0.16\textwidth, height=0.13\textwidth]{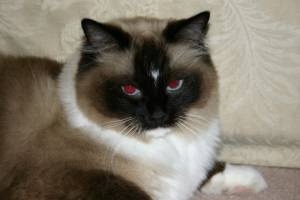}}\hfill
     \subfloat[][American Pitbull Terrier vs American Bulldog]{\includegraphics[width=0.16\textwidth, height=0.13\textwidth]{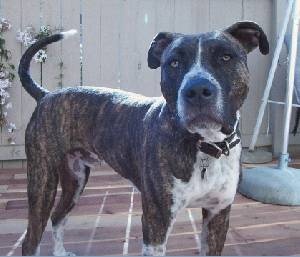}
     \includegraphics[width=0.16\textwidth, height=0.13\textwidth]{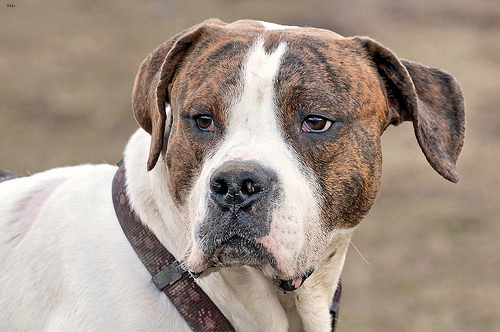}}\hfill
     \subfloat[][Staffordshire Bull Terrier vs American Pitbull Terrier]{\includegraphics[width=0.16\textwidth, height=0.13\textwidth]{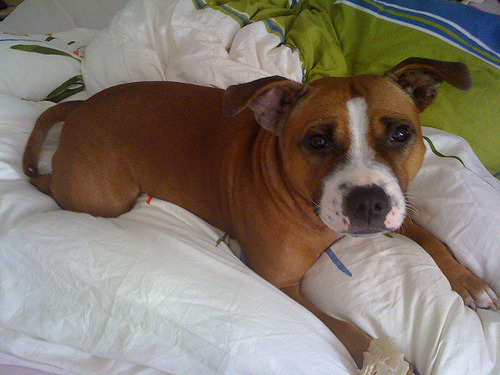}
     \includegraphics[width=0.16\textwidth, height=0.13\textwidth]{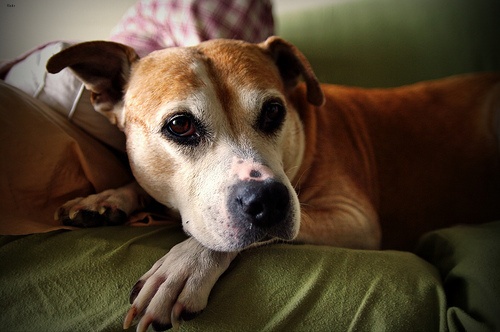}}\hfill
     
     \subfloat[][Spotted Catbird vs Gray Catbird]{\includegraphics[width=0.16\textwidth, height=0.13\textwidth]{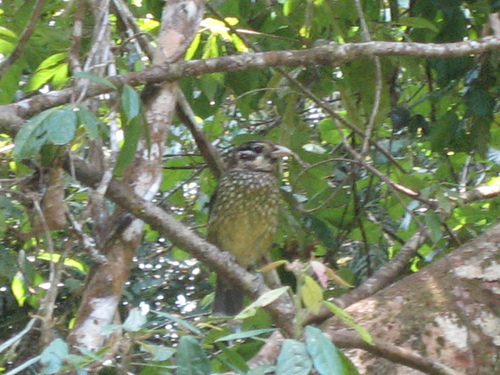}
     \includegraphics[width=0.16\textwidth, height=0.13\textwidth]{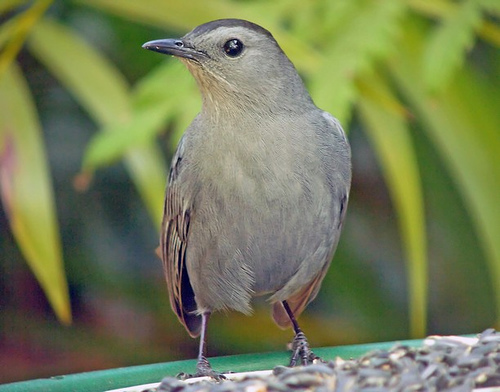}}\hfill
     \subfloat[][Red Winged Blackbird vs Brewer Blackbird]{\includegraphics[width=0.16\textwidth, height=0.13\textwidth]{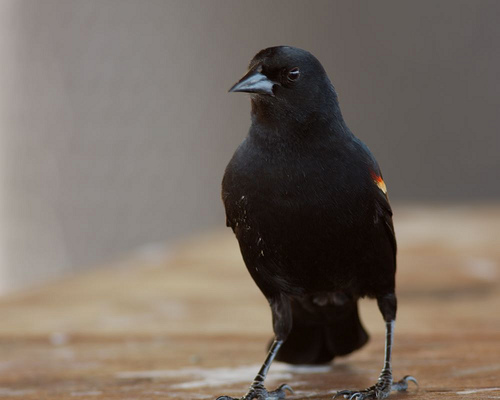}
     \includegraphics[width=0.16\textwidth, height=0.13\textwidth]{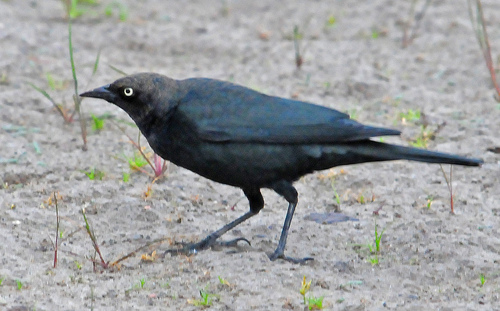}}\hfill
     \subfloat[][Laysan Albatross vs Sooty Albatross]{\includegraphics[width=0.16\textwidth, height=0.13\textwidth]{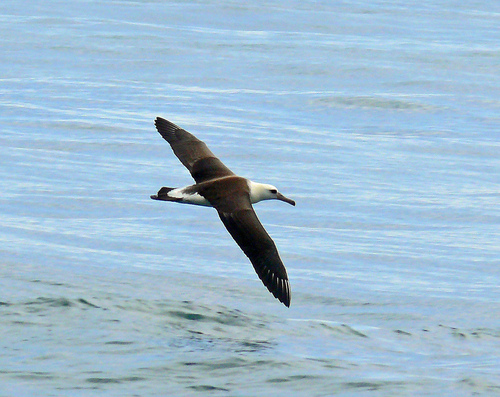}
     \includegraphics[width=0.16\textwidth, height=0.13\textwidth]{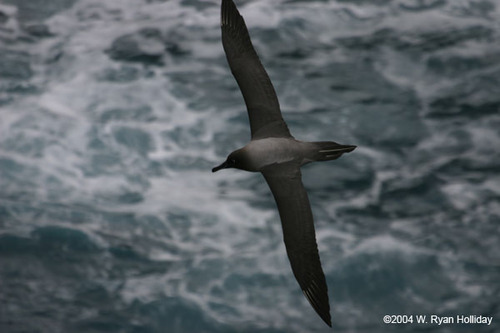}}\hfill
     
     \subfloat[][English Marigold vs Dandelion]{\includegraphics[width=0.16\textwidth, height=0.13\textwidth]{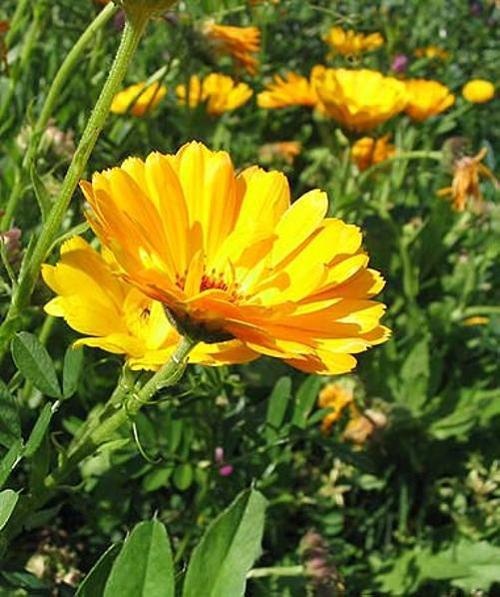}
     \includegraphics[width=0.16\textwidth, height=0.13\textwidth]{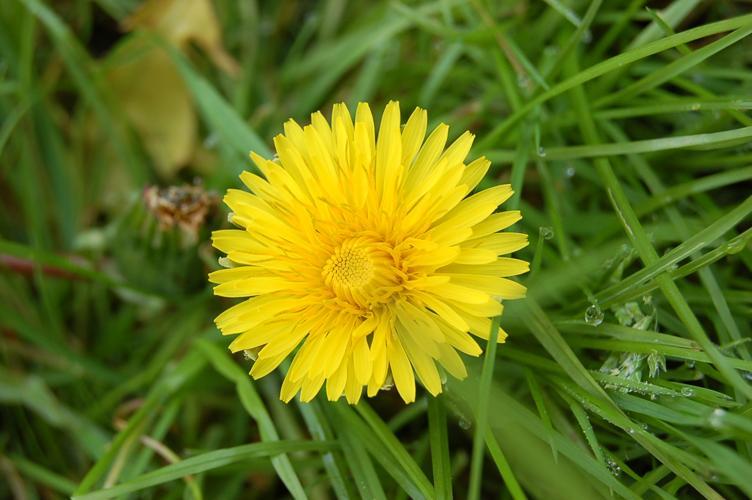}}\hfill
     \subfloat[][Sweet Pea vs Lenten Rose]{\includegraphics[width=0.16\textwidth, height=0.13\textwidth]{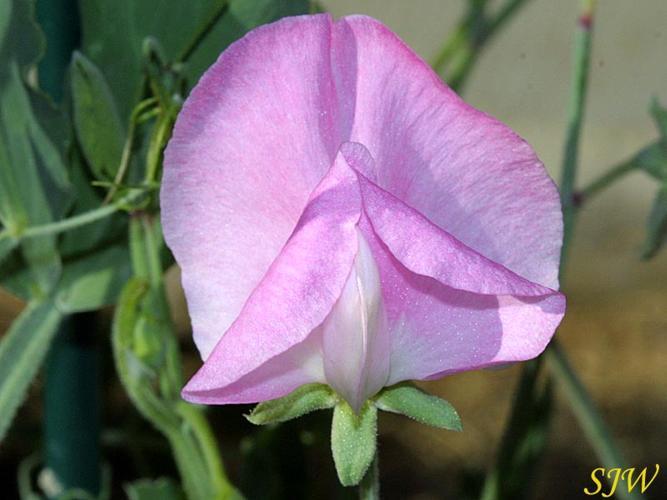}
     \includegraphics[width=0.16\textwidth, height=0.13\textwidth]{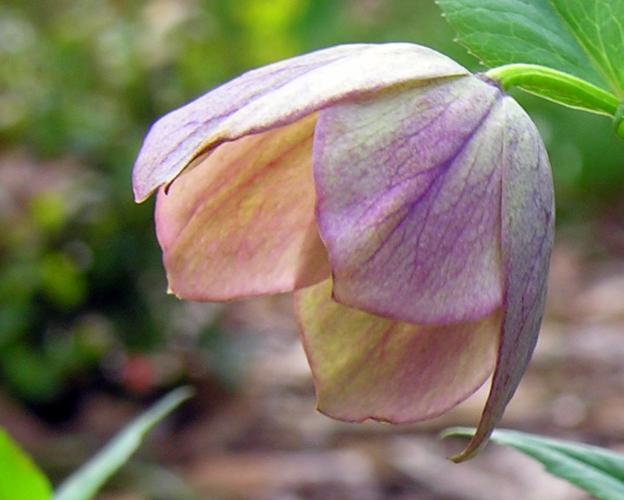}}\hfill
     \subfloat[][Clematis vs Hibiscus]{\includegraphics[width=0.16\textwidth, height=0.13\textwidth]{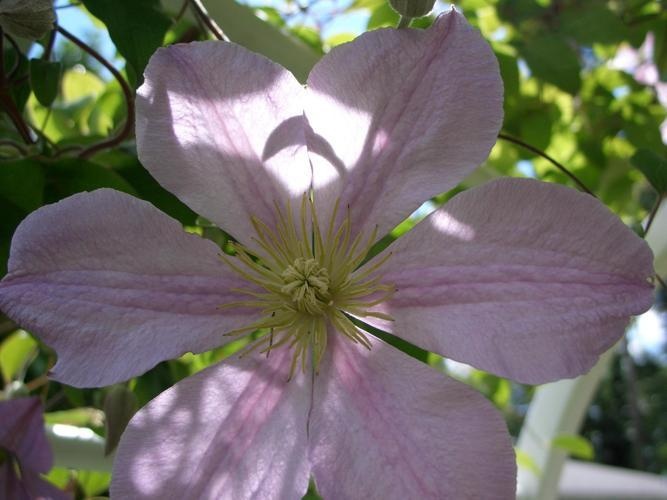}
     \includegraphics[width=0.16\textwidth, height=0.13\textwidth]{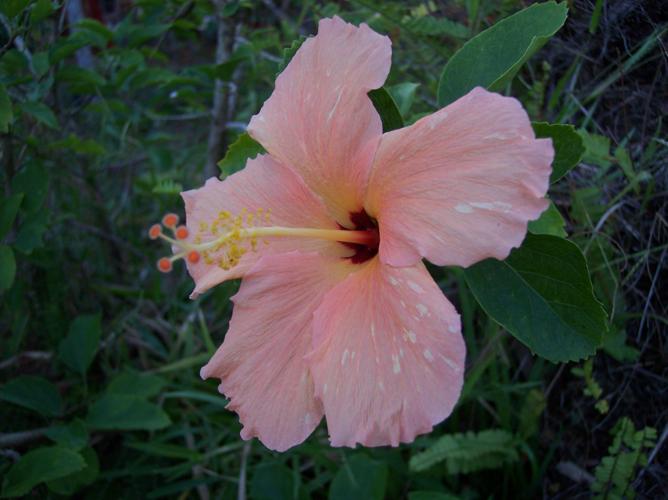}}\hfill
     
     \caption{Some of the example images, which are incorrectly classified by our model (left) against the label they are mistaken for (right - selected from the mistaken subcategories): Aircraft (a-c), Stanford Cars (d-f), Oxford-IIIT Pets (g-i), Caltech-UCSD Birds - CUB-200 (j-l), and Oxford Flowers (m-o). It can be seen that the mistaken labelling comes from classes with extremely similar appearance features and/or perspective changes, often being from the same manufacturer (Boeing 747, Audi, etc.). We have also noticed that subcategories can have very specific defining features that are not clearly visible in every image due to poor angles or lighting conditions (e.g. The chin of a Ragdoll and legs of a Birman cat).}
     \label{fig:examples}
\end{figure*}

\noindent We have provided the additional qualitative analysis of our model's performance by selecting a few example images, which are wrongly classified against the label they are mistaken for (selected from the mistaken subcategories). This is presented in Figure \ref{fig:examples}. It is evident that the mistaken labels come from classes with extremely similar features, often being from the same manufacturer (Boeing 747, Audi, etc.). We have also noticed that subcategories can have very specific defining features that are not clearly visible in every image due to poor angles or lighting conditions (e.g. The chin of a Ragdoll and legs of a Birman cat shown in Fig. \ref{fig:examples}g).

\clearpage
\vspace{3mm}
\noindent We have also included an additional qualitative analysis of discriminating ability (Figure 5 to Figure 9) of our model using t-SNE to visualize class separability and compactness on the different datasets as well as various backbone CNNs.

\begin{figure*}[ht]
     \centering
     \subfloat[][Base CNN (Inception-V3)]{\includegraphics[width=0.33\textwidth]{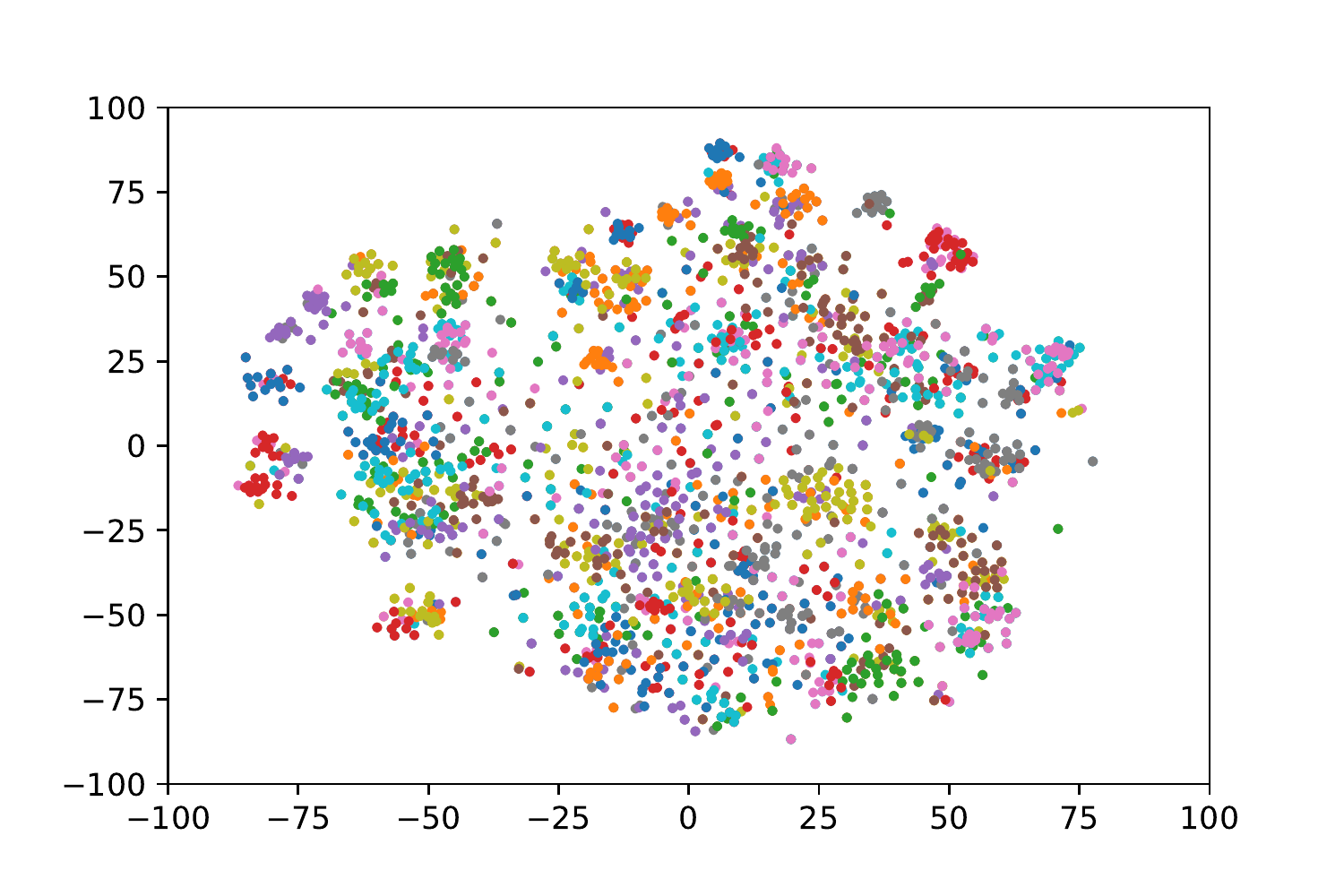}}\hfill
     \subfloat[][Attentional Pooling (Inception-V3)]{\includegraphics[width=0.33\textwidth]{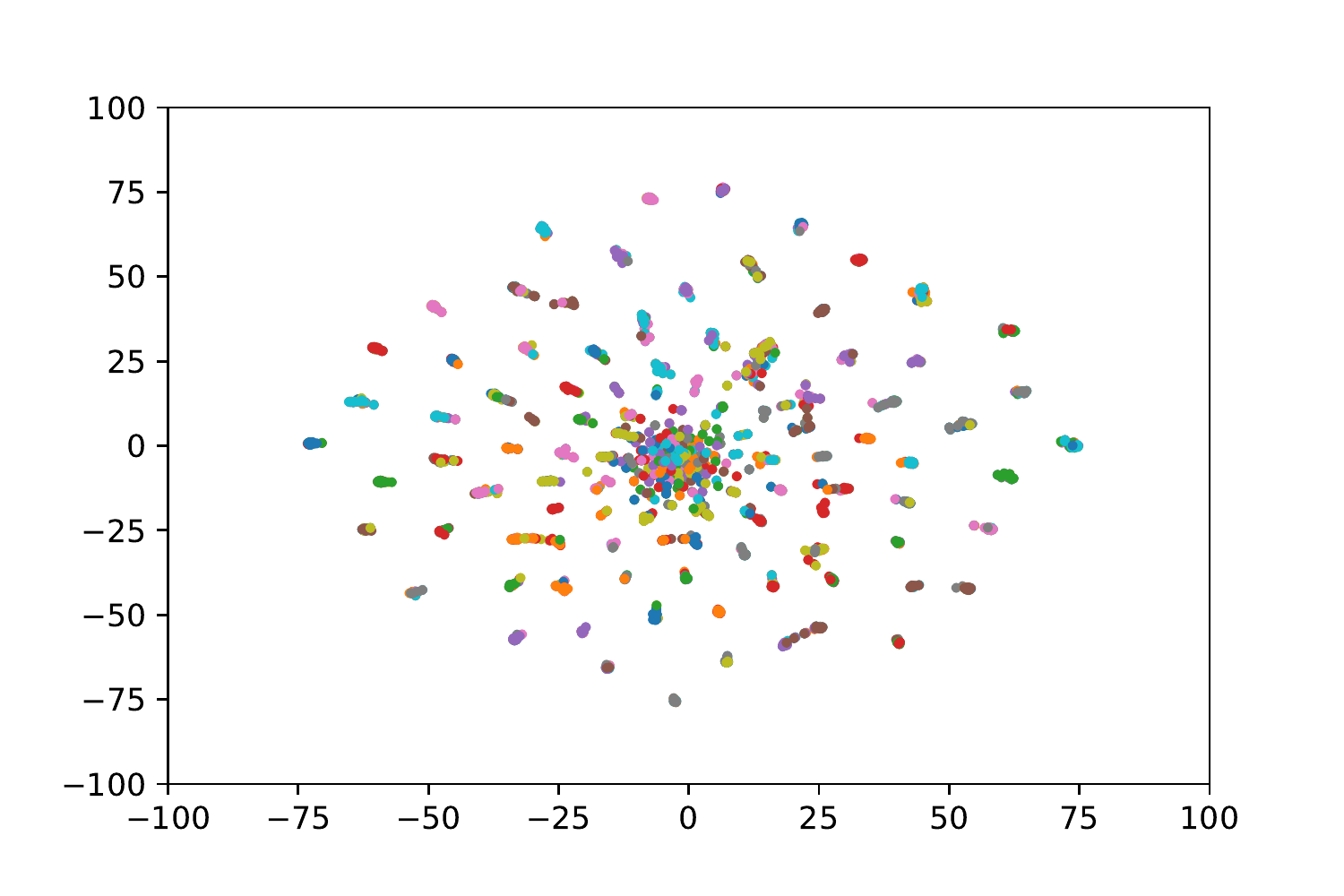}}\hfill
     \subfloat[][CAP + Encoding (Inception-V3)]{\includegraphics[width=0.33\textwidth]{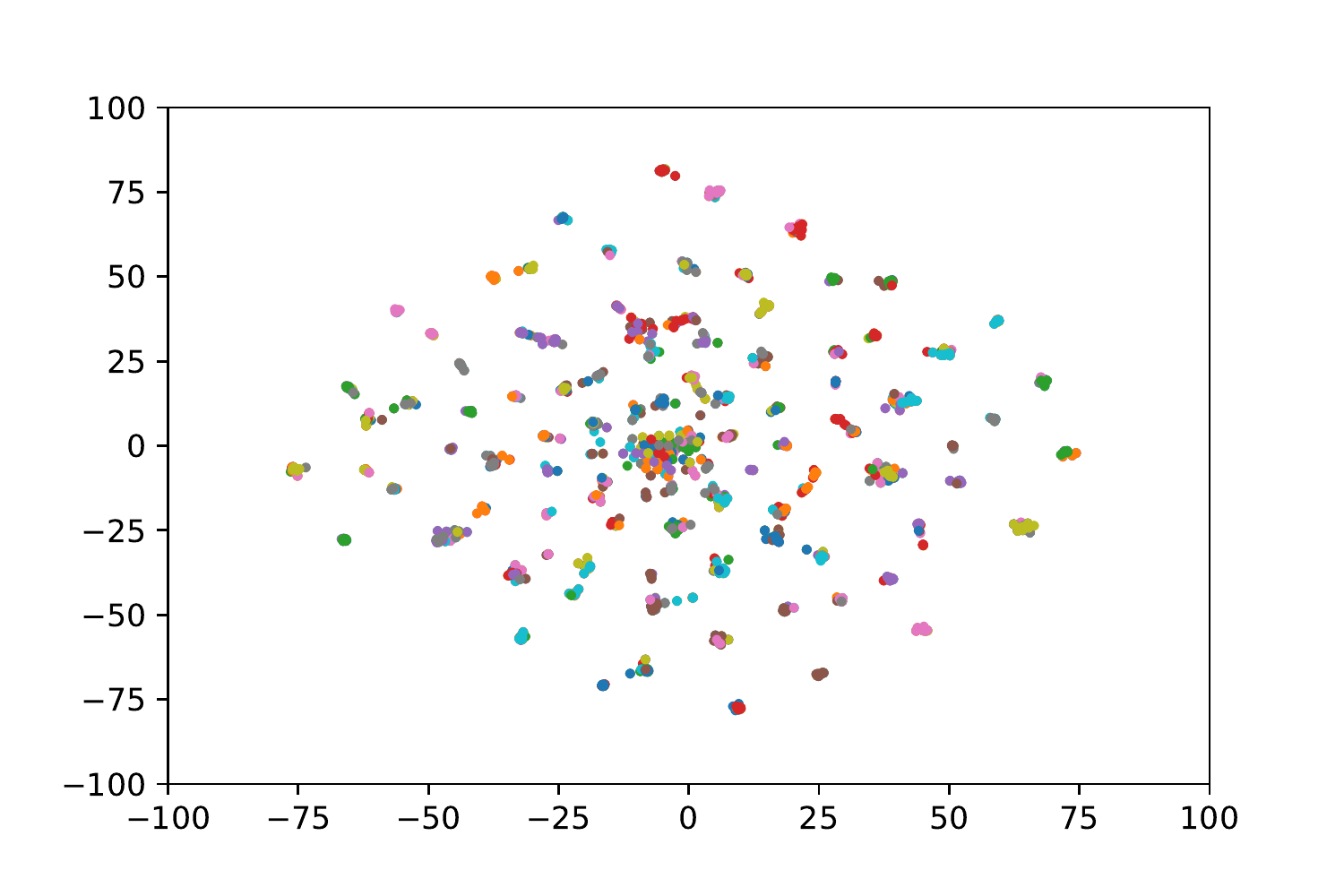}}\hfill
     
     \subfloat[][Base CNN (NASNetMobile)]{\includegraphics[width=0.33\textwidth]{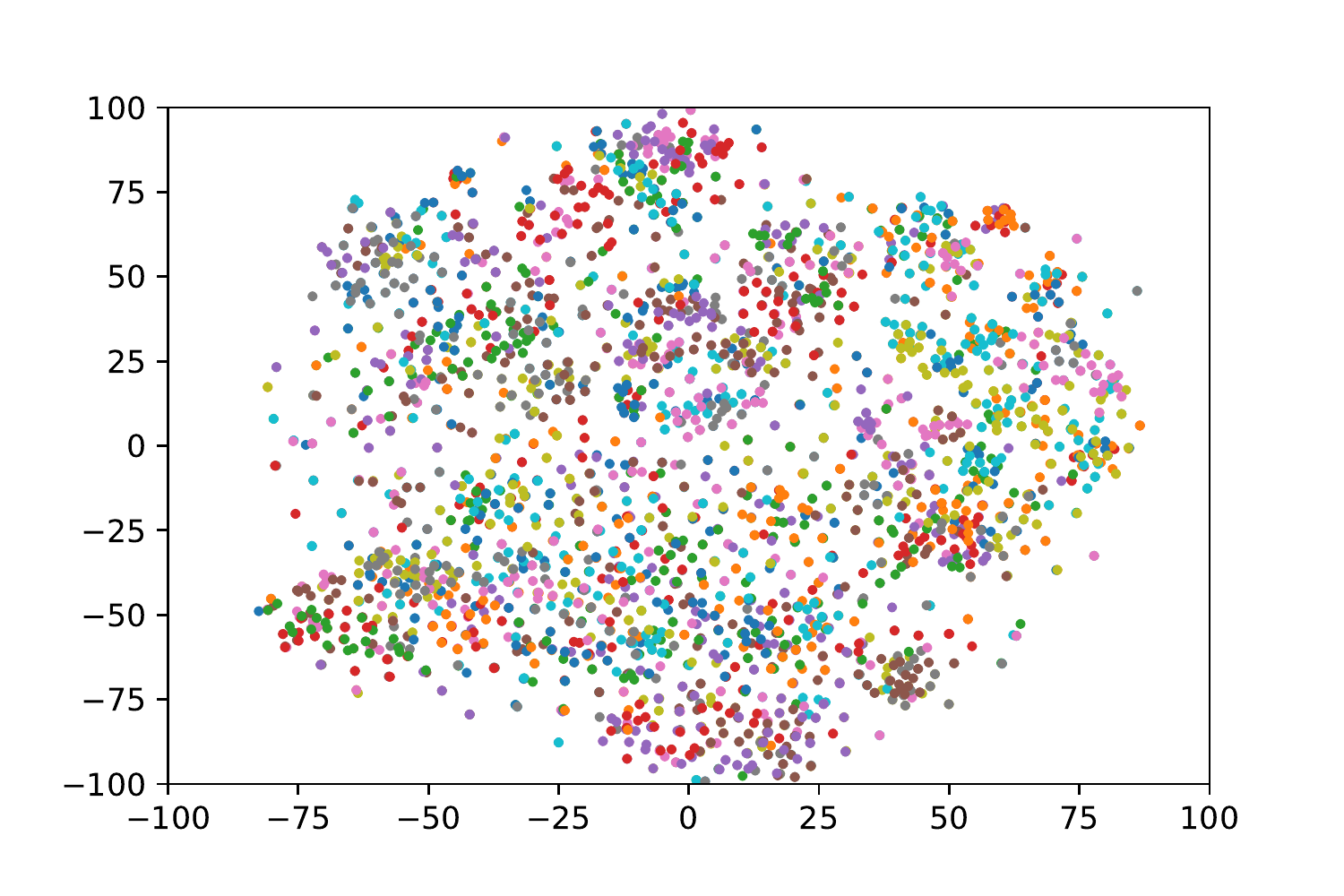}}\hfill
     \subfloat[][Attentional Pooling (NASNetMobile)]{\includegraphics[width=0.33\textwidth]{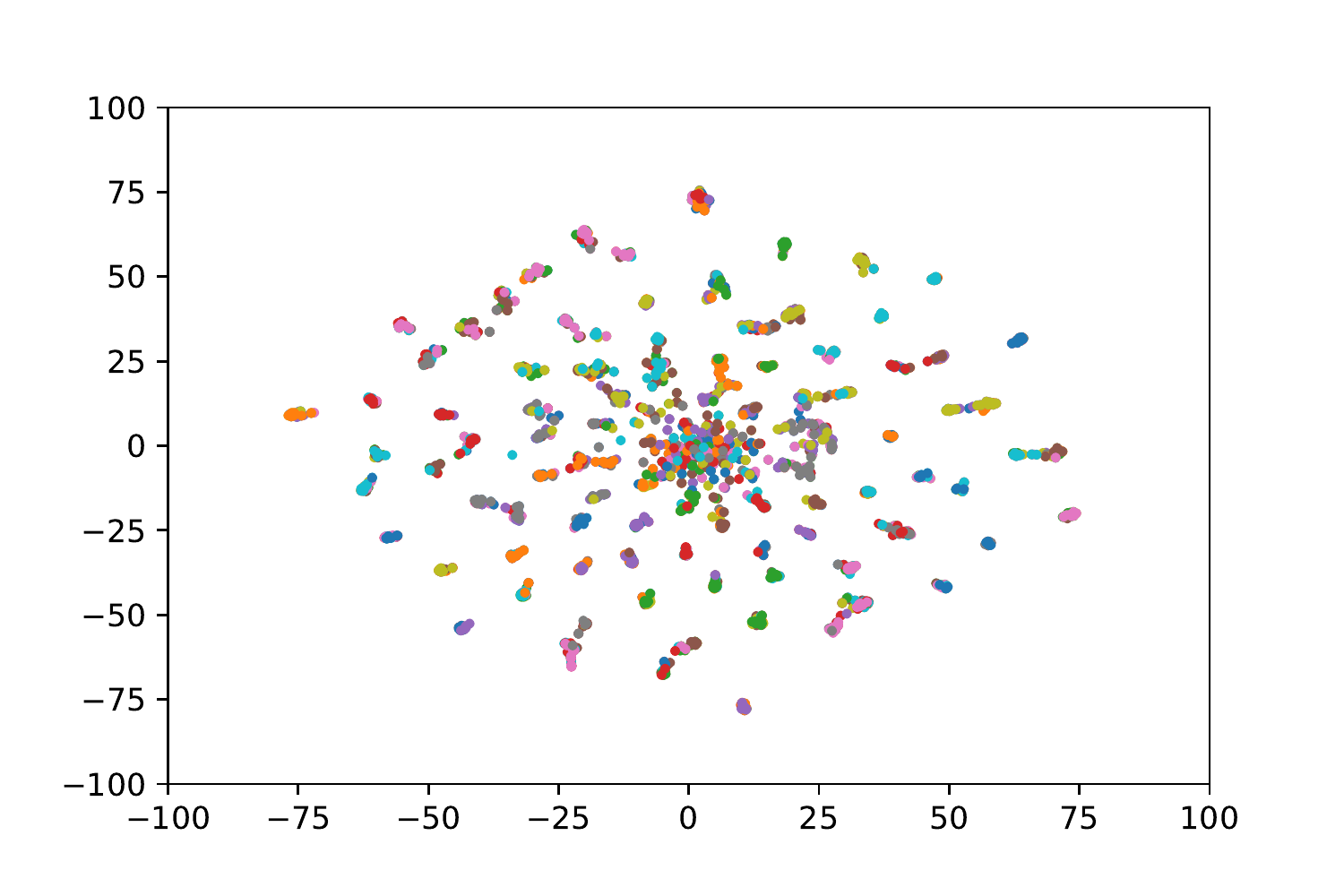}}\hfill
     \subfloat[][CAP + Encoding (NASNetMobile)]{\includegraphics[width=0.33\textwidth]{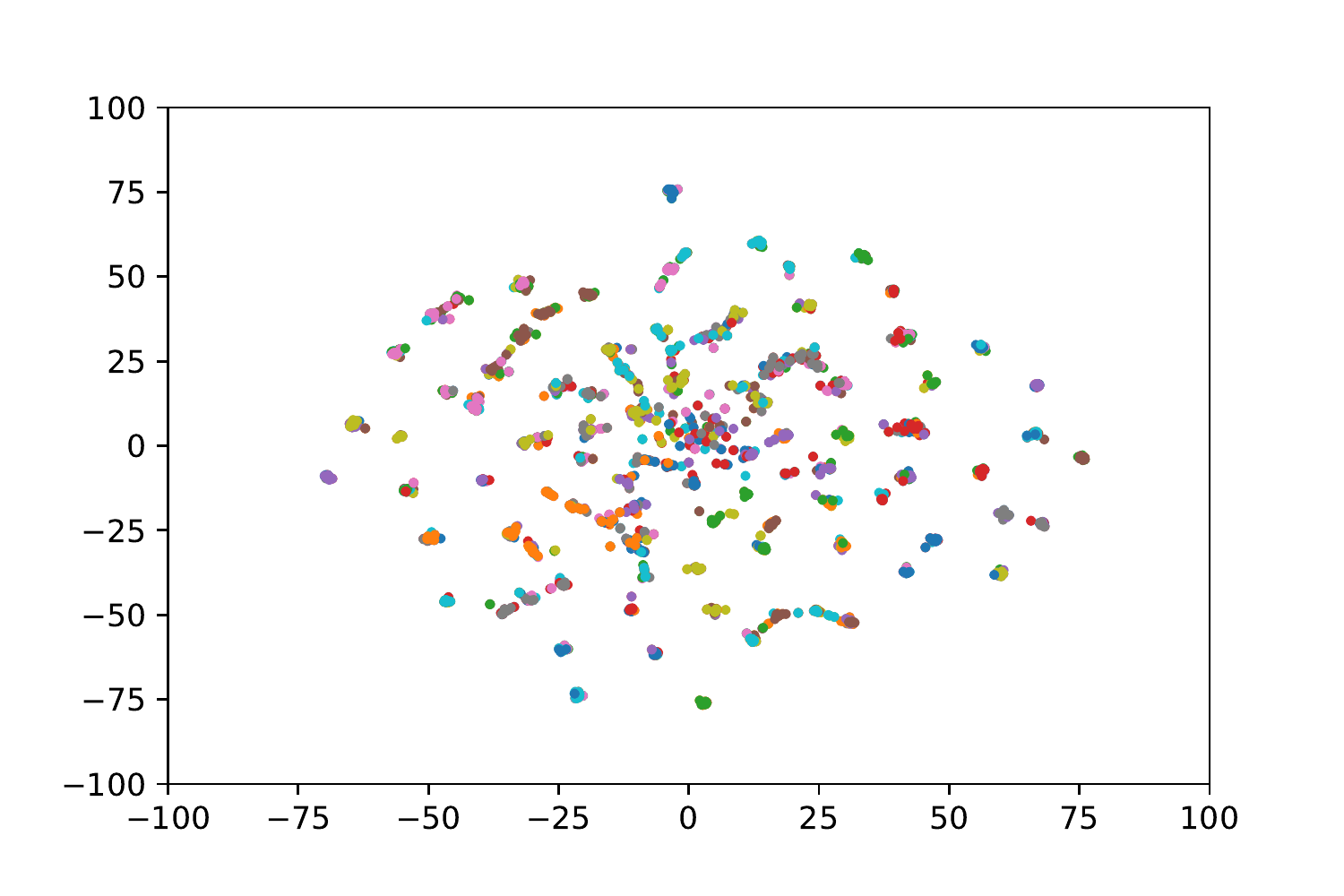}}\hfill
     \caption{Qualitative analysis of discriminating ability using t-SNE to monitor class separability and compactness. Visualization of \textbf{Aircraft} test images using Inception-V3 and NASNetMobile as a base CNN: (a \& d) output of the base CNN, (b \& e) feature maps from our attentional pooling (CAP), and (c \& f) our model's final feature maps (CAP+Encoding). Best view in color.}
     \label{fig:qual_1}
\end{figure*}

\begin{figure*}
     \centering
     \subfloat[][Base CNN (Inception-V3)]{\includegraphics[width=0.33\textwidth]{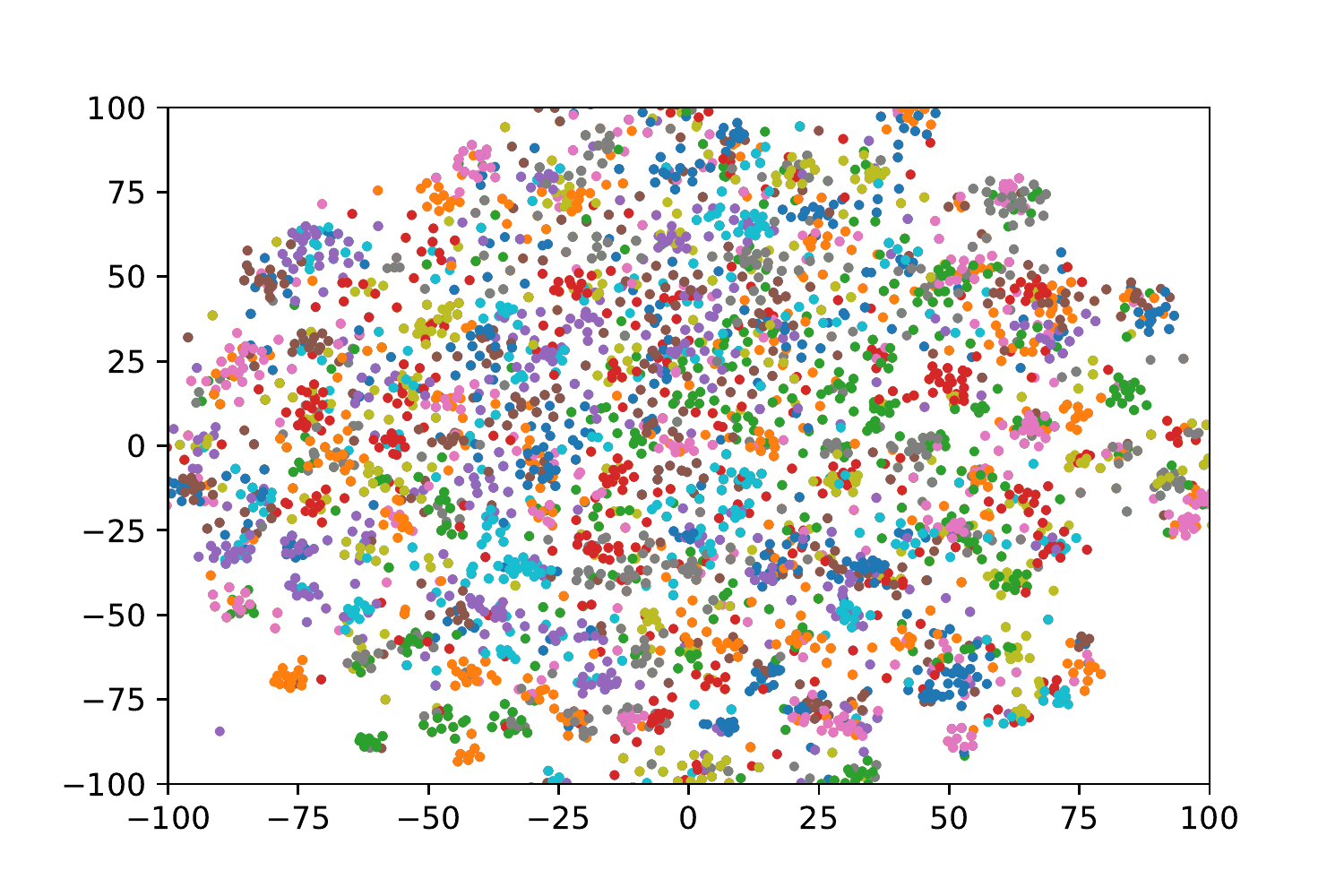}}\hfill
     \subfloat[][Attentional Pooling (Inception-V3)]{\includegraphics[width=0.33\textwidth]{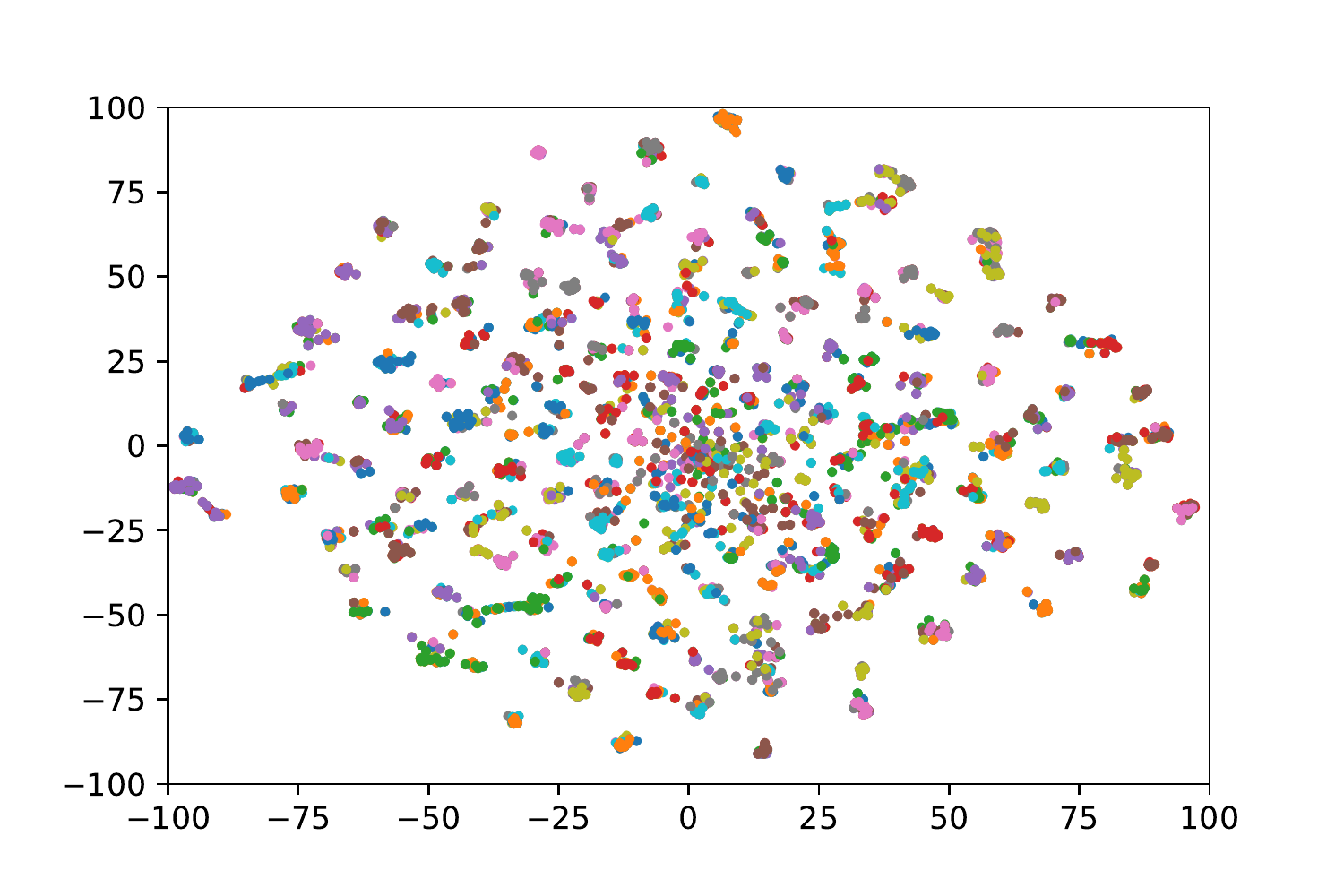}}\hfill
     \subfloat[][CAP + Encoding (Inception-V3)]{\includegraphics[width=0.33\textwidth]{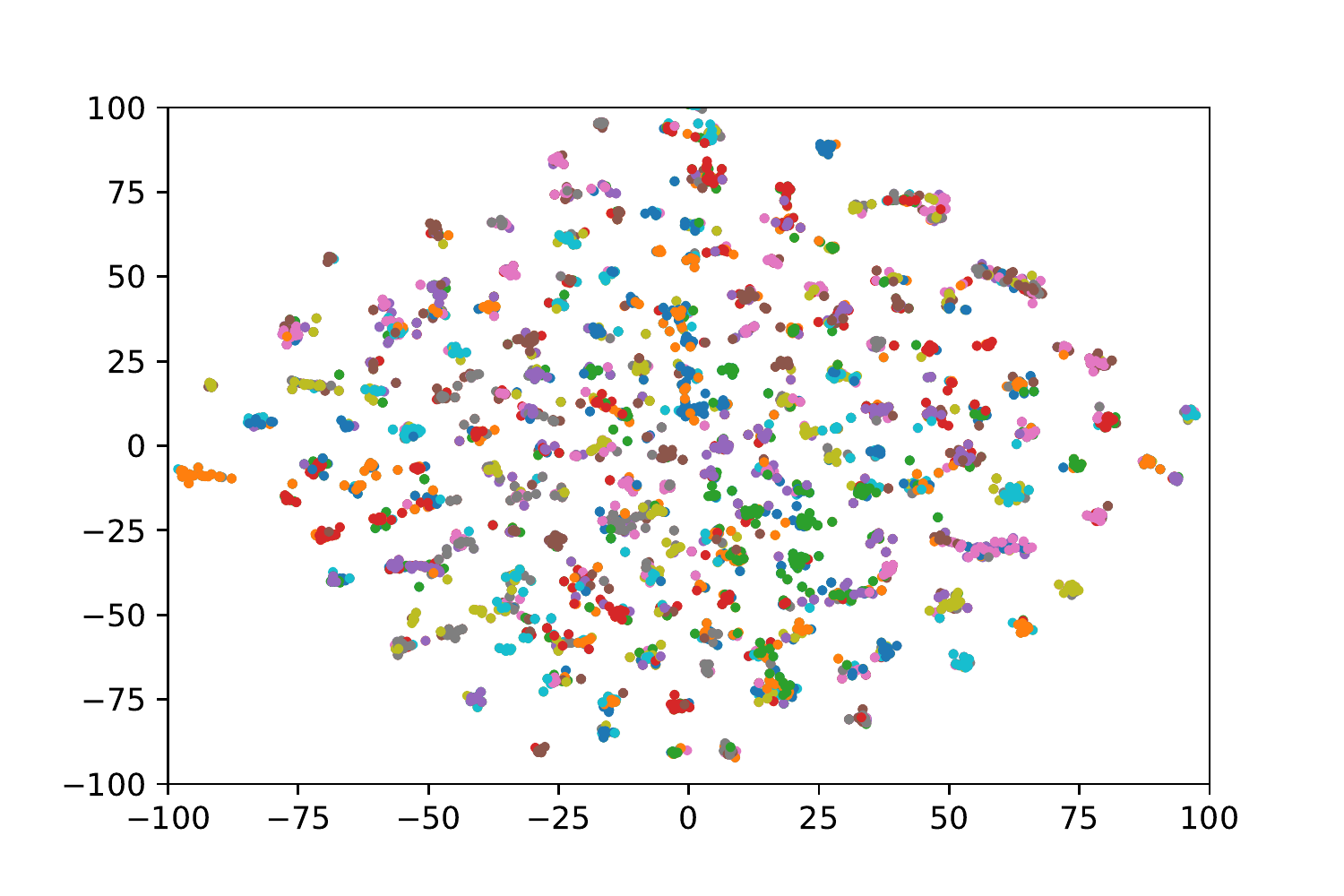}}\hfill
     
     \subfloat[][Base CNN (NASNetMobile)]{\includegraphics[width=0.33\textwidth]{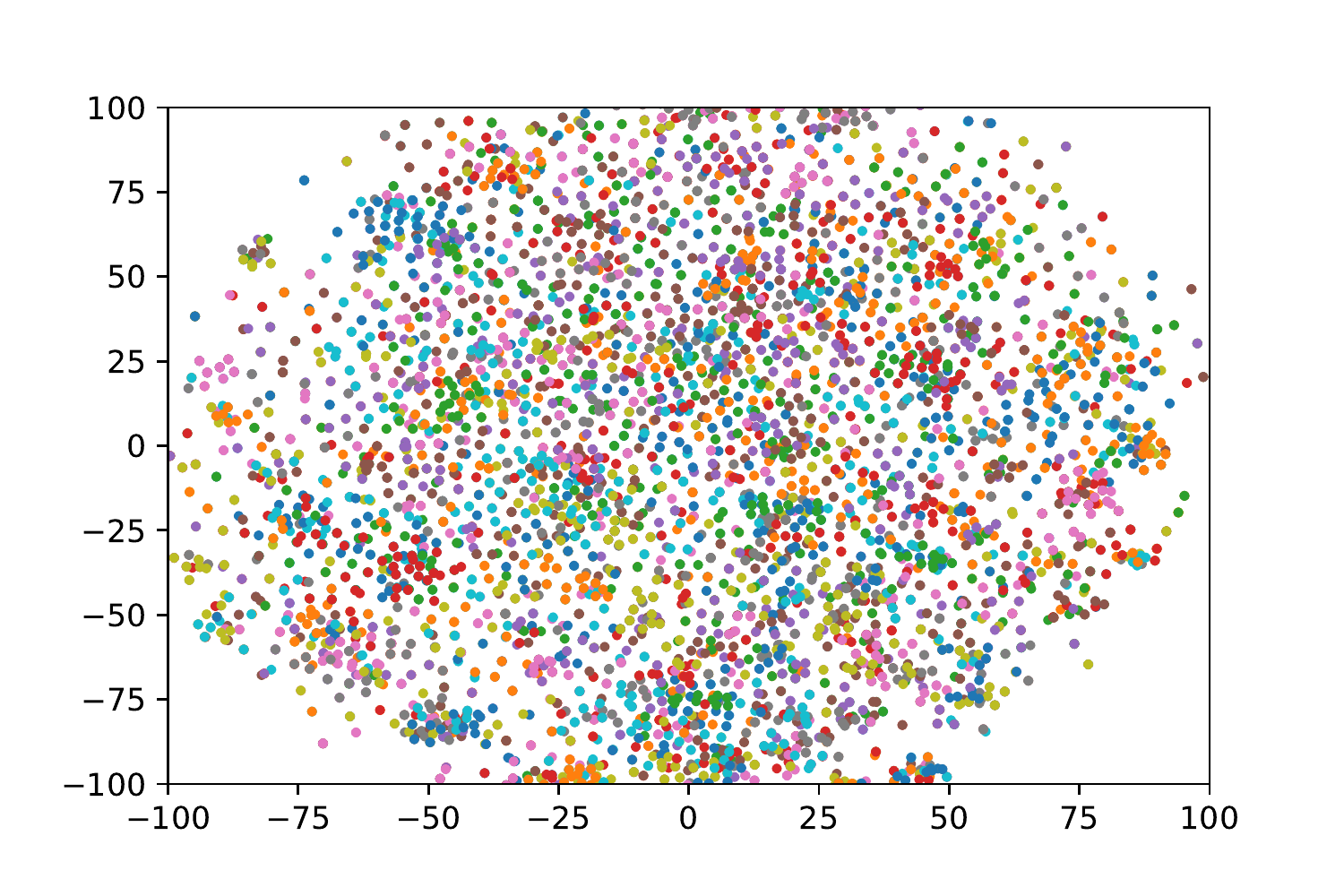}}\hfill
     \subfloat[][Attentional Pooling (NASNetMobile)]{\includegraphics[width=0.33\textwidth]{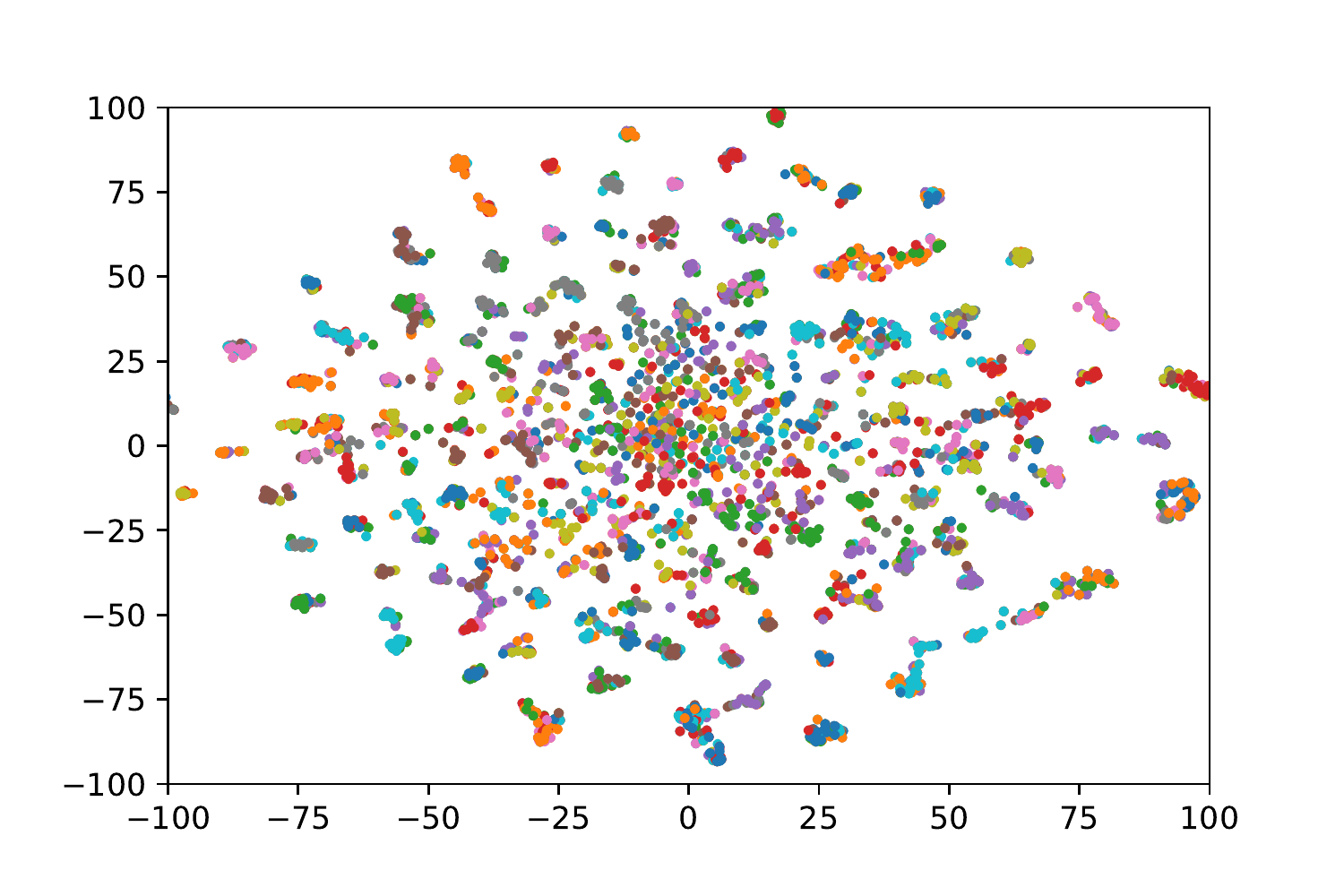}}\hfill
     \subfloat[][CAP + Encoding (NASNetMobile)]{\includegraphics[width=0.33\textwidth]{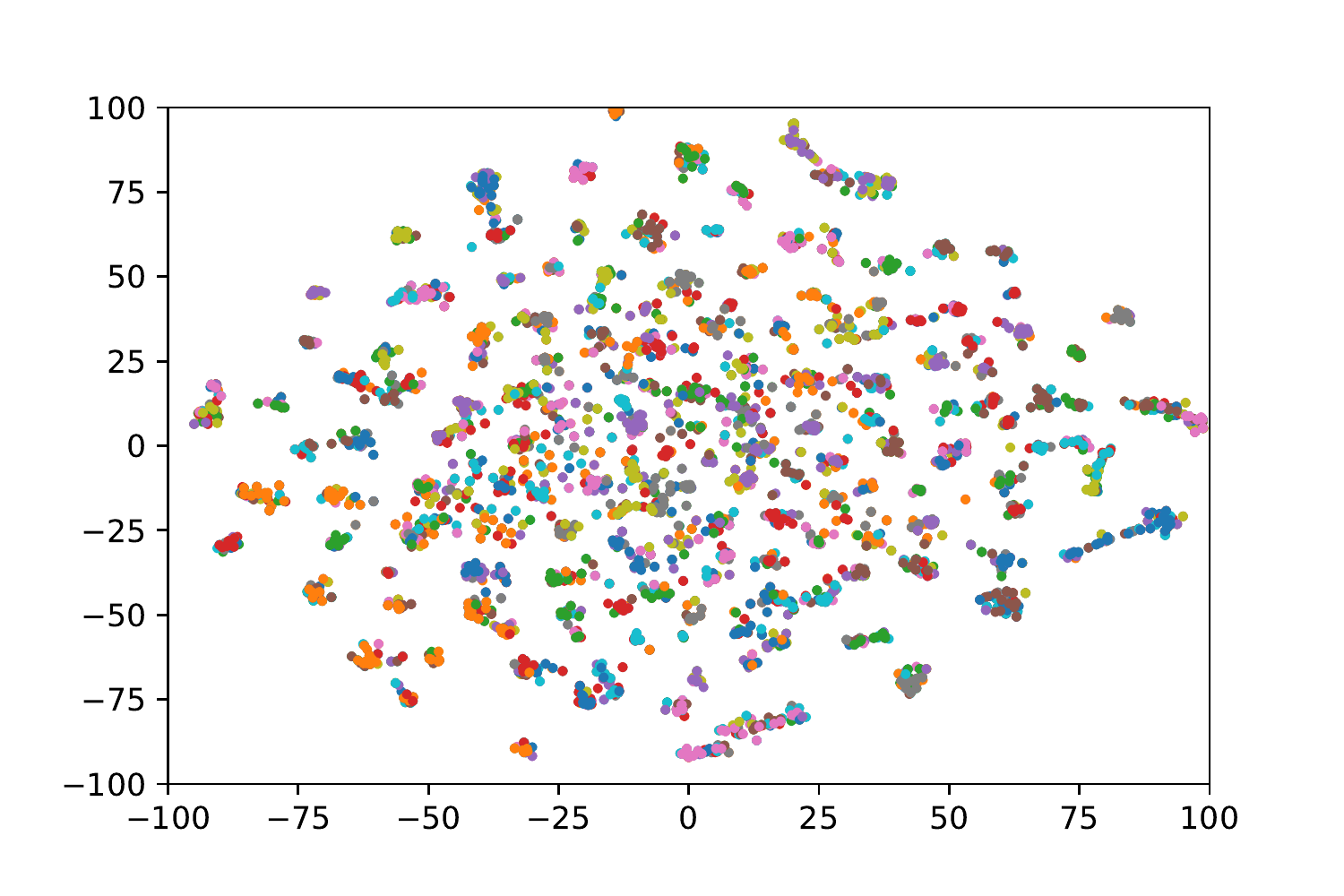}}\hfill
     \caption{Qualitative analysis of discriminating ability using t-SNE to monitor class separability and compactness. Visualization of \textbf{Stanford Cars} test images using Inception-V3 and NASNetMobile as a base CNN: (a \& d) output of the base CNN, (b \& e) feature maps from our attentional pooling (CAP), and (c \& f) our model's final feature maps (CAP+Encoding). Best view in color.}
     \label{fig:qual_2}
\end{figure*}

\begin{figure*}
     \centering
     \subfloat[][Base CNN (Inception-V3)]{\includegraphics[width=0.33\textwidth]{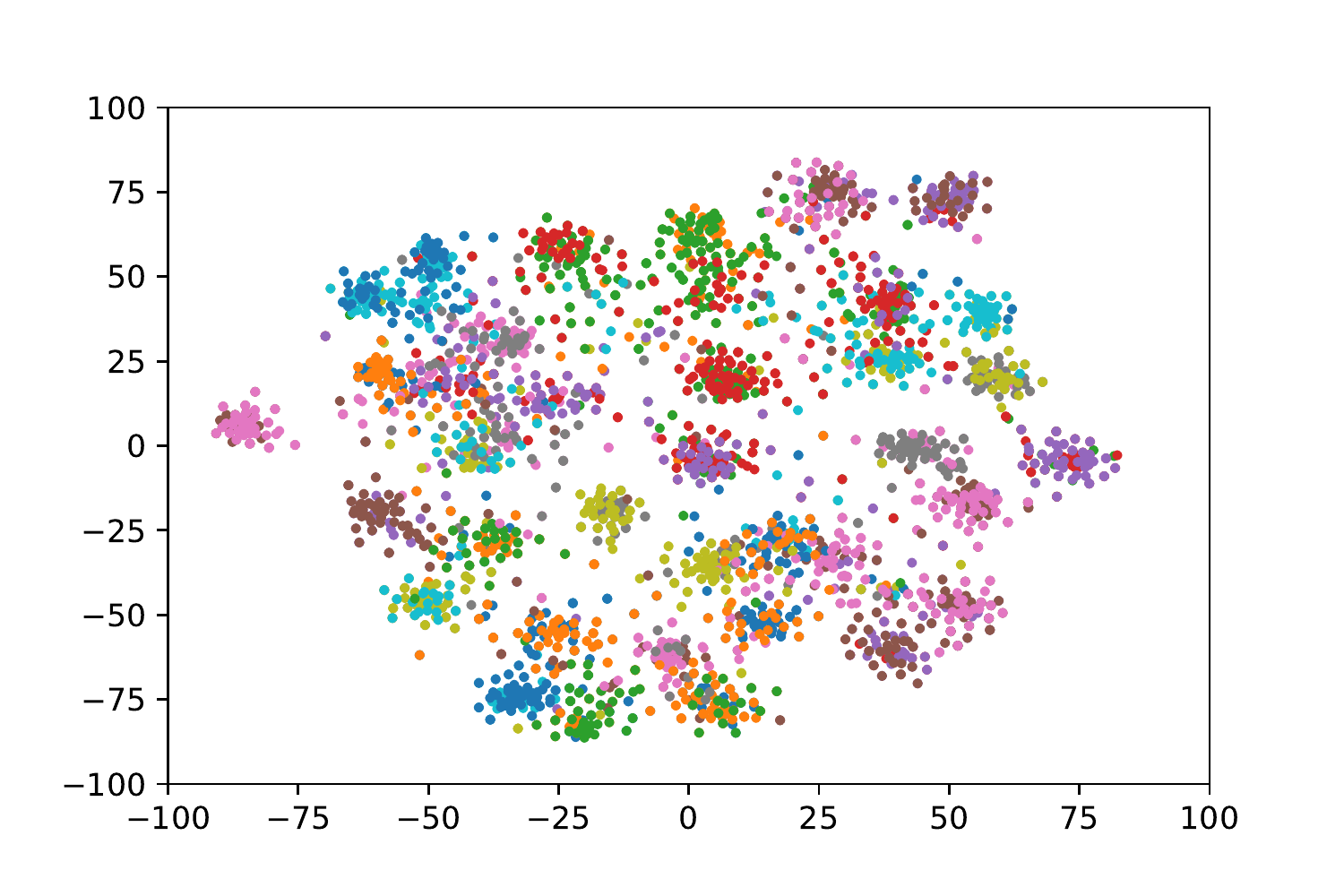}}\hfill
     \subfloat[][Attentional Pooling (Inception-V3)]{\includegraphics[width=0.33\textwidth]{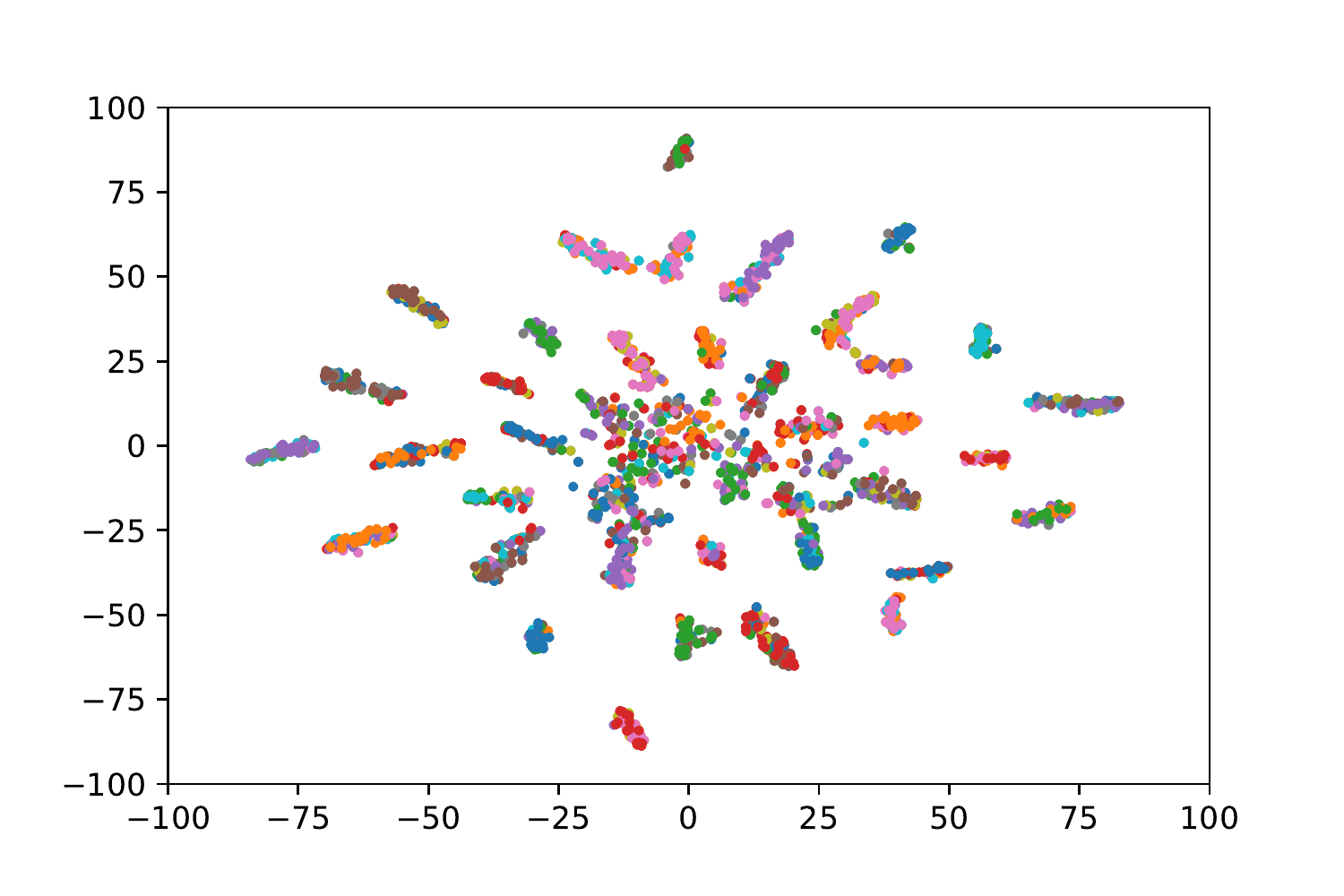}}\hfill
     \subfloat[][CAP + Encoding (Inception-V3)]{\includegraphics[width=0.33\textwidth]{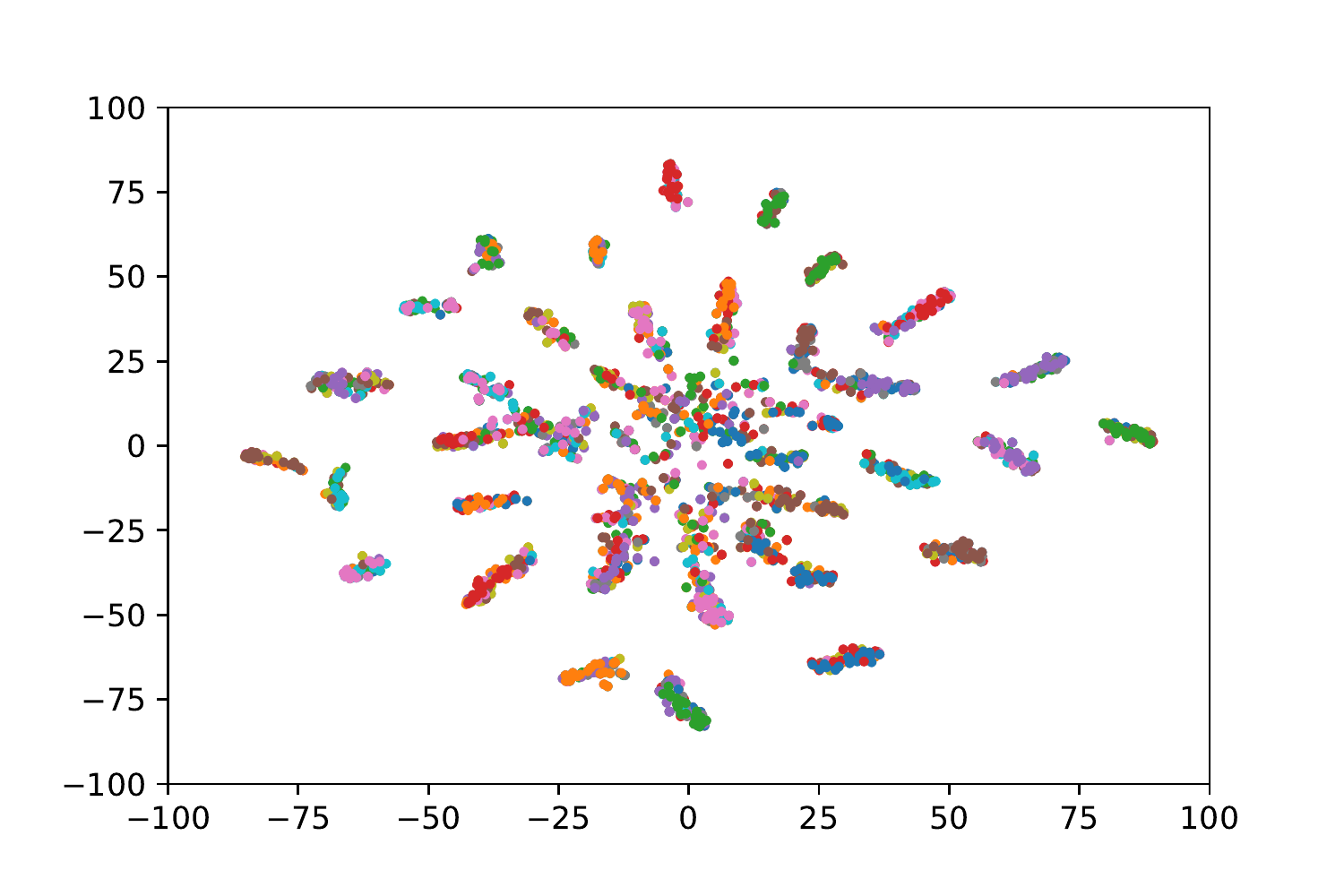}}\hfill
     
     \subfloat[][Base CNN (NASNetMobile)]{\includegraphics[width=0.33\textwidth]{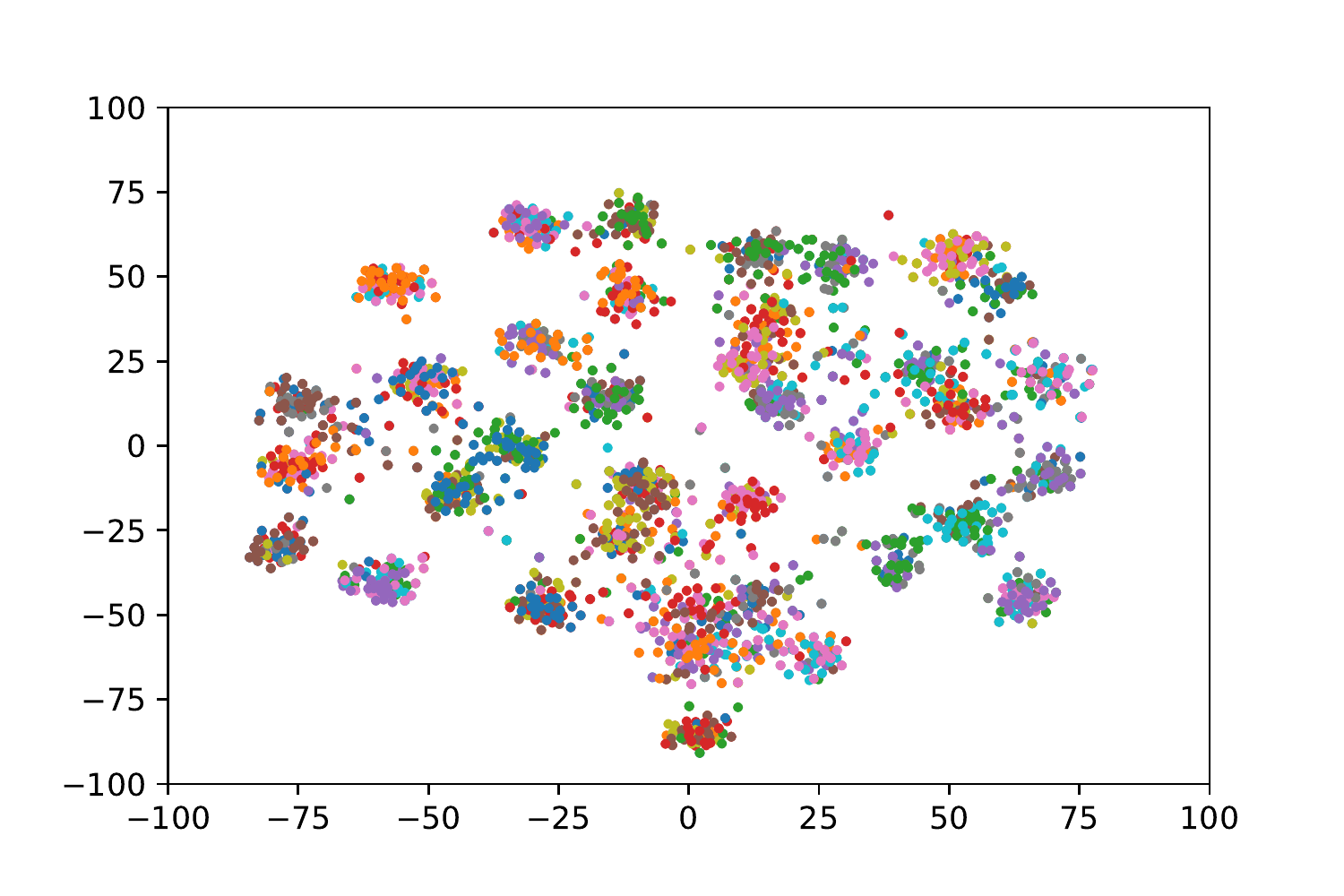}}\hfill
     \subfloat[][Attentional Pooling (NASNetMobile)]{\includegraphics[width=0.33\textwidth]{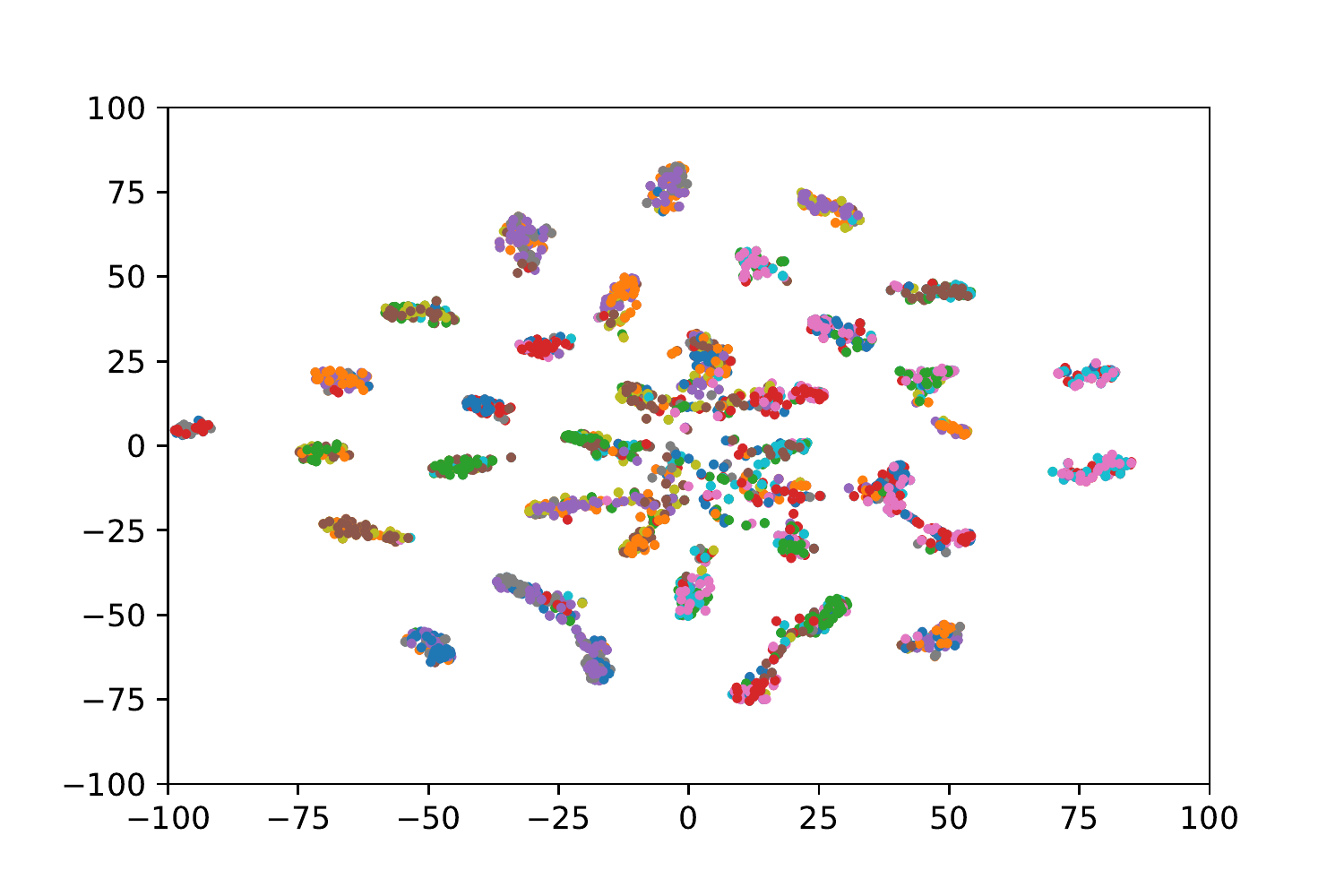}}\hfill
     \subfloat[][CAP + Encoding (NASNetMobile)]{\includegraphics[width=0.33\textwidth]{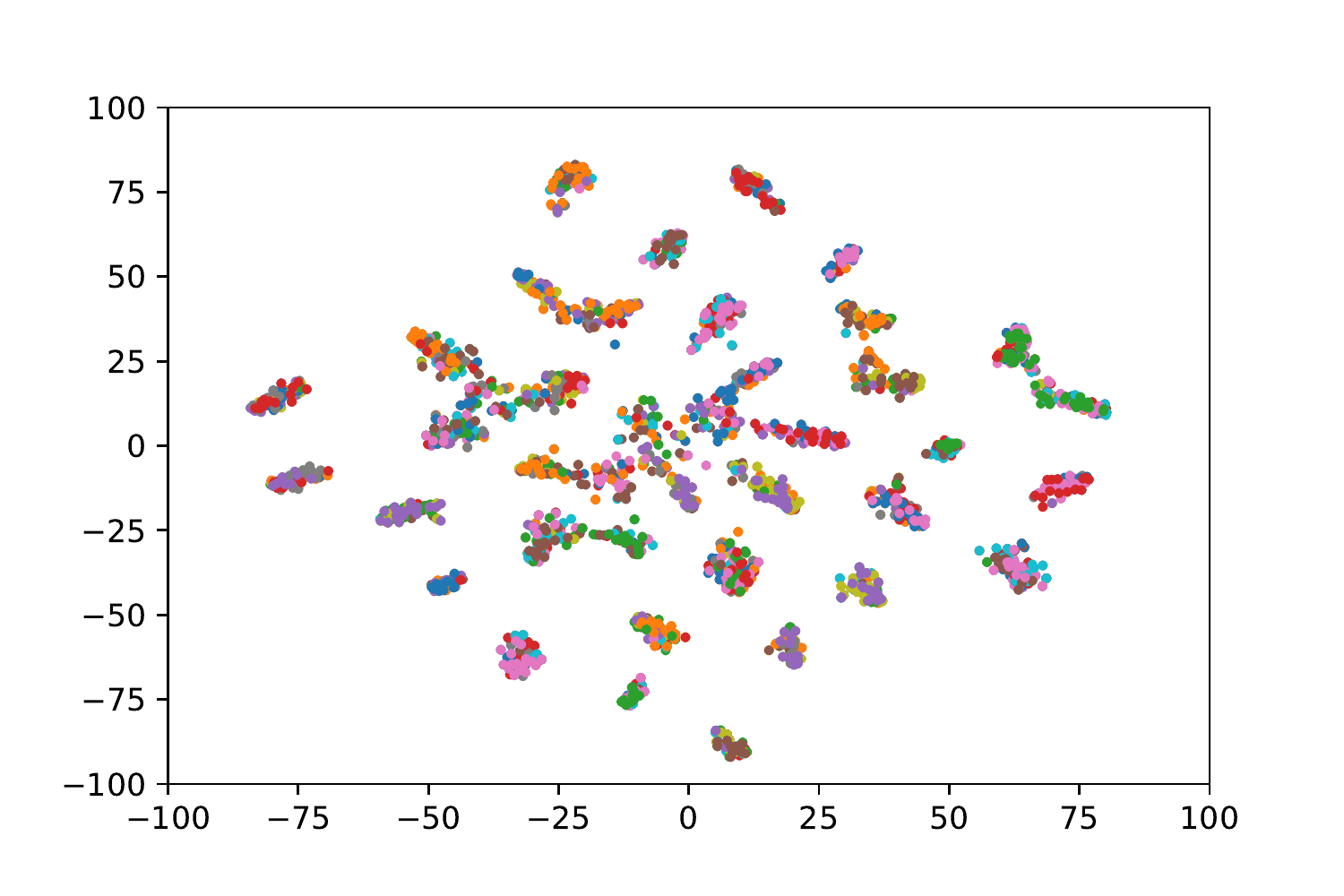}}\hfill
     \caption{Qualitative analysis of discriminating ability using t-SNE to monitor class separability and compactness. Visualization of \textbf{Oxford-IIIT Pets} test images using Inception-V3 and NASNetMobile as a base CNN: (a \& d) output of the base CNN, (b \& e) feature maps from our attentional pooling (CAP), and (c \& f) our model's final feature maps (CAP+Encoding). Best view in color.}
     \label{fig:qual_3}
\end{figure*}

\begin{figure*}
     \centering
     \subfloat[][Base CNN (MobileNetV2)]{\includegraphics[width=0.33\textwidth]{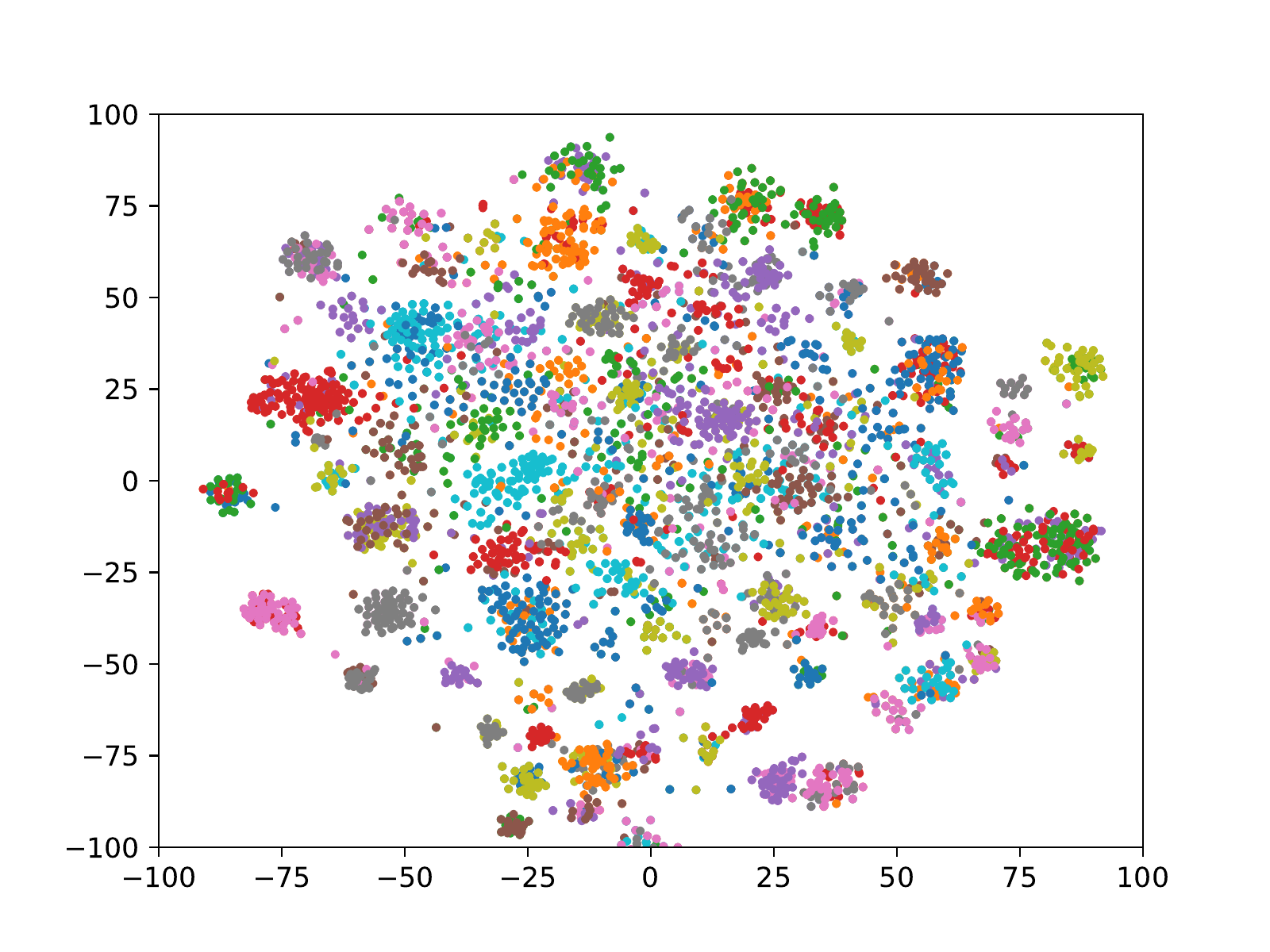}}\hfill
     \subfloat[][Attentional Pooling (MobileNetV2)]{\includegraphics[width=0.33\textwidth]{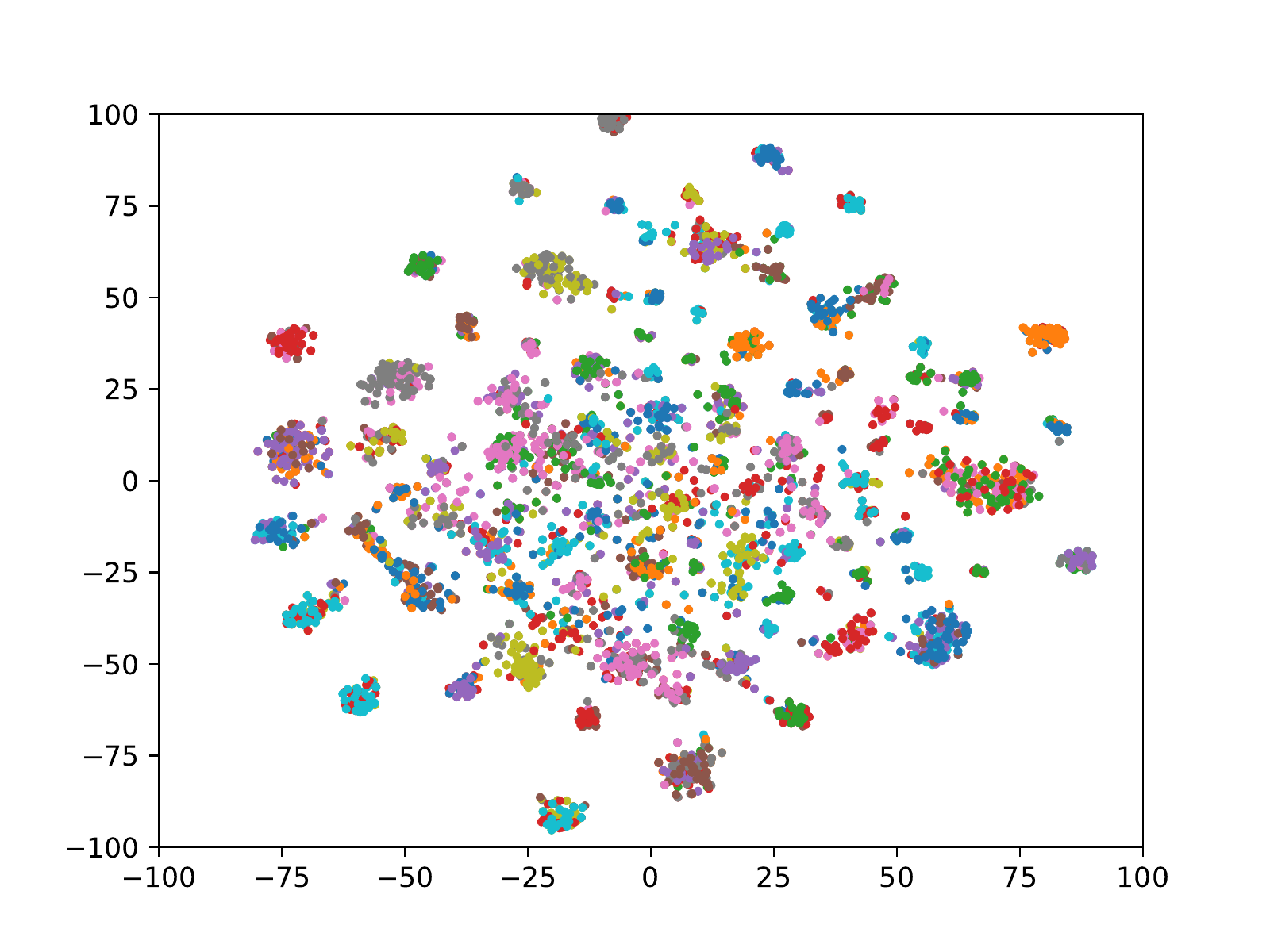}}\hfill
     \subfloat[][CAP + Encoding (MobileNetV2)]{\includegraphics[width=0.33\textwidth]{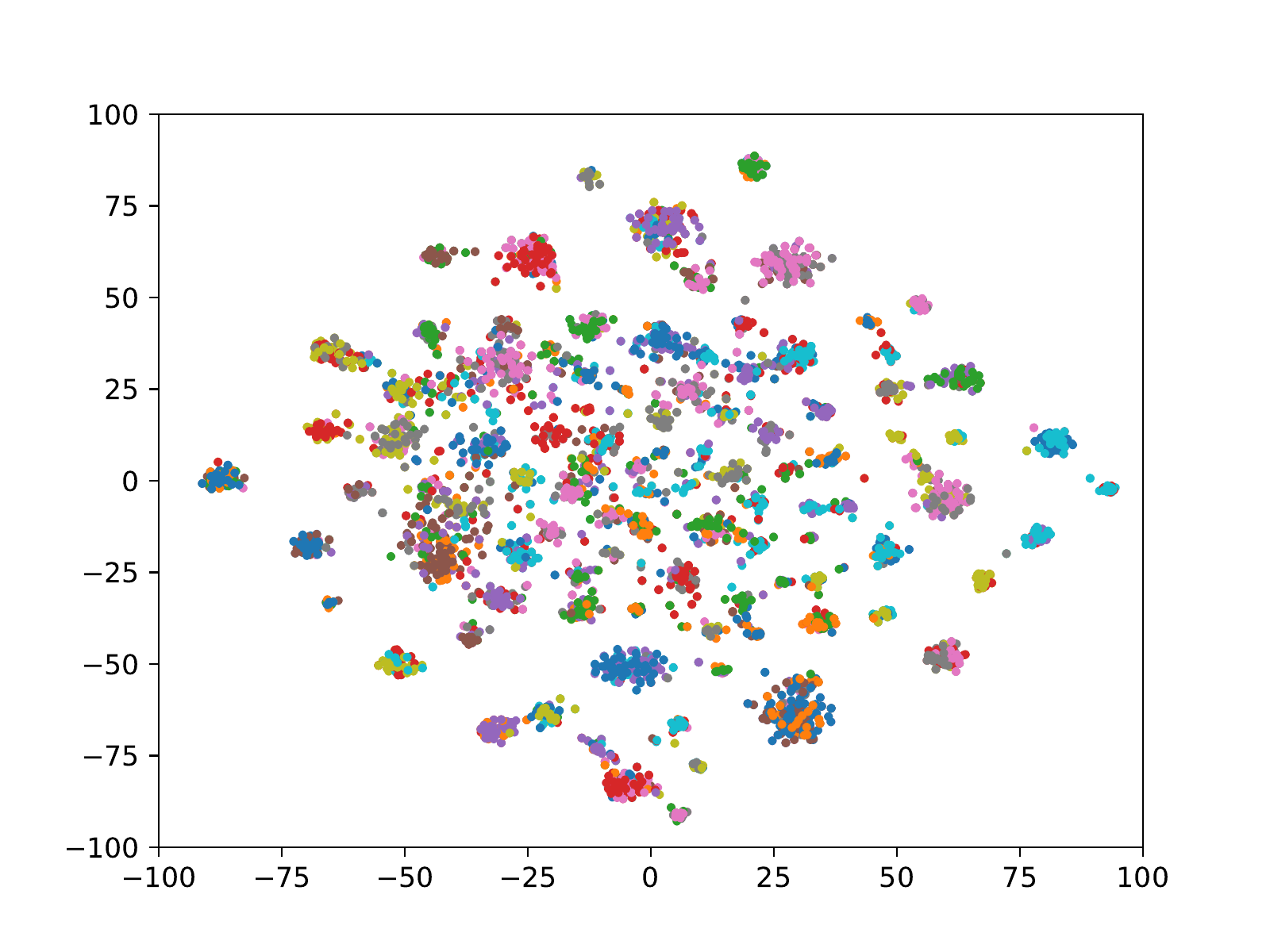}}\hfill
     
     \subfloat[][Base CNN (NASNetMobile)]{\includegraphics[width=0.33\textwidth]{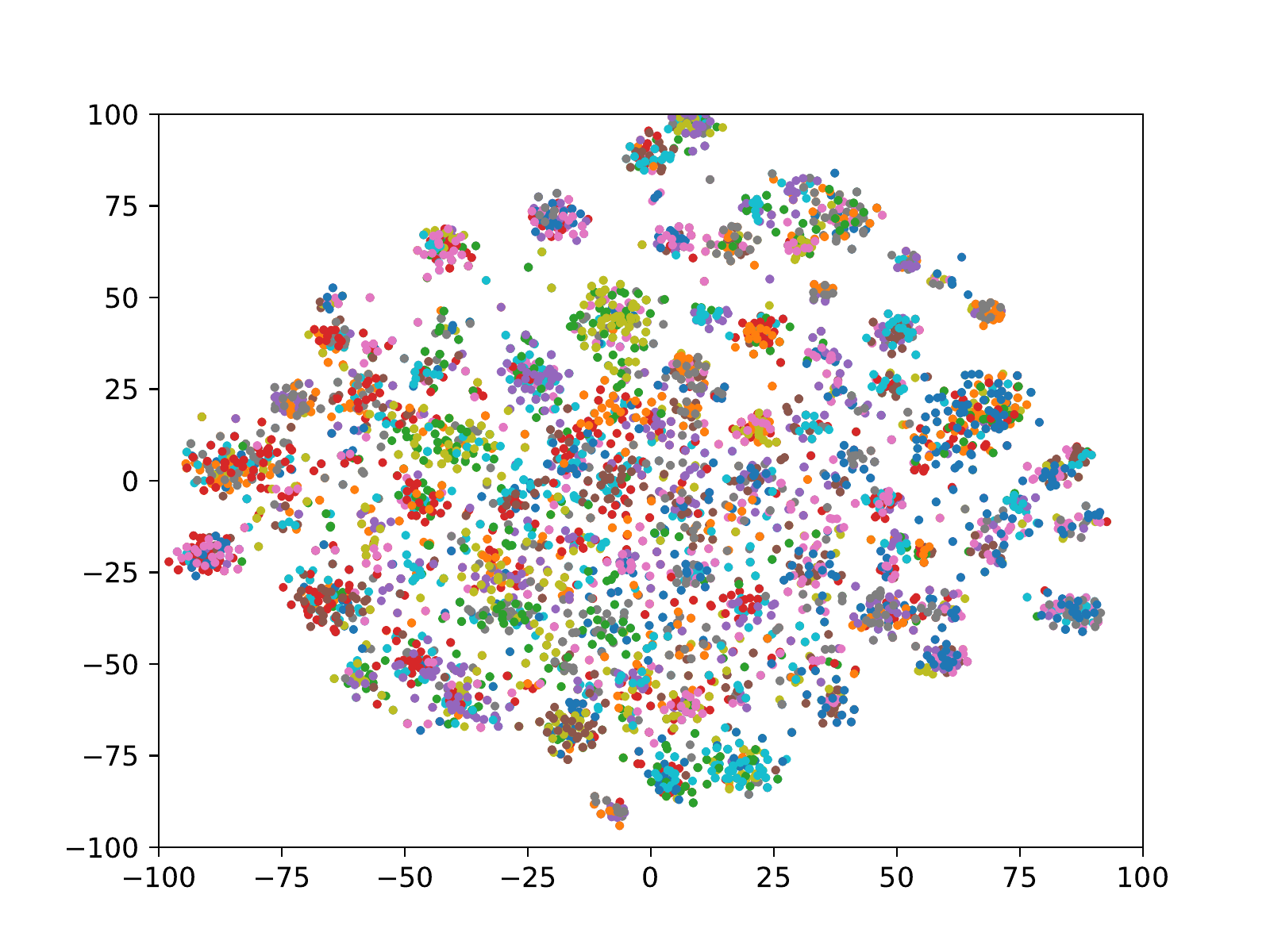}}\hfill
     \subfloat[][Attentional Pooling (NASNetMobile)]{\includegraphics[width=0.33\textwidth]{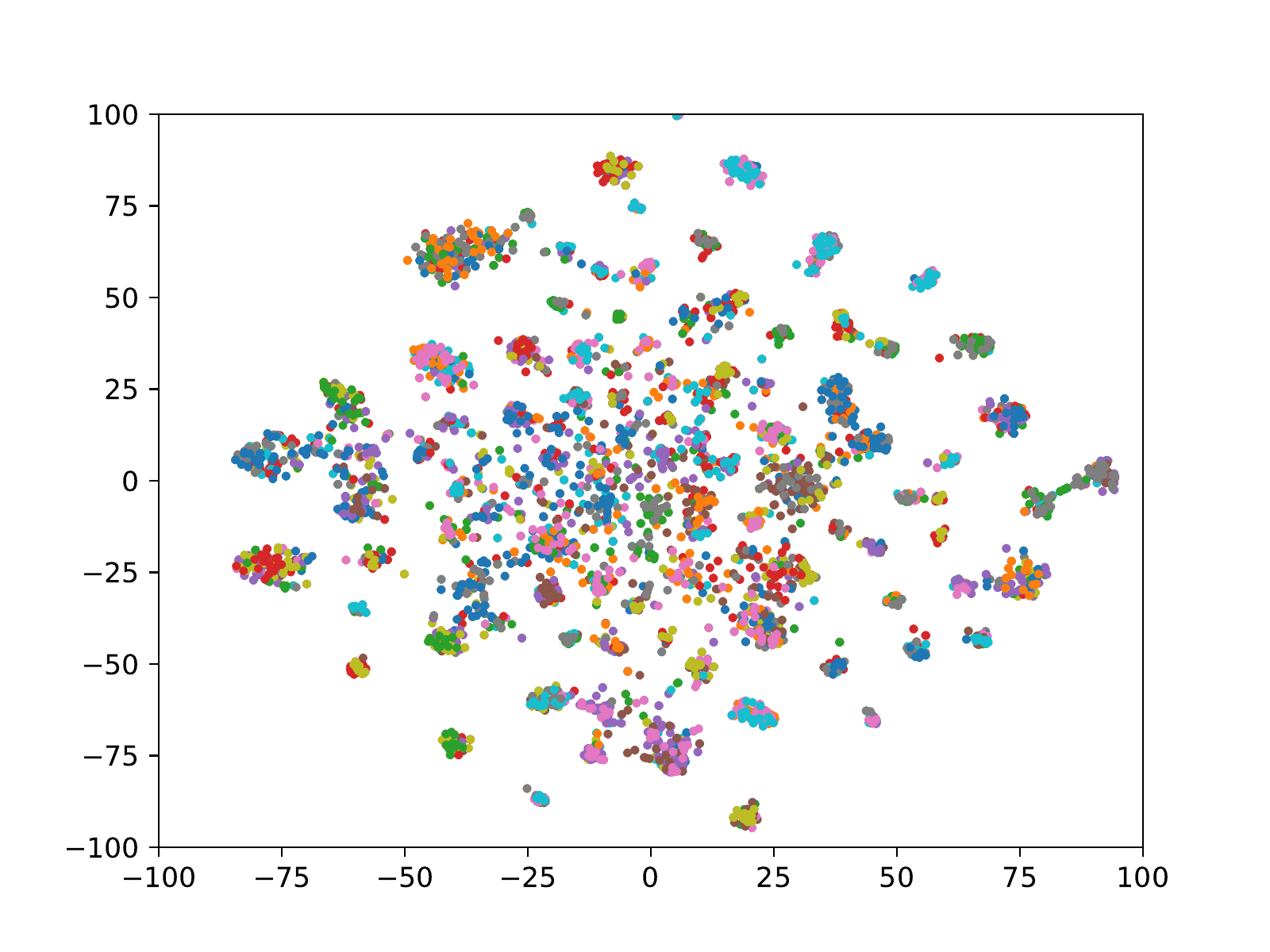}}\hfill
     \subfloat[][CAP + Encoding (NASNetMobile)]{\includegraphics[width=0.33\textwidth]{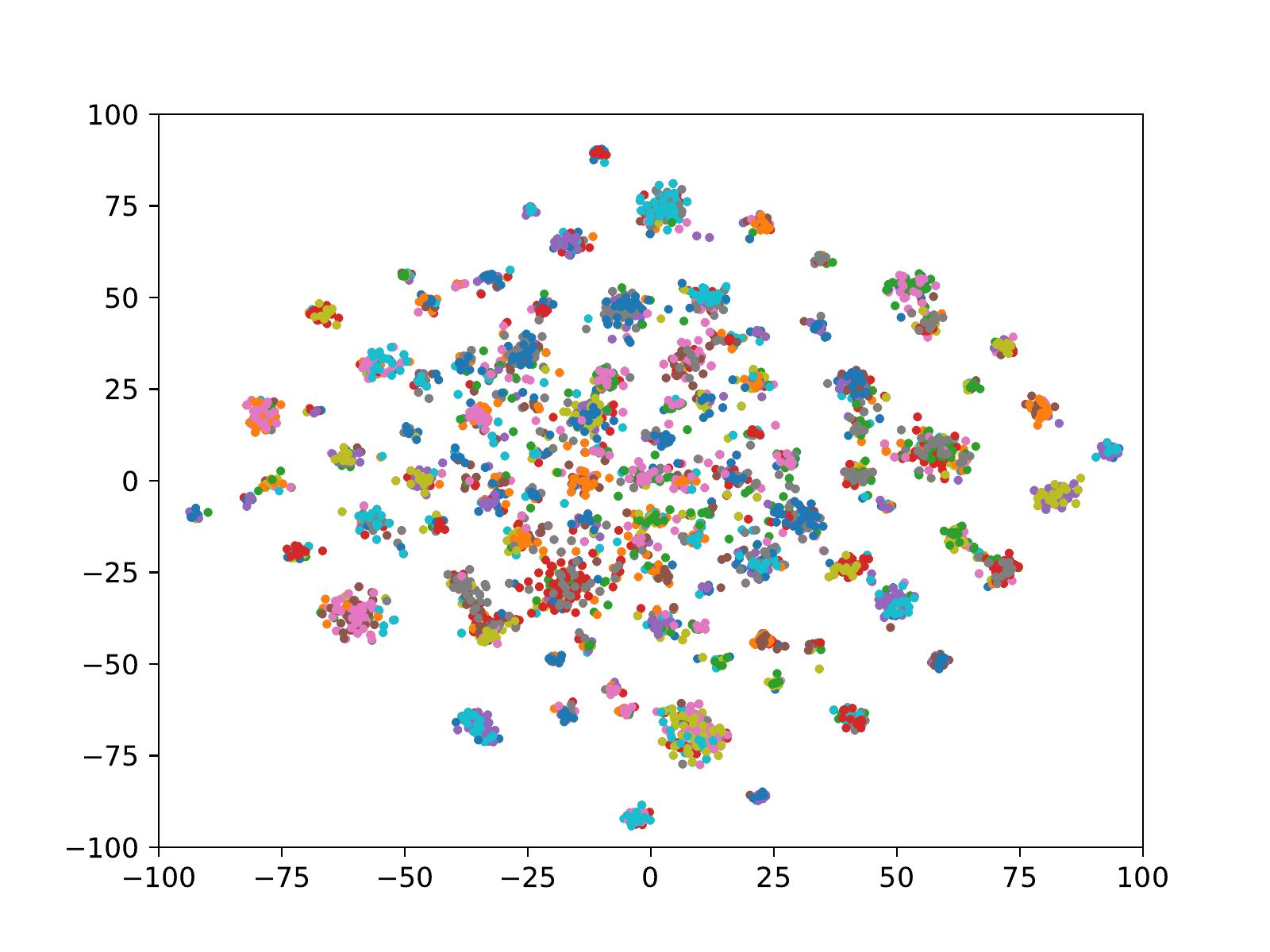}}\hfill
     \caption{Qualitative analysis of discriminating ability using t-SNE to monitor class separability and compactness. Visualization of \textbf{Oxford Flowers} test images using MobileNetV2 and NASNetMobile as a base CNN: (a \& d) output of the base CNN, (b \& e) feature maps from our attentional pooling (CAP), and (c \& f) our model's final feature maps (CAP+Encoding). Best view in color.}
     \label{fig:qual_4}
\end{figure*}

\begin{figure*}
     \centering
     \subfloat[][Base CNN (MobileNetV2)]{\includegraphics[width=0.33\textwidth]{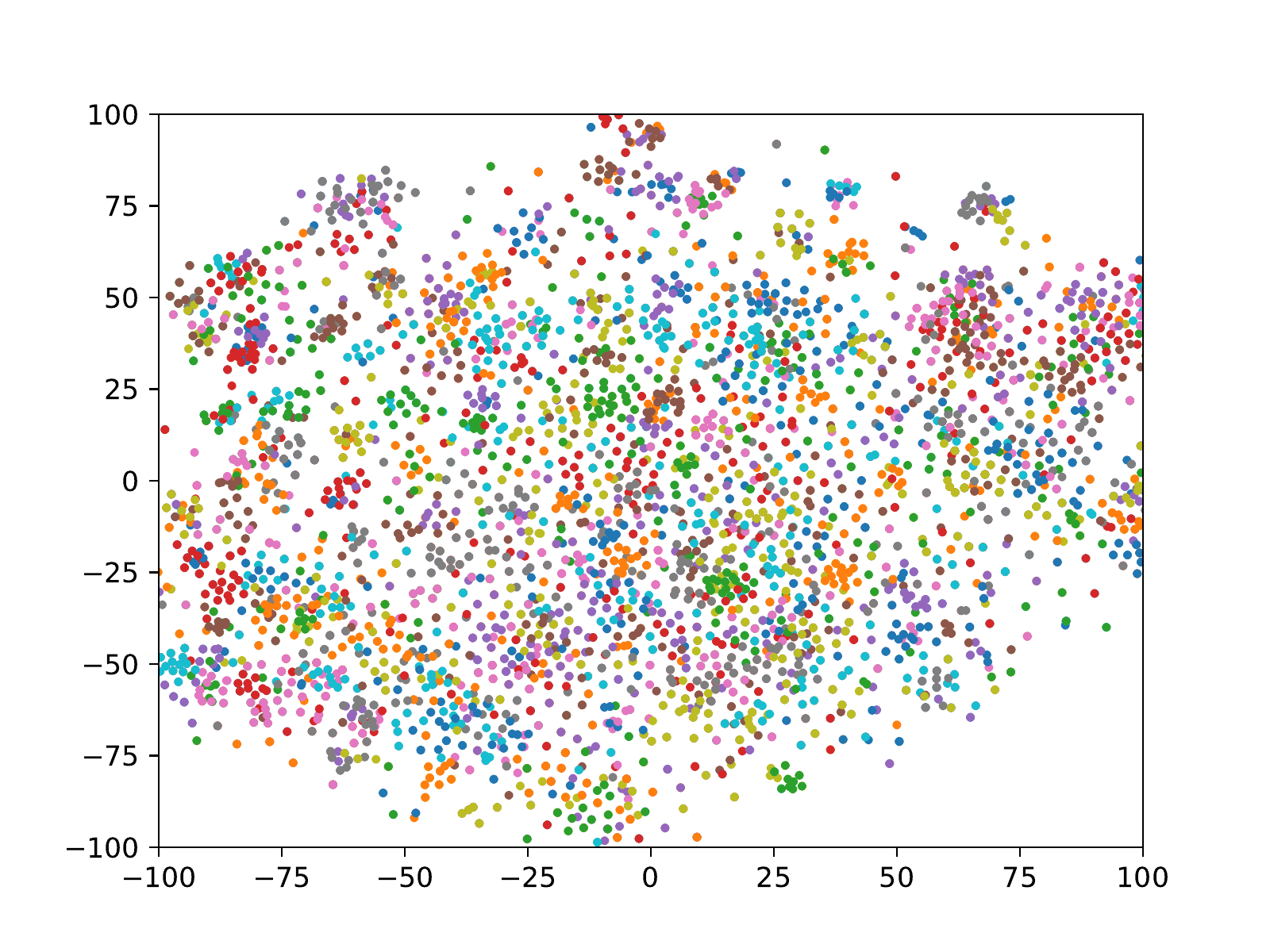}}\hfill
     \subfloat[][Attentional Pooling (MobileNetV2)]{\includegraphics[width=0.33\textwidth]{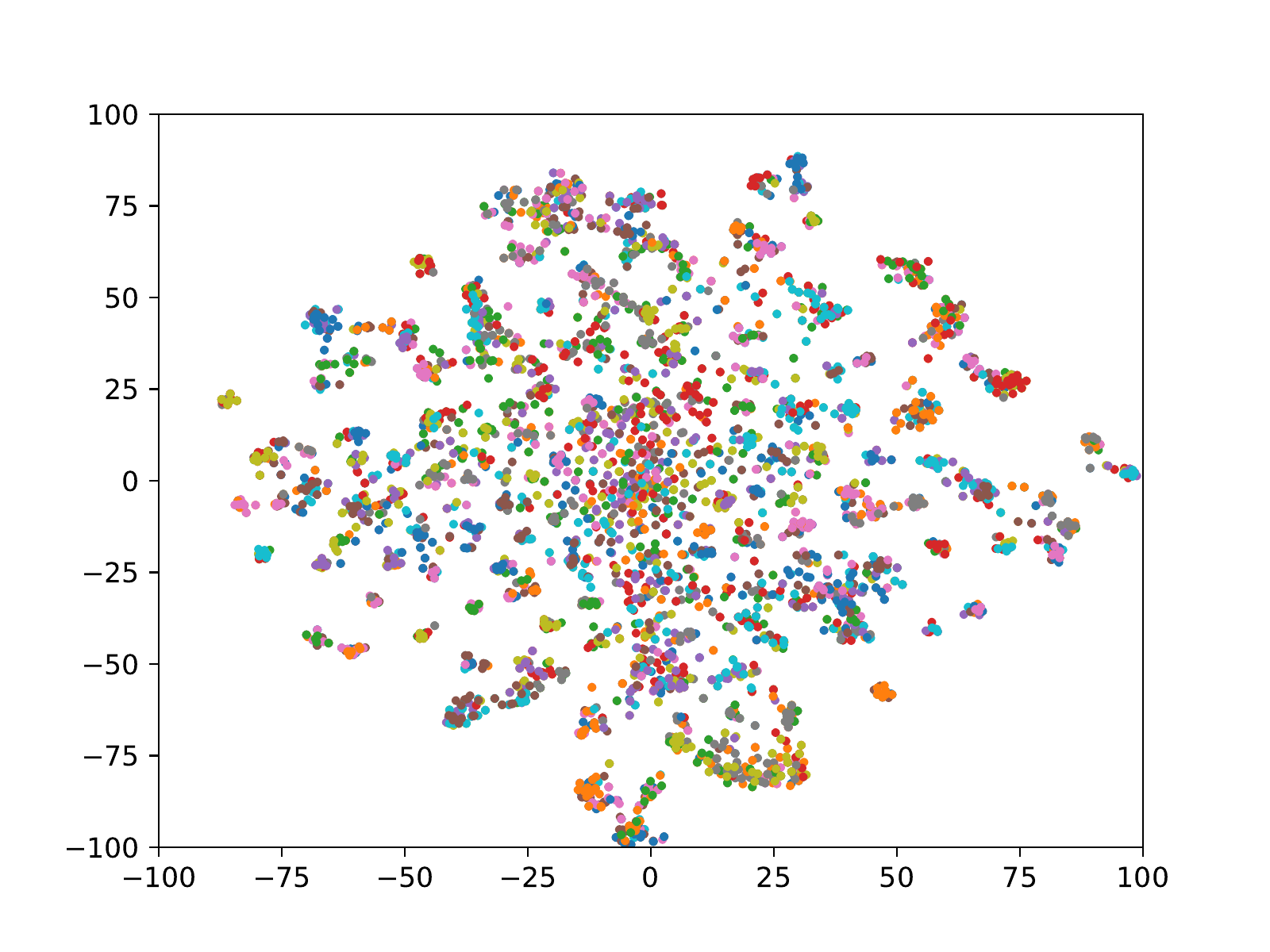}}\hfill
     \subfloat[][CAP + Encoding (MobileNetV2)]{\includegraphics[width=0.33\textwidth]{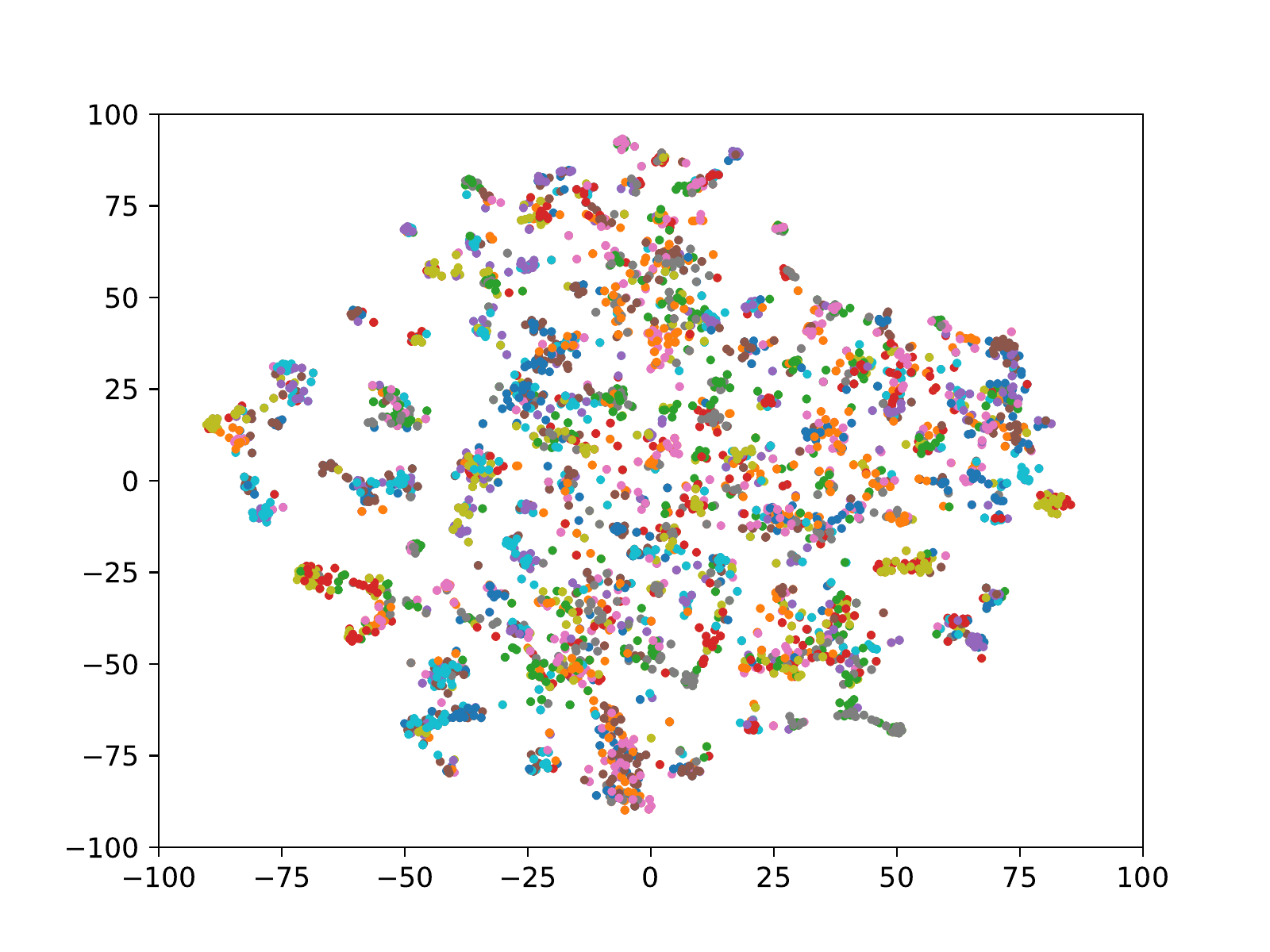}}\hfill
     
     \subfloat[][Base CNN (NASNetMobile)]{\includegraphics[width=0.33\textwidth]{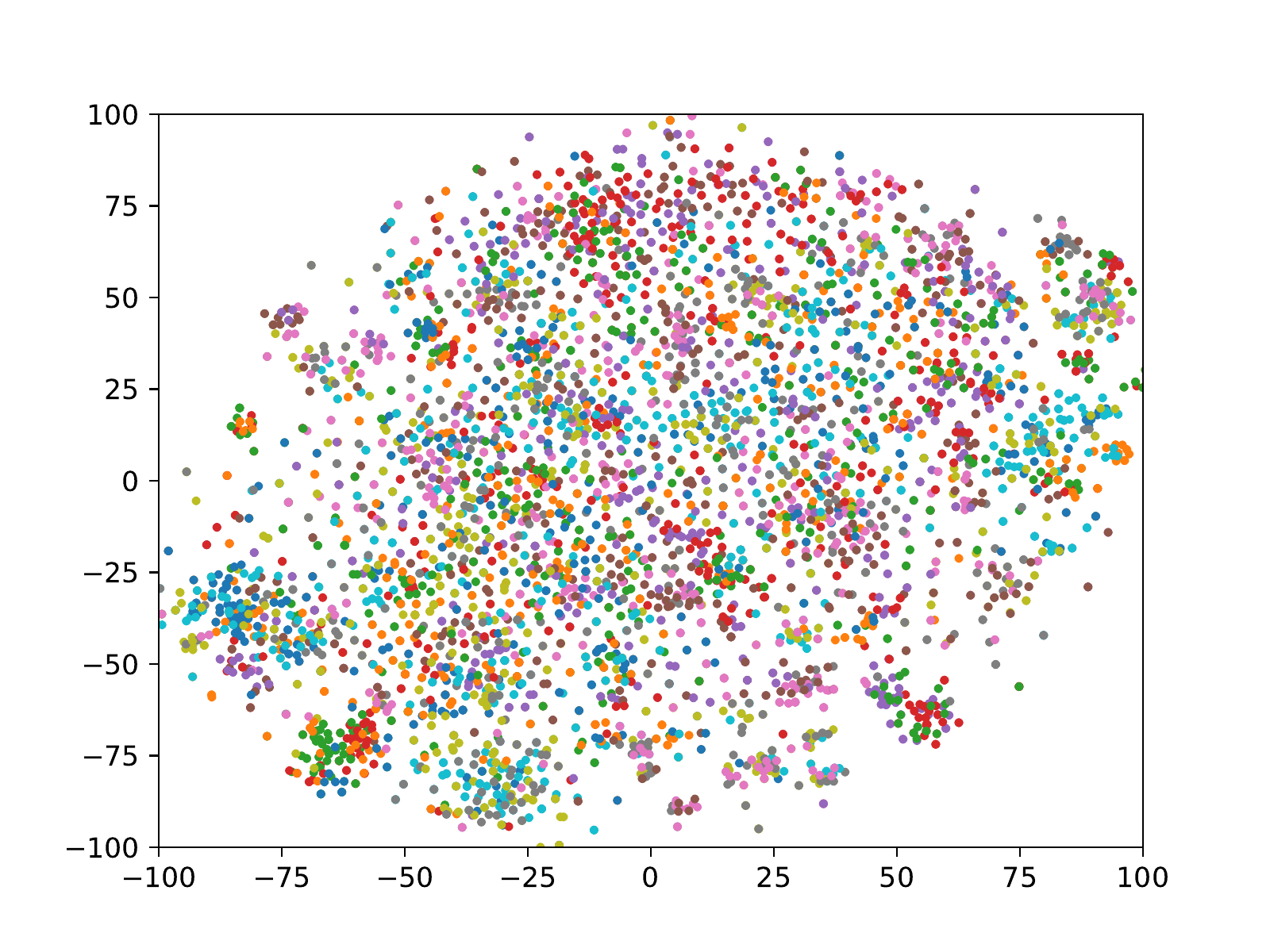}}\hfill
     \subfloat[][Attentional Pooling (NASNetMobile)]{\includegraphics[width=0.33\textwidth]{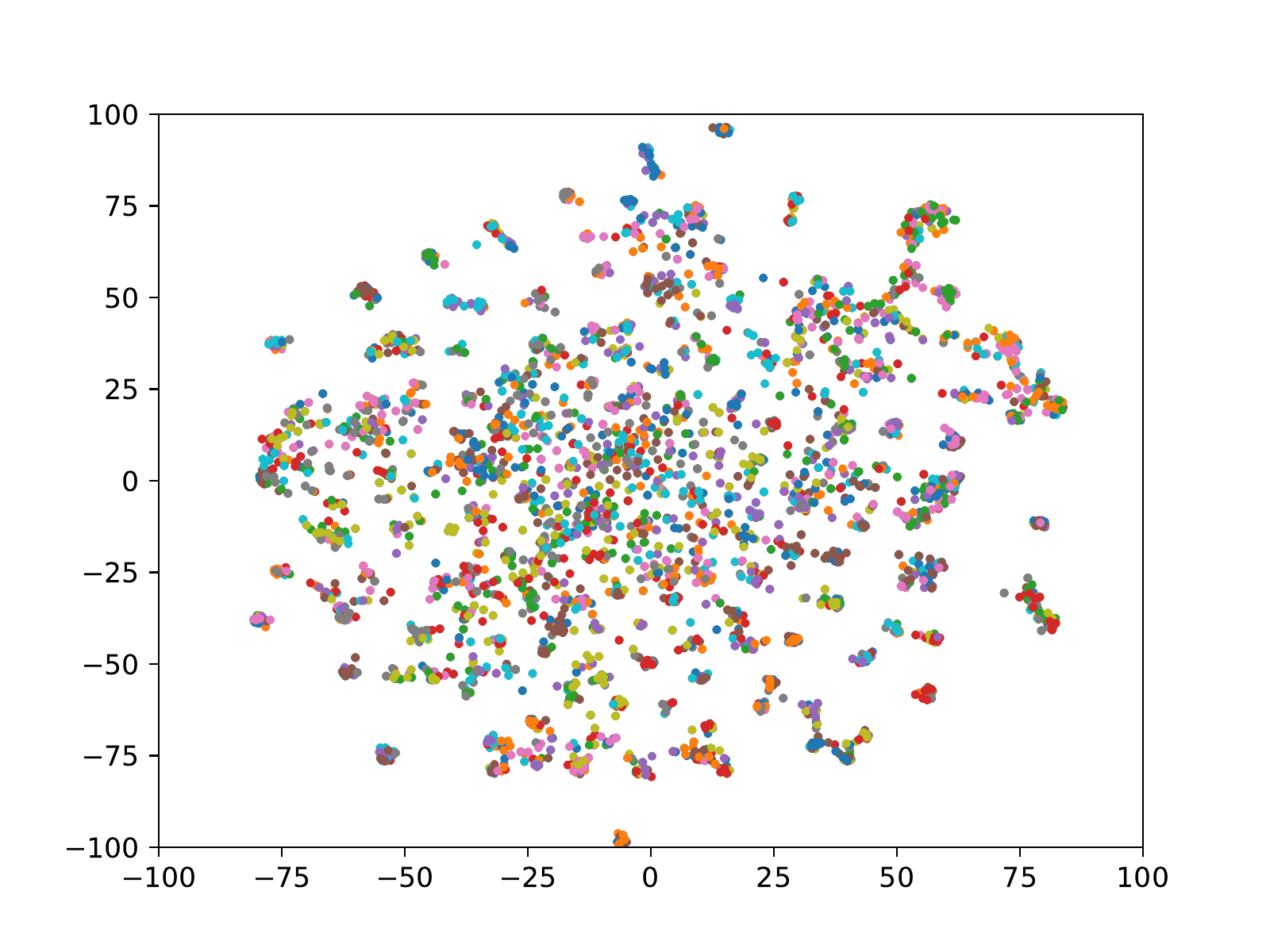}}\hfill
     \subfloat[][CAP + Encoding (NASNetMobile)]{\includegraphics[width=0.33\textwidth]{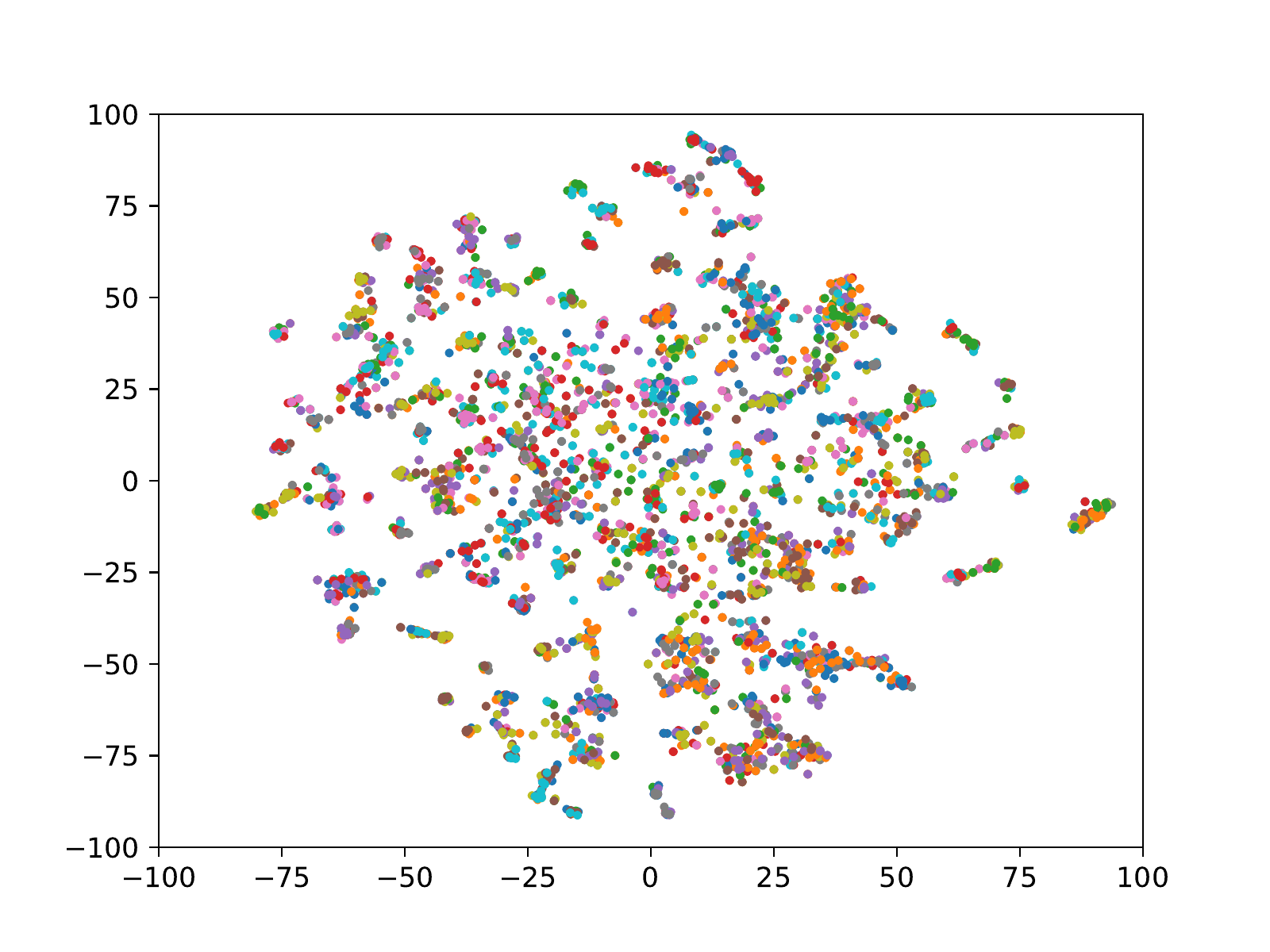}}\hfill
     \caption{Qualitative analysis of discriminating ability using t-SNE to monitor class separability and compactness. Visualization of the \textbf{Caltech-UCSD Birds (CUB-200)} test images using MobileNetV2 and NASNetMobile as a base CNN: (a \& d) output of the base CNN, (b \& e) feature maps from our attentional pooling (CAP), and (c \& f) our model's final feature maps (CAP+Encoding). Best view in color.}
     \label{fig:qual_5}
\end{figure*}

\end{document}